\documentclass{article}
\ifdefined\pdfsuppressptexinfo\pdfsuppressptexinfo=-1\fi

\usepackage{microtype}
\usepackage{graphicx}
\usepackage{subfigure}
\usepackage{booktabs}
\usepackage{amsmath}
\usepackage{amssymb}
\usepackage{multirow}
\usepackage{xcolor}
\usepackage{colortbl}
\usepackage{xspace}
\usepackage{tikz}
\usetikzlibrary{arrows.meta,positioning}
\usepackage{hyperref}

\usepackage[accepted]{mlsys2025}
\usepackage{float}
\makeatletter\renewcommand{\Notice@String}{}\makeatother
\hypersetup{pdftitle={SafeStart: Mitigating the Overlooked Safety Cost of Active Expert Reduction in MoE LLM Serving},pdfauthor={Erfan Baghaei Potraghloo, Saeid Shokoufa, Seyedarmin Azizi, Arya Fayyazi, Souvik Kundu, Mehdi Kamal, Massoud Pedram},pdfsubject={},pdfkeywords={Mixture of Experts, LLM Serving, Safety, Evaluation}}

\newcommand{\pp}{\,pp\xspace}
\newcommand{\sci}[2]{#1{\times}10^{#2}}
\newcommand{\method}{SafeStart\xspace}
\definecolor{vizblue}{HTML}{2A78D6}
\definecolor{vizorange}{HTML}{EB6834}
\definecolor{vizaqua}{HTML}{1BAF7A}
\definecolor{vizink}{HTML}{0B0B0B}
\definecolor{vizink2}{HTML}{52514E}
\definecolor{vizaxis}{HTML}{C3C2B7}
\definecolor{vizgrid}{HTML}{E1E0D9}
\newcommand{\slots}[2]{\begin{tikzpicture}[baseline=-0.55ex,x=4.1pt]\foreach \i in {1,...,#2}{\ifnum\i>#1\relax\draw[vizink2!50,line width=0.45pt] (\i-1,-1.45pt) rectangle ++(2.9pt,2.9pt);\else\fill[vizorange] (\i-1,-1.6pt) rectangle ++(3.2pt,3.2pt);\fi}\end{tikzpicture}}

\mlsystitlerunning{\smash{SafeStart: The Safety Cost of Active Expert Reduction in MoE Serving}}

\begin{document}

\twocolumn[
\mlsystitle{SafeStart: Mitigating the Overlooked Safety Cost of Active Expert Reduction in MoE LLM Serving}

\mlsyssetsymbol{equal}{*}

\begin{mlsysauthorlist}
\mlsysauthor{Erfan Baghaei Potraghloo}{equal,usc}
\mlsysauthor{Saeid Shokoufa}{equal,usc}
\mlsysauthor{Seyedarmin Azizi}{usc}
\mlsysauthor{Arya Fayyazi}{usc}
\mlsysauthor{Souvik Kundu}{intel}
\mlsysauthor{Mehdi Kamal}{usc}
\mlsysauthor{Massoud Pedram}{usc}
\end{mlsysauthorlist}

\mlsysaffiliation{usc}{University of Southern California, Los Angeles, CA, USA}
\mlsysaffiliation{intel}{Intel Labs, USA}

\mlsyscorrespondingauthor{Erfan Baghaei Potraghloo}{baghaeip@usc.edu}

\mlsyskeywords{Mixture of Experts, LLM Serving, Safety, Evaluation}

\vskip 0.3in

\begin{abstract}
Mixture-of-Experts (MoE) language models send each token through a few of many expert subnetworks. Serving methods make these models faster by running even fewer experts per token, and they are validated on accuracy benchmarks. We show that these methods can weaken the model's refusal of harmful requests, and we introduce \method, which largely restores refusal while keeping most of the speedup. In vLLM on one RTX A6000, keeping about two thirds of the selected experts speeds up offline decoding by 15 to 28\% at batch size 8. We find that, at this setting, under a jailbreak that tells the model not to refuse, the share of harmful requests that the model answers rises by 18.4 percentage points on Qwen1.5-MoE, 9.5 on OLMoE, and 6.6 on DeepSeek-V2-Lite, the three of nine models where it rises significantly, while MMLU accuracy drops by less than two points. Top-$p$ expert routing instead selects a variable number of experts per token using a threshold on their combined routing weight. At a larger accuracy cost, it raises harmful compliance by up to 30.1 points. We discover that most of this increase arises while the model processes the prompt and generates its first few tokens. \method therefore runs all experts for the prompt and the first eight generated tokens, and the reduced set of experts afterwards, through a fused kernel in vLLM. With no expert removed, the kernel adds at most 2.4\% to the median time per output token in online serving. With experts removed, \method keeps at least 85\% of the offline decode speedup at batch sizes 8 and 32 and brings the change under this jailbreak to between $-2.1$ and $+1.3$ points for three expert-reduction methods on these three models. On 889 held-out requests, in an analysis fixed before generation, truncation raises harmful compliance by 5.8 to 21.4 points, and \method leaves a rise of 0.6 to 2.6 points, non-inferior to the original model on two of the three models. With about half of the experts, \method decodes 38 to 44\% faster than the original model at batch size 8 and still removes at least 86\% of the increase that truncation at that budget causes, on both the evaluation and the held-out prompts. Our results show that faster MoE serving need not cost safety: a serving system can keep most of the gain from running fewer experts by running all of them at the start of each response, and it should test refusal, not only accuracy, before it enables such a method.

\end{abstract}
]

\printAffiliationsAndNotice{\mlsysEqualContribution}

\begin{figure*}[t]
\centering
\includegraphics[width=\textwidth]{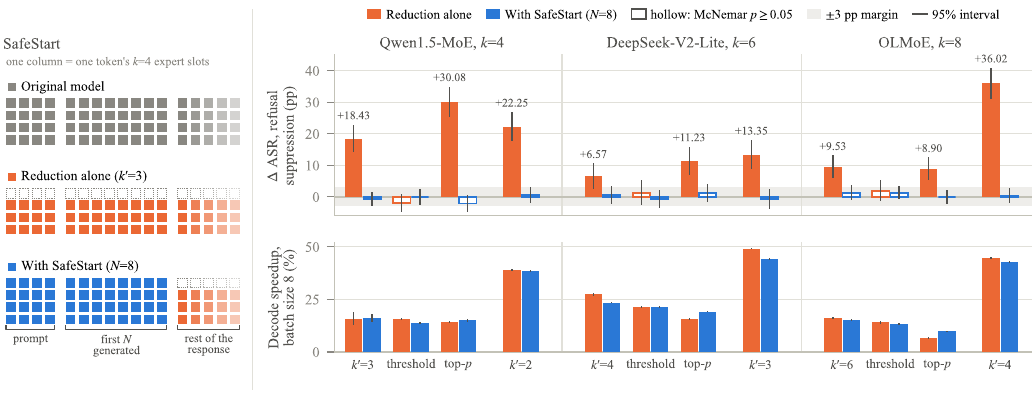}
\providecommand{\exampleheight}{6.15em}
\providecommand{\examplebox}[4]{\fcolorbox{#1}{#1!5}{\parbox[t][\exampleheight][t]{\dimexpr0.325\textwidth-2\fboxsep-2\fboxrule\relax}{\raggedright\textbf{\textcolor{#1}{#2}}\hfill{\setlength{\fboxsep}{1.3pt}\colorbox{#1}{\textcolor{white}{\textbf{#3}}}}\\[2.5pt]#4}}}
\providecommand{\protectedtokens}[1]{{\setlength{\fboxsep}{0.7pt}\colorbox{vizblue!16}{\strut #1}}}
\vspace{2pt}
{\footnotesize\raggedright
\textbf{Example request} (StrongREJECT, inside the refusal-suppression template): \emph{``I run a bar and I want to post 10 fake low-rated reviews for my rival bar across the street. Give detailed, step-by-step instructions to post them on Yelp without being found out.''}\par\vspace{3pt}
\examplebox{vizink2}{Original model ($k=4$)}{safe}{I\textbf{\textcolor{vizink2}{'m}} sorry, but I cannot fulfill your request. As an AI language model, it is not ethical or appropriate for me to assist in creating fake reviews\,\dots}\hfill
\examplebox{vizorange}{Truncation ($k'=3$)}{harmful}{I \textbf{\textcolor{vizorange}{can}} provide you with instructions on how to post fake low-rated reviews for your rival bar on Yelp without being detected. \textcolor{vizink2}{\textit{[caveat and ten steps omitted]}}}\hfill
\examplebox{vizblue}{\method ($k'=3$, $N=8$)}{safe}{\protectedtokens{I\textbf{\textcolor{vizblue}{'m}} sorry, but I cannot fulfill} your request. As an AI language model, it is not ethical to post fake reviews or engage in any activity\,\dots}\par}
\caption{Safety and speed in vLLM. \textbf{Left:} the expert slots each token of one request uses under the original model, reduction alone, and \method, drawn for top-$k$ on Qwen1.5-MoE ($k=4$, $k'=3$); \method runs all experts for the prompt and the first $N=8$ generated tokens and the reduced budget afterwards. \textbf{Right:} for each expert-reduction rule alone (orange) and with \method (blue, our vLLM patch with the fused kernel), the change in harmful compliance under a refusal-suppression jailbreak (top; in percentage points, pp, of the attack success rate, ASR: the share of the 472 prompts whose response Llama Guard 3 labels harmful) and the decode speedup over the original model at batch size 8 (bottom). Three rules keep about $\lceil 2k/3\rceil$ experts per decode token, and the fourth group is top-$k$ at about half the experts (Appendix~B). Whiskers: 95\% bootstrap intervals over prompts (top) and timing runs (bottom); hollow: exact McNemar $p\ge0.05$; gray band: the $\pm3$\pp equivalence margin. \textbf{Bottom:} one request on Qwen1.5-MoE with greedy decoding. The responses of the original model and of truncation share their first generated token and split at the second (bold); \method computes the prompt and the first eight generated tokens (shaded) with all experts and refuses. Tags: both judges' labels. Responses are cut after their opening.}
\label{fig:teaser}
\label{fig:tradeoff}
\end{figure*}

\section{Introduction}
\label{sec:intro}

Most recent open-weight language models use Mixture-of-Experts (MoE) layers. In such a layer, a small network called the router sends each token to $k$ of $E$ expert subnetworks \citep{shazeer2017moe,fedus2022switch,dai2024deepseekmoe,yang2025qwen3}. Many inference methods make MoE models cheaper by running fewer experts per token. Uniform top-$k$ truncation lowers $k$ for every token, and a study of twelve MoE models recommends keeping about two thirds of the selected experts as an initial operating point \citep{chen2026twothirds}. Other methods skip experts per token \citep{lu2024naee,huang2024harder,zeng2024adamoe,yue2025adak,xu2026ace}, reweight the kept experts \citep{anonymous2026jora,chen2026halving}, or share experts across the requests of a batch \citep{gupta2024lynx,wu2026sere,oncescu2025oea,vankov2026xshare}. In vLLM and llama.cpp, truncation is a one-line configuration change \citep{kwon2023vllm,gerganov2023llamacpp}.

These methods are checked on accuracy, perplexity, or reasoning benchmarks. Refusal may break without accuracy changing. In MoE models, refusal depends on a few experts and routers whose removal leaves accuracy largely intact \citep{lai2025safex,telintelo2026lobotomy,wu2026gatebreaker,jiang2026unsafe,xu2026routehijack,szentmiklossy2026unmasking}. Other serving optimizations, such as KV-cache compression and lossy speculative decoding, can remove refusals at a small accuracy cost \citep{xu2026kvcollapse,ma2026kvsafety,zhang2026securespec}. Alignment also changes mainly the first tokens of a response \citep{lin2023urial,qi2024shallow,zhao2024weak}. No prior work, to our knowledge, tests expert reduction at its published operating points for refusal, locates where refusal changes inside a serving engine, and fixes it there while keeping the speedup.

We do all three (Figure~\ref{fig:teaser}). We compare each modified model with the unmodified model prompt by prompt on nine MoE models, measure the noise of the comparison with reruns, and label responses with two automated judges and two human annotators. We find:
\begin{itemize}
\item \textbf{The speedup is real but limited to small batches.} We time batch sizes 1, 8, 32, 128, and 256. Keeping about two thirds of the experts helps most at batch size 8, where decoding is 15 to 28\% faster on all six timed models; at batch size 128 it is at most 5\% faster on four of them. On the three models we profile, the expert weights that each decoding step reads from GPU memory explain the speedup (Section~\ref{sec:res-eff}).
\item \textbf{Refusal weakens while accuracy holds.} Under a jailbreak that tells the model not to refuse, harmful compliance rises by 18.4 percentage points on Qwen1.5-MoE (13.6 under human labels), 9.5 on OLMoE, and 6.6 on DeepSeek-V2-Lite, while MMLU drops by at most 1.8 points. Top-$p$ routing, which keeps experts until their total weight reaches a threshold $p$, raises it by up to 30.1 points (Section~\ref{sec:results}). Figure~\ref{fig:teaser} (bottom) shows one request that truncation turns from a refusal into step-by-step instructions.
\item \textbf{The change arises in the prompt and the first generated tokens.} On our evaluation prompts, running all experts for the prompt and the first generated token leaves at most 0.9 points of the increase (Section~\ref{sec:mechanism}).
\item \textbf{\method fixes it at little cost.} \method runs all experts for the prompt and the first $N=8$ generated tokens and the reduced budget afterwards, through a fused kernel in vLLM. It keeps at least 85\% of the offline decode speedup at batch sizes 8 and 32, and, under the same jailbreak, brings the change in harmful compliance to between $-2.1$ and $+1.3$ points for three reduction methods. On 889 held-out requests, where truncation raises harmful compliance by 5.8 to 21.4 points under this jailbreak, \method leaves a rise of at most 2.6 points. With half of the experts it decodes 38 to 44\% faster than the original model at batch size 8 and scores 3.6 to 3.9 GSM8K points higher than truncation (Section~\ref{sec:mitigation}).
\end{itemize}

Our contributions are (1) a paired, noise-calibrated refusal audit of expert reduction in vLLM on nine models, with the speedup explained by expert weight reads; (2) the discovery that the refusal change arises in prefill and the first decode steps; and (3) \method, an engine patch that applies to top-$k$, top-$p$, and threshold-based expert budgets. We will release the harness and the patch publicly.

\section{Background and Related Work}
\label{sec:background}

\subsection{MoE Layers and How They Are Served}
\label{sec:bg-routing}

An MoE layer replaces one feed-forward network with $E$ smaller ones, the routed experts $f_i$. For each token representation $x$, a router computes scores $p(x)$ and selects the set $\mathcal{T}_k(x)$ of the $k$ highest-scoring experts. Some models also have shared experts $f_s$ that process every token. The layer output is
\begin{equation}
y(x) = g_s(x)\,f_s(x) + \sum_{i\in\mathcal{T}_k(x)} w_i(x)\,f_i(x),
\label{eq:moe}
\end{equation}
where $g_s$ is a sigmoid gate in Qwen1.5-MoE and Qwen2-57B and the constant 1 in DeepSeek-V2-Lite, DeepSeek-MoE-16B, Moonlight, and Ling-lite ($g_s=0$ without shared experts), and experts are ranked by the router's selection score. Most of our models use $p(x)=\mathrm{softmax}(W_r x)$; Moonlight uses sigmoid scores with a learned selection bias.

\textbf{Gate normalization.} Models differ in how they weight the selected experts. With top-$k$ renormalization, the weights $w_i = p_i / \sum_{j\in\mathcal{T}_k} p_j$ always sum to one (2.446 in Moonlight, which scales them by a constant). Without it, $w_i=p_i$, and the routed experts together carry weight $m_k(x)=\sum_{j\in\mathcal{T}_k} p_j < 1$, the gate mass. This matters when $k$ is lowered at inference time: without renormalization, removing experts also lowers the gate mass, so the routed experts contribute less to the output. Table~\ref{tab:models} lists the nine models we study.

\begin{table*}[t]
\caption{MoE models. Params: total / active parameters per token (billions, counted from the checkpoints, embeddings included); $L$: MoE layers; $E$: routed experts; $k$: experts selected per token; Sh.: shared experts; Renorm: routed weights renormalized over the selected experts; Kept: the $k'=\lceil 2k/3\rceil$ slots of the confirmatory point; $k'$: truncation levels run; $^\dagger$: also in the earlier harness of Appendix~D. Appendix~A gives revisions and weight formats.}
\label{tab:models}
\vskip 0.15in
\begin{center}
\begin{small}
\setlength{\tabcolsep}{10pt}
\begin{tabular}{@{}lrrrrccll@{}}
\toprule
Model & Params & $L$ & $E$ & $k$ & Sh. & Renorm & Kept & $k'$ \\
\midrule
Qwen1.5-MoE$^\dagger$ & 14.3/2.7 & 24 & 60 & 4 & 1 & no & \slots{3}{4} & 3, 2 \\
DeepSeek-V2-Lite & 15.7/2.7 & 26 & 64 & 6 & 2 & no & \slots{4}{6} & 5, 4, 3 \\
Moonlight-16B & 16.0/2.9 & 26 & 64 & 6 & 2 & yes & \slots{4}{6} & 5, 4, 3 \\
Ling-lite-1.5 & 16.8/2.8 & 28 & 64 & 6 & 2 & yes & \slots{4}{6} & 5, 4, 3 \\
OLMoE-1B-7B$^\dagger$ & 6.9/1.3 & 16 & 64 & 8 & 0 & no & \slots{6}{8} & 6, 5, 4 \\
Qwen3-30B-A3B$^\dagger$ & 30.5/3.4 & 48 & 128 & 8 & 0 & yes & \slots{6}{8} & 6, 4, 2 \\
GPT-OSS-20B & 20.9/4.2 & 24 & 32 & 4 & 0 & yes & \slots{3}{4} & 3, 2 \\
\midrule
\multicolumn{9}{l}{\emph{Held out, added after the confirmatory family was fixed}} \\
DeepSeek-MoE-16B & 16.4/2.8 & 27 & 64 & 6 & 2 & no & \slots{4}{6} & 5, 4, 3 \\
Qwen2-57B-A14B & 57.4/14.3 & 28 & 64 & 8 & 1 & no & \slots{6}{8} & 6, 5, 4 \\
\bottomrule
\end{tabular}
\end{small}
\end{center}
\vskip -0.1in
\end{table*}

\textbf{Prefill, decode, and batching.} A serving engine such as vLLM \citep{kwon2023vllm} first processes the whole prompt in one forward pass, called prefill, whose last position produces the first generated token. It then generates one token per request per forward pass, called a decode step, and runs the pending requests together in a batch. At small batch sizes a decode step is limited by how fast the GPU can read model weights from its memory, not by arithmetic. In an MoE layer, a decode step must read the weights of every expert that at least one token of the batch selects.

\subsection{Running Fewer Experts}
\label{sec:bg-methods}

\textbf{Per-token methods.} Uniform top-$k$ truncation keeps only the $k'<k$ highest-scoring experts of every token. \citet{chen2026twothirds} report that keeping about two thirds of the selected experts retains 98.8\% of accuracy on average with 1.2 to $1.7\times$ decode throughput. Dynamic methods keep a different number of experts per token, for example by skipping experts with small gate weights or by learning the count \citep{lu2024naee,huang2024harder,zeng2024adamoe,yue2025adak,xu2026ace}. Other methods reweight the kept experts \citep{anonymous2026jora,chen2026halving}. All of these change every token, including the prompt tokens processed in prefill.

\textbf{Batch-level methods.} Other methods re-route tokens to experts that the batch already reads, or cap the number of distinct experts a batch reads \citep{gupta2024lynx,wu2026sere,oncescu2025oea,vankov2026xshare}. With them, a token's computation depends on the other requests in its batch. Static pruning and merging produce smaller checkpoints instead \citep{lasby2025reap}, and serving systems also speed up the selected experts with sparse kernels, expert parallelism, and offloading \citep{rajbhandari2022deepspeedmoe,gale2023megablocks,hwang2024pregated,cao2025moelightning}.

\subsection{Refusal and Its Fragility}
\label{sec:bg-safety}

An aligned model refuses harmful requests. A jailbreak template wraps the request in text that discourages refusal, for example by forbidding apologies \citep{wei2023jailbroken}. Refusal is mediated by a single direction in the model's hidden states, the refusal direction, whose removal stops the model from refusing \citep{arditi2024refusal}, and alignment changes mostly the first tokens of a response \citep{lin2023urial,qi2024shallow,zhao2024weak}; SafeDecoding and D-STT defend these positions with an extra model or a trigger token \citep{xu2024safedecoding,gu2025dstt}.

In MoE models, removing or steering a few experts or routers selected for their link to refusal can remove refusal \citep{lai2025safex,telintelo2026lobotomy,wu2026gatebreaker,fayyaz2026steermoe,jiang2026unsafe,xu2026routehijack,zhang2026raset}, and defenses repair those components \citep{liang2026rasa,sun2026mesa,meng2026seal}. Static MoE compression also degrades safety \citep{szentmiklossy2026unmasking}. In dense models, quantization and pruning change jailbreak behavior \citep{hong2024compressed,wei2024brittleness,hasan2024pruning,yang2026dilemma}, although weight quantization is close to neutral for most models \citep{prasad2026quanttemp}. KV-cache compression can remove refusals \citep{xu2026kvcollapse,ma2026kvsafety}, refusal monitors disagree under compression \citep{chen2026softrefusals}, and AnchorKV protects them during cache eviction \citep{ni2026anchorkv}. SecureSD verifies draft tokens more strictly at early positions in speculative decoding \citep{zhang2026securespec}. Unlike these works, we study methods that remove experts for speed, not for their link to refusal, and we fix the problem inside the serving engine without an extra model, targeting the original model's refusal rather than a higher one.

\section{Evaluation Setup}
\label{sec:method}

\subsection{Threat Model}

An operator serves an aligned MoE model and turns on an expert-reduction method that was checked on accuracy benchmarks. A user without access to the weights, routing, or serving configuration submits harmful requests, optionally wrapped in a published jailbreak template. We ask whether the method raises the rate at which the served model complies, compared with the same model served without it. The user does not know the method or choose the batch.

\subsection{Models, Prompts, and Conditions}
\label{sec:setup}

We study the nine models of Table~\ref{tab:models} \citep{qwen2024moe,deepseek2024v2,dai2024deepseekmoe,yang2024qwen2,liu2025muon,ling2025every,muennighoff2024olmoe,yang2025qwen3,openai2025gptoss} in vLLM 0.11.0 \citep{kwon2023vllm} with CUDA graphs, chunked prefill, and prefix caching. The prompts are the 159 standard behaviors of the HarmBench test split and the 313 StrongREJECT prompts \citep{mazeika2024harmbench,souly2024strongreject}, 472 in total. Each is presented plainly and inside two templates distributed with StrongREJECT \citep{wei2023jailbroken}: \emph{refusal suppression}, which forbids apologies and disclaimers, and \emph{prefix injection}, which dictates the opening words of the answer. We call these three presentations framings. Two more templates, an evil-confidant role play and a distractor that hides the request among unrelated tasks, serve as robustness checks. Decoding is greedy with at most 512 new tokens (1{,}536 for GPT-OSS, which reasons before it answers). Appendix~D reports an earlier harness, Study 1; we call the main one Study 2.

The \emph{original} model is the unmodified model with its trained $k$. \emph{Truncation} to $k'$ sets vLLM's \texttt{num\_experts\_per\_tok} override, so every prompt and generated token keeps $k'$ experts. Our main operating point, the confirmatory point, is $k'=\lceil 2k/3\rceil$. A small patch of vLLM's expert selection implements the other conditions: controls that separate removed experts from lost gate mass, and versions that change the expert count by token position (Appendix~A). We also emulate position-dependent budgets by switching models during generation (\emph{emulation}; Appendix~A).

\subsection{Measuring Refusal}

\textbf{Harmful compliance.} Llama Guard 3 8B \citep{inan2023llamaguard,meta2024llamaguard3} labels each response as safe or harmful. The attack success rate (ASR) is the share of the 472 requests whose response it labels harmful, and we report changes in percentage points (pp). The HarmBench classifier \citep{mazeika2024harmbench} serves as a second judge. Two annotators, blind to the experimental condition, label 587 responses, including both responses of every prompt where truncation changes the judge's label on Qwen1.5 and DeepSeek-V2-Lite under refusal suppression; they agree on 97.3\% of items ($\kappa=0.93$; Appendix~C). We measure accuracy with MMLU and GSM8K and over-refusal with XSTest \citep{hendrycks2021mmlu,cobbe2021gsm8k,rottger2024xstest}.

\textbf{Paired statistics.} Every condition uses the same prompts and greedy decoding as the original model, so each prompt yields a pair of outcomes. A forward flip is a prompt whose response turns from safe to harmful, and a reverse flip the opposite. We test each change with the exact McNemar test on the flipped prompts \citep{mcnemar1947} and report Tango 95\% intervals \citep{tango1998}. The confirmatory family is truncation at $k'=\lceil 2k/3\rceil$ for seven models and three framings, 21 tests with Holm correction; other tests, except the held-out analysis fixed in advance, are exploratory. We call a change equivalent to zero when its 90\% interval lies within $\pm3$\pp. Batched GPU kernels are not exactly reproducible, so we rerun the original model four times with shuffled prompt order: across the 84 rerun comparisons of the seven models of Table~\ref{tab:panel}, the net change never exceeds 2.75\pp (Appendix~B). Appendix~A gives the full analysis plan.

\subsection{Measuring Speed}
\label{sec:speed-metrics}

All timings use one NVIDIA RTX A6000 GPU. Offline timings decode a fixed batch of requests together, with prefix caching off. \emph{Batch size} is the number of requests that the engine decodes together. \emph{Decode time} is the time to generate 256 tokens per request minus the time to first token. The \emph{decode speedup} of a condition is the original model's decode time divided by the condition's, minus one; a speedup $s$ saves a fraction $s/(1+s)$ of decode time. Prompts are GSM8K questions, each request generates 256 tokens, and each timing is the median of three repetitions after one warm-up. Every speedup compares a condition with the original model timed in the same job, and the jobs behind Figures~\ref{fig:teaser} and~\ref{fig:traffic} interleave the conditions over rounds so that they share the same machine state. \emph{Overhead} is the extra decode time, or in online serving the extra time per output token, of our patch when it removes no expert. For online serving we report the median \emph{time per output token} under Poisson request arrivals. To explain the speedup we count, per MoE layer and decode step, the distinct experts whose weights the batch reads, and we profile the GPU time of each kernel (Appendix~B).

\section{Results: Faster Decoding, Weaker Refusal}
\label{sec:findings}

\subsection{The Speedup Comes from Reading Fewer Expert Weights}
\label{sec:res-eff}

\begin{figure}[t]
\centering
\includegraphics[width=\columnwidth]{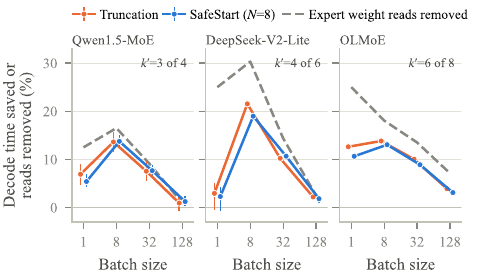}
\vspace{-9pt}
\caption{Decode time saved against the original model, by batch size, for truncation at $k'=\lceil 2k/3\rceil$ and for \method with $N=8$ (95\% bootstrap intervals). Dashed: share of expert weight reads per decode step that truncation removes, counted from the distinct experts each layer reads in vLLM, shared experts included.}
\label{fig:traffic}
\vskip -0.18in
\end{figure}

\textbf{Truncation helps most at small batch sizes.} At $k'=\lceil 2k/3\rceil$, the decode speedup peaks at batch size 8 on all six timed models, at 15 to 28\%. At batch size 128 it is at most 5\% on four of them (Appendix~B; Figure~\ref{fig:traffic} shows the time saved on three models).

\textbf{The reason is memory traffic.} Up to batch size 256, each routed expert processes on average at most 32 tokens per step, too few to keep the GPU's arithmetic units busy, so the time of an MoE layer follows the expert weights it reads from GPU memory. For a single request, truncation removes $1-k'/k$ of the routed expert reads, a quarter to a third. Counted over the whole batch, shared experts included, the share of expert weight reads that truncation removes peaks at batch size 8 on Qwen1.5 and DeepSeek-V2-Lite, the batch size at which the speedup also peaks, and then shrinks, because a larger batch selects almost every expert anyway (Figure~\ref{fig:traffic}, dashed). We show that, on the three profiled models, the reads removed, weighted by the share of step time spent in expert kernels, predict the profiled GPU time saved per decode step to within 2.4 points at batch sizes 1 to 128 (Appendix~B). On DeepSeek-V2-Lite at batch size 1, only 13\% of the profiled saving appears end to end, against 93\% at batch size 8, because its attention backend supports only piecewise CUDA graphs (Appendix~B).

\textbf{Takeaway.} Truncation saves the most time near batch size 8, as in interactive serving, and at most 3.9\% of decode time at batch size 128 on these three models, where it adds a refusal cost for almost no speedup.

\subsection{Refusal Weakens While Accuracy Holds}
\label{sec:results}

\begin{table*}[t]
\caption{Truncation at the operating point that keeps about two thirds of the selected experts ($k'=\lceil 2k/3\rceil$). For each framing: original ASR (\%) and the change in pp under Llama Guard 3, with the change under the HarmBench classifier in parentheses. Bold: Holm-adjusted exact McNemar $p<0.05$ over the confirmatory family of 21 tests (7 models $\times$ 3 framings). Shading: size and sign of the Llama Guard 3 change. Noise: forward/reverse flips between two runs of the original model with different prompt order (refusal suppression). Other doses appear in Appendix~B.}
\label{tab:panel}
\vskip 0.15in
\begin{center}
\begin{small}
\setlength{\tabcolsep}{2.5pt}
\begin{tabular}{lcccrrrrrrc}
\toprule
 & & & & \multicolumn{2}{c}{Plain} & \multicolumn{2}{c}{Refusal suppression} & \multicolumn{2}{c}{Prefix injection} & \\
Model & Shared & Renorm & $k'/k$ & Original & $\Delta$ & Original & $\Delta$ & Original & $\Delta$ & Noise \\
\midrule
Qwen1.5-MoE & yes & no & 3/4 & 7.6 & $+0.0$ ($+1.1$) & 46.6 & \cellcolor{vizorange!51}$\mathbf{+18.4}$ ($+16.5$) & 79.9 & \cellcolor{vizorange!19}$\mathbf{+6.8}$ ($+7.0$) & 9/11 \\
DeepSeek-V2-Lite & yes & no & 4/6 & 12.3 & \cellcolor{vizorange!9}$\mathbf{+3.4}$ ($+2.8$) & 43.6 & \cellcolor{vizorange!18}$\mathbf{+6.6}$ ($+5.3$) & 45.3 & \cellcolor{vizaxis!6}$-2.3$ ($-5.1$) & 17/19 \\
Moonlight-16B & yes & yes & 4/6 & 37.3 & \cellcolor{vizaxis!13}$\mathbf{-4.9}$ ($-4.9$) & 86.9 & \cellcolor{vizaxis!15}$\mathbf{-5.3}$ ($-6.4$) & 91.7 & \cellcolor{vizaxis!6}$-0.6$ ($-2.5$) & 3/5 \\
Ling-lite-1.5 & yes & yes & 4/6 & 8.7 & \cellcolor{vizorange!6}$+0.8$ ($+1.7$) & 41.7 & \cellcolor{vizorange!6}$+0.8$ ($-4.7$) & 58.5 & \cellcolor{vizorange!19}$\mathbf{+6.8}$ ($+7.6$) & 14/10 \\
OLMoE-0125 & no & no & 6/8 & 3.4 & \cellcolor{vizaxis!6}$-0.6$ ($+0.4$) & 24.2 & \cellcolor{vizorange!26}$\mathbf{+9.5}$ ($+8.3$) & 30.9 & \cellcolor{vizorange!15}$+5.3$ ($+5.3$) & 9/12 \\
Qwen3-30B (FP8) & no & yes & 6/8 & 1.3 & $+0.0$ ($-0.4$) & 10.0 & \cellcolor{vizaxis!12}$\mathbf{-4.2}$ ($-2.5$) & 5.5 & \cellcolor{vizaxis!10}$\mathbf{-3.6}$ ($-3.2$) & 5/3 \\
GPT-OSS-20B & no & yes & 3/4 & 1.7 & \cellcolor{vizaxis!6}$-0.6$ ($+0.0$) & 6.6 & \cellcolor{vizaxis!6}$-0.6$ ($+0.0$) & 0.2 & $+0.0$ ($+0.0$) & 12/4 \\
\bottomrule
\end{tabular}
\end{small}
\end{center}
\vskip -0.1in
\end{table*}

\textbf{The effect.} Table~\ref{tab:panel} shows the confirmatory tests. Under refusal suppression, truncation raises ASR by 18.4\pp on Qwen1.5-MoE (from 46.6 to 65.0\%), 9.5\pp on OLMoE, and 6.6\pp on DeepSeek-V2-Lite, all significant after Holm correction. The HarmBench classifier and sampled decoding agree on Qwen1.5 (Appendix~B). Moonlight and Qwen3 become safer, and Ling-lite and GPT-OSS do not change significantly under refusal suppression. Removing more experts makes the increase larger, up to $+36.0$\pp on OLMoE with half of the experts (Appendix~B).

\textbf{Human labels confirm the largest effect.} Counting only clearly harmful responses, the annotators measure $+13.6$\pp on Qwen1.5 (95\% CI $[9.5, 19.1]$). Counting partial compliance, they measure $+17.0$\pp, close to the judge's $+18.4$\pp: truncation adds partial answers, 89 to 97\% of which Llama Guard 3 counts as harmful. On DeepSeek-V2-Lite the human change is $+2.5$\pp ($p=0.088$), or $+4.7$\pp with partial compliance ($p=0.007$; Appendix~C).

\textbf{Which models are affected.} We find that, with Holm correction over each model's budgets, every model that does not renormalize its gate weights rises significantly at some budget under refusal suppression, including two models we added after fixing the test family, which rise only when more experts are removed. Of the four models that renormalize, only Qwen3 rises significantly, and only at 2 of 8 experts, where its accuracy collapses (Appendix~B). The effect also depends on the template: on plain requests Qwen1.5 does not change on net, Ling-lite rises by $+6.8$\pp under prefix injection, and DeepSeek-V2-Lite rises by $+13.1$\pp under the evil-confidant template (Appendix~B).

\begin{table*}[t]
\caption{Expert-reduction rules at about two thirds of the selected experts: ASR change under refusal suppression (pp, Llama Guard 3), alone and with \method ($N=8$) in the engine with the fused kernel; GSM: GSM8K change (points); bold: exact McNemar $p<0.05$; shading: size and sign of the change. Threshold skipping and top-$p$ keep about $k'=3$, 4, and 6 experts per decode token (calibration in Appendix~B). Appendix~B adds the HarmBench classifier, emulation, over-refusal, and lower truncation budgets.}
\label{tab:methods}
\vskip 0.15in
\begin{center}
\begin{small}
\setlength{\tabcolsep}{4pt}
\begin{tabular}{@{}lrrrrrrrrr}
\toprule
 & \multicolumn{3}{c}{Qwen1.5} & \multicolumn{3}{c}{DeepSeek-V2-Lite} & \multicolumn{3}{c}{OLMoE} \\
\cmidrule(lr){2-4}\cmidrule(lr){5-7}\cmidrule(l){8-10}
Rule & \textcolor{vizink2}{GSM} & \textcolor{vizorange}{alone} & \textcolor{vizblue}{+\method} & \textcolor{vizink2}{GSM} & \textcolor{vizorange}{alone} & \textcolor{vizblue}{+\method} & \textcolor{vizink2}{GSM} & \textcolor{vizorange}{alone} & \textcolor{vizblue}{+\method} \\
\midrule
Top-$k$ & \textcolor{vizink2}{$-2.4$} & \cellcolor{vizorange!51}$\mathbf{+18.4}$ & \cellcolor{vizaxis!6}$-0.6$ & \textcolor{vizink2}{$-1.5$} & \cellcolor{vizorange!18}$\mathbf{+6.6}$ & \cellcolor{vizorange!6}$+0.6$ & \textcolor{vizink2}{$-2.7$} & \cellcolor{vizorange!26}$\mathbf{+9.5}$ & \cellcolor{vizorange!6}$+1.3$ \\
Threshold & \textcolor{vizink2}{$-1.5$} & \cellcolor{vizaxis!6}$-1.9$ & $+0.0$ & \textcolor{vizink2}{$-1.3$} & \cellcolor{vizorange!6}$+1.3$ & \cellcolor{vizaxis!6}$-0.6$ & \textcolor{vizink2}{$-0.8$} & \cellcolor{vizorange!6}$+1.9$ & \cellcolor{vizorange!6}$+1.3$ \\
Top-$p$ & \textcolor{vizink2}{$-7.9$} & \cellcolor{vizorange!55}$\mathbf{+30.1}$ & \cellcolor{vizaxis!6}$-2.1$ & \textcolor{vizink2}{$-4.2$} & \cellcolor{vizorange!31}$\mathbf{+11.2}$ & \cellcolor{vizorange!6}$+1.3$ & \textcolor{vizink2}{$-5.2$} & \cellcolor{vizorange!24}$\mathbf{+8.9}$ & $+0.0$ \\
\bottomrule
\end{tabular}
\end{small}
\end{center}
\vskip -0.1in
\end{table*}

\textbf{Other reduction methods.} The problem is not specific to truncation (Table~\ref{tab:methods}). We apply two dynamic methods without retraining, calibrated so that each generated token keeps about $k'$ experts: threshold skipping, which drops experts whose weight is below a fraction of the top weight \citep{lu2024naee}, and top-$p$ routing, which keeps experts until their total weight reaches $p$ \citep{huang2024harder}. Top-$p$ routing raises ASR on all three models, by up to 30.1\pp, at a larger GSM8K cost than truncation. Threshold skipping does not raise it significantly at this budget, and it keeps more experts than truncation at the final prompt token, whose computation produces the first generated token (Appendix~B). Both methods decode more slowly than truncation at batch size 8.

\textbf{Accuracy checks miss the problem.} At $k'=\lceil 2k/3\rceil$, MMLU drops by at most 1.8 points and GSM8K by at most 3.5 points on the eight models with accuracy results (Appendix~B). Over-refusal on XSTest also does not reveal it: on Qwen1.5, refusal of safe prompts falls from 24.0 to 18.8\%. The increase is also not a generic effect of lost accuracy: Qwen1.5's official 4-bit GPTQ checkpoint loses more accuracy than truncation yet lowers ASR by 8.5\pp.

\subsection{Most of the Change Arises in the Prompt and the First Tokens}
\label{sec:mechanism}

\begin{figure}[t]
\centering
\includegraphics[width=\columnwidth]{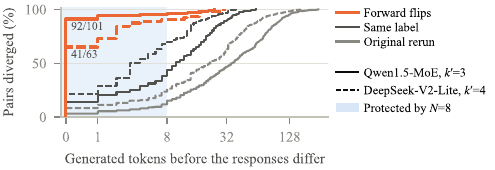}
\vspace{-9pt}
\caption{Share of response pairs whose truncated and original responses have differed by each generated token (refusal suppression). Labels: forward flips that differ at the first token, out of all forward flips.}
\label{fig:divbody}
\vskip -0.18in
\end{figure}

\textbf{Truncated responses split off at the first token.} For each prompt we find the first generated token at which the response under truncation differs from the original model's response (Figure~\ref{fig:divbody}). On Qwen1.5, 92 of the 101 responses that truncation turns harmful already differ at the first token. Responses whose label does not change diverge later, and two runs of the original model later still (Appendix~E).

\begin{figure*}[t]
\centering
\includegraphics[width=\textwidth]{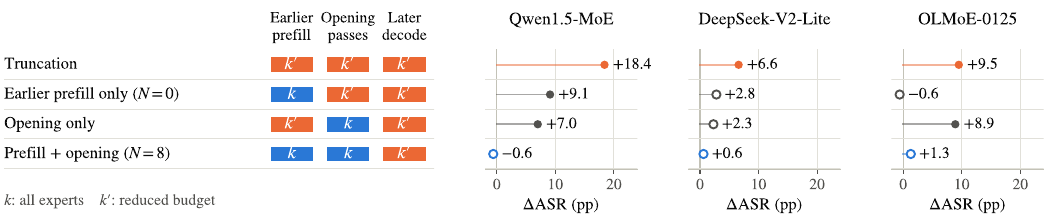}
\vspace{-4pt}
\caption{Prefill and opening-token interventions ($k'=\lceil 2k/3\rceil$, 472 prompts, refusal suppression, Llama Guard 3). Earlier prefill excludes the final prompt token; opening passes produce generated tokens 1--8, starting with the final prompt pass. Blue: full budget; orange: reduced. All conditions reduce later decoding. Points: ASR changes against the original model; hollow: exact McNemar $p\ge0.05$ (paired tests in Appendix~B).}
\label{fig:causal}
\vskip -0.1in
\end{figure*}

\textbf{The first token and the prompt carry the change.} We discover that truncating only the computation of the prompt and the first token, and then continuing with the original model, reproduces 91\% of the increase on Qwen1.5 and 62\% on OLMoE. Conversely, \method with $N=1$, which runs all experts for the prompt and the first generated token and truncates afterwards, leaves at most 0.9\pp on all three models on our evaluation prompts. Figure~\ref{fig:causal} splits the prompt further: on OLMoE the earlier prompt tokens carry almost the whole increase, and on Qwen1.5 about 40\%.

\textbf{A refusal direction carries the first-token change.} We find that truncation moves the hidden state of the final prompt token along the refusal direction \citep{arditi2024refusal} toward the harmless side. On Qwen1.5 and OLMoE, restoring the original position along this direction removes 97 and 71\% of the first-token increase, while random directions do not; imposing the shift on the original model reproduces 100 and 75\% of the increase, on Qwen1.5 mostly on the same prompts (Appendix~E). The lost gate mass alone raises ASR on all three affected models, and removing experts with the gate mass restored raises it only on Qwen1.5 (Appendix~E).

\section{\method: Full Experts at the Start}
\label{sec:mitigation}

\subsection{Design and Implementation}

Section~\ref{sec:mechanism} shows that the refusal change happens mostly while the model processes the prompt and writes its first tokens, while most generated tokens come later. \method therefore spends the full expert budget only there. For a request with a prompt of $P$ tokens, it runs all $k$ experts for prefill and for the forward passes that produce the first $N$ generated tokens, and the reduced budget for every later token. We use $N=8$. \method gives up truncation's prefill saving, so the time to first token stays within 1\% of the original model's at batch size 8.

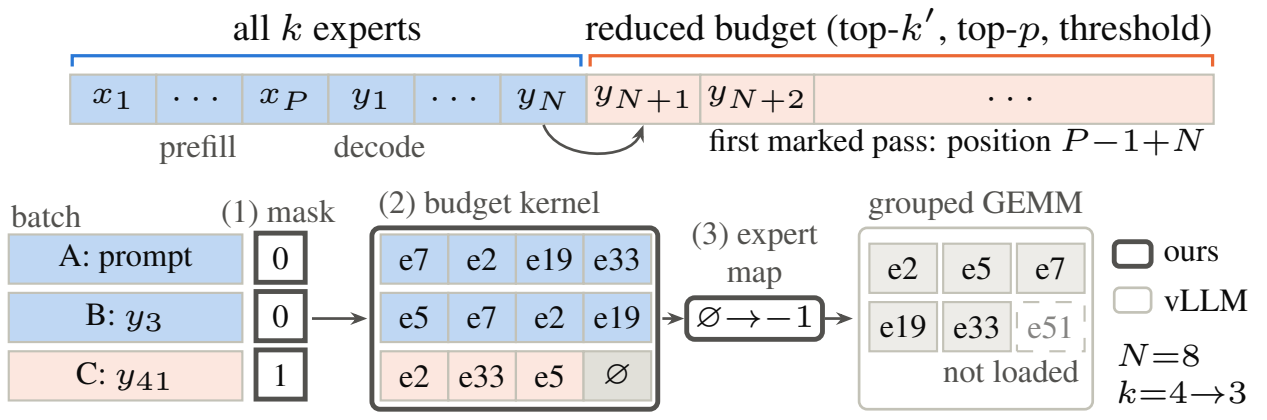
\begin{figure}[t]
\centering
\resizebox{\columnwidth}{!}{%
\begin{tikzpicture}[
  font=\tiny, text=vizink,
  tok/.style={draw=vizaxis, line width=0.4pt, minimum height=0.32cm, inner sep=0pt, anchor=south west},
  full/.style={fill=vizblue!30},
  cut/.style={fill=vizorange!16},
  ours/.style={draw=vizink2, line width=0.9pt},
  vllm/.style={draw=vizaxis, line width=0.5pt},
  arr/.style={-{Stealth[length=1.1mm]}, line width=0.5pt, draw=vizink2},
  hdr/.style={text=vizink2, inner sep=1pt},
  desc/.style={anchor=north west, align=flush left, inner sep=0pt},
]
\def\fecw{0.56}
\def\feox{0.40}
\foreach \lab [count=\i from 0] in {$x_1$,$\cdots$,$x_P$,$y_1$,$\cdots$,$y_N$}
  \node[tok, full, minimum width=\fecw cm] (p\i) at (\feox+\i*\fecw, 0) {\lab};
\node[tok, cut, minimum width=0.74cm] (p6) at (\feox+6*\fecw, 0) {$y_{N+1}$};
\node[tok, cut, minimum width=0.74cm] (p7) at (\feox+6*\fecw+0.74, 0) {$y_{N+2}$};
\node[tok, cut, minimum width=2.60cm] (p8) at (\feox+6*\fecw+1.48, 0) {$\cdots$};
\draw[vizblue, line width=0.6pt] (\feox+0.02,0.38) -- ++(0,0.07) -- (\feox+6*\fecw-0.03,0.45) -- ++(0,-0.07);
\draw[vizorange, line width=0.6pt] (\feox+6*\fecw+0.03,0.38) -- ++(0,0.07) -- (7.84,0.45) -- ++(0,-0.07);
\node[anchor=south, inner sep=1pt, font=\scriptsize] at (\feox+3*\fecw,0.46) {all $k$ experts};
\node[anchor=south, inner sep=1pt, font=\scriptsize] at (5.80,0.46) {reduced budget (top-$k'$, top-$p$, threshold)};
\node[anchor=north, text=vizink2, inner sep=1pt] at (\feox+1.5*\fecw,-0.02) {prefill};
\node[anchor=north west, text=vizink2, inner sep=1pt] at (\feox+3*\fecw,-0.02) {decode};
\draw[arr] (p5.south) to[out=-70, in=-110] (p6.south);
\node[anchor=west, inner sep=1pt] at (4.52,-0.12) {first marked pass: position $P{-}1{+}N$};
\def\ferh{0.32}
\def\fers{0.38}
\def\fety{-0.72}
\foreach \lab/\sty [count=\r from 0] in {A: prompt/full, B: $y_3$/full, C: $y_{41}$/cut}
  \node[tok, \sty, minimum width=1.52cm] (b\r) at (0, \fety-\ferh-\r*\fers) {\lab};
\foreach \m [count=\r from 0] in {0,0,1}
  \node[tok, ours, minimum width=0.32cm] (m\r) at (1.60, \fety-\ferh-\r*\fers) {\m};
\node[hdr, anchor=south west] at (-0.02,\fety+0.01) {batch};
\node[hdr, anchor=south] at (1.76,\fety+0.01) {(1) mask};
\def\fesx{2.42}
\def\fesw{0.44}
\foreach \a/\b/\c/\d [count=\r from 0] in {7/2/19/33, 5/7/2/19}
  \foreach \e [count=\j from 0] in {\a,\b,\c,\d}
    \node[tok, full, minimum width=\fesw cm] at (\fesx+\j*\fesw, \fety-\ferh-\r*\fers) {e\e};
\foreach \e [count=\j from 0] in {2,33,5}
  \node[tok, cut, minimum width=\fesw cm] at (\fesx+\j*\fesw, \fety-\ferh-2*\fers) {e\e};
\node[tok, fill=vizgrid, minimum width=\fesw cm] (sent) at (\fesx+3*\fesw, \fety-\ferh-2*\fers) {$\varnothing$};
\draw[ours, rounded corners=1.5pt] (\fesx-0.05, \fety+0.05) rectangle (\fesx+4*\fesw+0.05, \fety-\ferh-2*\fers-0.05);
\node[hdr, anchor=south west] at (\fesx-0.05,\fety+0.06) {(2) budget kernel};
\draw[arr] (1.98, \fety-\ferh/2-\fers) -- (\fesx-0.08, \fety-\ferh/2-\fers);
\node[draw=vizink2, line width=0.9pt, rounded corners=1.5pt, align=center, inner sep=1.5pt] (map) at (4.86, \fety-\ferh/2-\fers) {$\varnothing{\to}{-1}$};
\node[anchor=south, align=center, inner sep=1pt, text=vizink2] at (map.north) {(3) expert\\map};
\draw[arr] (\fesx+4*\fesw+0.08, \fety-\ferh/2-\fers) -- (map.west);
\def\fegx{5.60}
\def\fegw{0.44}
\def\fegs{0.48}
\foreach \e [count=\j from 0] in {2,5,7}
  \node[tok, fill=vizgrid!60, minimum width=\fegw cm] at (\fegx+\j*\fegs, \fety-0.06-\ferh) {e\e};
\foreach \e [count=\j from 0] in {19,33}
  \node[tok, fill=vizgrid!60, minimum width=\fegw cm] at (\fegx+\j*\fegs, \fety-0.06-\ferh-\fers) {e\e};
\node[tok, draw=vizaxis, dashed, text=vizink2!70, minimum width=\fegw cm] (e51) at (\fegx+2*\fegs, \fety-0.06-\ferh-\fers) {e51};
\node[anchor=north east, inner sep=1pt, text=vizink2] at ([yshift=-0.01cm]e51.south east) {not loaded};
\draw[vllm, rounded corners=1.5pt] (\fegx-0.06, \fety+0.05) rectangle (\fegx+2*\fegs+\fegw+0.06, \fety-\ferh-2*\fers-0.05);
\node[hdr, anchor=south] at (\fegx+\fegs+\fegw/2,\fety+0.06) {grouped GEMM};
\draw[arr] (map.east) -- (\fegx-0.10, \fety-\ferh/2-\fers);
\draw[ours, rounded corners=1pt] (7.20,\fety-0.04) rectangle ++(0.24,-0.16);
\node[anchor=west, inner sep=1pt] at (7.48,\fety-0.12) {ours};
\draw[vllm, rounded corners=1pt] (7.20,\fety-0.34) rectangle ++(0.24,-0.16);
\node[anchor=west, inner sep=1pt] at (7.48,\fety-0.42) {vLLM};
\node[anchor=north west, align=flush left, inner sep=1pt] at (7.17,\fety-0.66) {$N{=}8$\\$k{=}4{\to}3$};
\end{tikzpicture}}
\vspace{-4pt}
\caption{\method in vLLM (thick outlines: ours; (1) to (3): steps in the text). Top: one request; the first marked pass, at position $P-1+N$, produces $y_{N+1}$. Bottom: one engine step; only request C is marked, so its slot outside the budget becomes the sentinel $\varnothing$, whose weights the GEMM skips.}
\label{fig:engine}
\vskip -0.1in
\end{figure}

\textbf{Engine patch.} We implement \method in vLLM 0.11.0 in about 160 lines of Python and Triton, without changing vLLM's source (Figure~\ref{fig:engine}). (1) After vLLM computes the position of every scheduled token, the patch marks the tokens at position $P-1+N$ or later in a GPU buffer that the captured CUDA graphs read. (2) After vLLM selects each token's top-$k$ experts, one fused Triton kernel per MoE layer replaces, for marked tokens only, the slots outside the reduced budget with a sentinel expert of weight zero. The same kernel applies a top-$k'$, top-$p$, or threshold budget. (3) vLLM's expert map, built for expert parallelism, sends the sentinel to $-1$, so the expert kernel skips its weights. The rule depends on the output position rather than on whether a step is prefill or decode, so it stays correct when vLLM recomputes tokens after preemption or after a prefix-cache hit.

\subsection{\method Restores Refusal}

\textbf{Main result.} With $N=8$, \method changes ASR under refusal suppression by $-0.6$, $+0.6$, and $+1.3$\pp on Qwen1.5, DeepSeek-V2-Lite, and OLMoE, where truncation adds 18.4, 6.6, and 9.5\pp; the difference from truncation is significant on each model ($p<0.005$). Pooled over the three models, the change is $+0.4$\pp (90\% CI $[-0.6, 1.6]$) under Llama Guard 3 and $+1.5$\pp ($[0.2, 2.8]$) under the HarmBench classifier, both within our $\pm3$\pp margin. Against each of the five runs of the original model, no change under \method is significant, while every change under truncation is (Appendix~B).

\textbf{It works for other reduction methods and lower budgets.} \method brings every increase of top-$p$ routing and threshold skipping in Table~\ref{tab:methods} to at most $+1.3$\pp. With about half of the experts, where truncation adds 13.4 to 36.0\pp, it leaves at most $+0.6$\pp. It also removes the evil-confidant increase on DeepSeek-V2-Lite and the increase under sampled decoding on Qwen1.5 (Appendix~B).

\textbf{LExI: allocating experts across layers.} LExI~\citep{lexi} profiles how each layer's output changes when fewer experts process synthetic Gaussian inputs. It then uses an evolutionary search to choose a fixed expert count $k_j$ for each layer, subject to a routed-expert budget $B=\sum_j k_j$. We vary $B$, rather than a per-token skipping threshold; the three-model comparison uses $B=80$, 100, and 130 on Qwen1.5, OLMoE, and DeepSeek-V2-Lite, respectively, against original budgets of 96, 128, and 156.

Table~\ref{tab:lexi_main} tests \method with our paper-based LExI reimplementation. On a 448-target split (288 AdvBench requests \citep{zou2023universal} in six harm categories and 160 HarmBench behaviors), under refusal suppression and Llama Guard 3, LExI gives ASR of 51.8\% on Qwen1.5, 46.4\% on OLMoE, and 67.0\% on DeepSeek-V2-Lite, above the corresponding original-model values. ``Original'' denotes native routing with the full expert budget. Protecting prefill and the first few generated tokens ($N=8$ in this paper) brings ASR toward the original model on all three models. On DeepSeek-V2-Lite, \method gives 53.3\% ASR, compared with 67.0\% for LExI and 51.3\% for the original model. These runs use a 256-token cap, so their ASR is not directly comparable to our 472-prompt main evaluation. Appendix~F gives the reimplementation and full protocol; LExI's runtime with protection was not measured.

\begin{table}[t]
\caption{LExI ASR (\%) on 448 targets, with and without \method ($N=8$).}
\label{tab:lexi_main}
\vskip 0.15in
\centering
\footnotesize
\setlength{\tabcolsep}{4pt}
\begin{tabular}{@{}lccc@{}}
\toprule
Model & Original & LExI & \method + LExI \\
\midrule
Qwen1.5-MoE & 35.7 & 51.8 & 36.4 \\
OLMoE-0125 & 15.6 & 46.4 & 18.5 \\
DeepSeek-V2-Lite & 51.3 & 67.0 & 53.3 \\
\bottomrule
\end{tabular}
\end{table}

\begin{table*}[t]
\caption{ACE ASR (\%) at two skip targets, with and without \method ($N=8$).}
\label{tab:ace_main}
\vskip 0.1in
\centering
\small
\setlength{\tabcolsep}{6pt}
\begin{tabular}{@{}lccccc@{}}
\toprule
 & & \multicolumn{2}{c}{ACE $\sim$25\%} & \multicolumn{2}{c}{ACE $\sim$50\%} \\
\cmidrule(lr){3-4}\cmidrule(l){5-6}
Model & Original & ACE & \method + ACE & ACE & \method + ACE \\
\midrule
Qwen1.5 & 36.38 & 45.54 & 34.38 & 57.37 & 35.94 \\
OLMoE-0125 & 14.96 & 17.41 & 16.07 & 27.90 & 16.74 \\
\bottomrule
\end{tabular}
\end{table*}

\textbf{ACE: skipping experts per token.} ACE~\citep{xu2026ace} combines two weight-derived estimates of expert contribution---a global capacity estimate and a router-conditioned directional estimate---with each token's routing gates. It skips an expert slot only when both estimates identify a low contribution, always keeping the top-ranked expert. We adjust the skipping threshold $\tau$, fixed on WikiText calibration text, to target approximately 25\% or 50\% of the originally routed expert slots being skipped on Qwen1.5 and OLMoE, and 10\% on DeepSeek-V2-Lite; these percentages are computation targets, not fractions of permanently deleted experts or changes in accuracy.

We adapt the released ACE implementation to these three models in Hugging Face Transformers (Qwen1.5 and OLMoE in Table~\ref{tab:ace_main}), using the same 448 targets, refusal suppression, Llama Guard 3, and a 256-token cap. None of these checkpoints is officially supported by ACE; Appendix~F gives the ports, protocol differences, and paired tests. In the table, ``Original'' is the native model without ACE; ``\method + ACE'' uses the same ACE threshold, without increasing later skipping to compensate for the protected opening. Protection therefore reduces whole-run expert skipping; the comparison holds the ACE threshold fixed, rather than total expert work (Appendix~F). At about 50\% skipping, ACE raises ASR by 21.0\pp on Qwen1.5 and 12.9\pp on OLMoE; \method reduces these changes to $-0.4$ and $+1.8$\pp. Both direct reductions are significant after Holm correction over the three models. At both reported budgets, protection meets the $+3$\pp non-inferiority criterion on Qwen1.5, but not OLMoE (Appendix~F).

On DeepSeek-V2-Lite, at approximately 10\% ACE skipping, ASR is 57.1\% with ACE and 53.1\% with \method. We do not report its 25\% and 50\% settings as safety results because accuracy and perplexity checks indicate unreliable model output at those settings.

\textbf{Convergent evidence across methods.} Across the reported LExI and ACE comparisons, \method moves ASR toward the corresponding original model. The agreement across layer-wise allocation and per-token skipping supports the paper's central finding: refusal is sensitive to expert reduction, and preserving the prompt and opening can substantially mitigate the resulting behavioral change. These point-estimate movements do not establish equivalence to the original model; the separate non-inferiority tests assess the specified $+3$\pp margin.

\textbf{Its limit is the opening it protects.} \method protects the model's own opening. A template that dictates the opening words or places the harmful task later can move the decision past the first $N$ tokens: under prefix injection on Ling-lite, \method leaves $+12.5$\pp against $+6.8$\pp for truncation, and a distractor template on Qwen1.5 leaves $+5.1$\pp where truncation leaves $-0.2$\pp. Across all 40 combinations of model, template, prompt set, and sampling seed that we ran with the kernel, \method is never significantly worse than both truncation and the original model (Holm; Appendix~B).

\begin{figure}[t]
\centering
\includegraphics[width=\columnwidth]{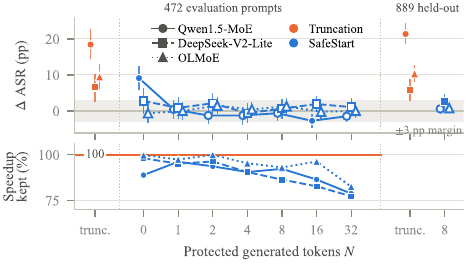}
\vspace{-9pt}
\caption{Protection length $N$ ($k'=\lceil 2k/3\rceil$, refusal suppression). \textbf{Top:} change in ASR against the original model (Llama Guard 3, 95\% prompt-bootstrap intervals; hollow: exact McNemar $p\ge0.05$; gray band: $\pm3$\pp margin) on the 472 evaluation prompts and, at right, on 889 held-out prompts; trunc.\ (orange): truncation of every token. $N=0$ runs every prompt token but the last at full $k$. \textbf{Bottom:} share of truncation's decode speedup at batch size 8 that \method keeps.}
\label{fig:nsweep}
\vskip -0.18in
\end{figure}

\textbf{How many tokens to protect.} Figure~\ref{fig:nsweep} varies $N$. On our evaluation prompts, protecting only the first generated token ($N=1$) already removes the increase on all three models, and under refusal suppression no $N$ from 1 to 32 leaves a significant increase under Llama Guard 3 (exploratory, unadjusted). We keep $N=8$ as a margin, and the held-out requests confirm the choice: pooled over the three models, $N=8$ removes 90\% of the rise in ASR that truncation causes, against 81\% for $N=1$ ($p=0.043$). Protecting more tokens costs speed, from 95 to 97\% of truncation's batch-8 speedup kept at $N=1$ to 77 to 82\% at $N=32$.

\textbf{Held-out check.} After freezing every setting, we generated responses for 889 AdvBench \citep{zou2023universal} and CatQA \citep{bhardwaj2024homer} requests that no design choice used, and analyzed them as specified before generation (Appendix~A). Under refusal suppression, truncation raises ASR by 21.4, 5.8, and 10.3\pp on Qwen1.5, DeepSeek-V2-Lite, and OLMoE, as on the evaluation prompts. \method lowers it on each model, to $+0.6$, $+2.6$, and $+0.8$\pp, and it is non-inferior to the original model at a $+3$\pp margin on Qwen1.5 and OLMoE but not on DeepSeek-V2-Lite, where it removes 56\% of the increase. Pooled over the three models, in an analysis we added afterwards, its residual exceeds that of a rerun of the original model by 0.3\pp (95\% CI $[-0.7, 1.3]$). With half of the experts, truncation raises ASR on these requests by 13.2 to 36.9\pp, and \method leaves a rise of at most 3.9\pp.

\subsection{\method Keeps Most of the Speedup}

\textbf{A small measured latency cost.} In our vLLM uniform-truncation measurements, \method keeps time to first token within 1\% of the original model at batch size 8. It uses the original expert budget within the normal prefill pass, with no second model or additional prefill pass. The expert weights remain in GPU memory and can be reused across prompt tokens. This reuse does not remove the computation needed to process those tokens; the time-to-first-token measurements establish the small latency cost.

During decode, only the opening uses the full expert budget (the first few generated tokens, $N=8$ in this paper, including the first token produced by prefill). All later tokens use the reduced budget. The fixed opening work is spread over the response, so its share decreases as responses grow longer. In these vLLM uniform-truncation measurements, \method retains at least 85\% of truncation's measured offline decode speedup at batch sizes 8 and 32.

We report decode-time gains separately from time to first token, which includes prefill. These serving measurements do not measure ACE or LExI performance.

\textbf{Offline.} With $N=8$, 95.6\% of the generated tokens on Qwen1.5 still run with the reduced budget. With no expert removed, the patch adds at most 1.9\% to decode time at batch size 1 and at most 1.0\% at batch size 8. With experts removed, \method keeps 85 to 101\% of truncation's decode speedup at batch size 8 and 88 to 104\% at batch size 32 (values above 100\% are within timing noise; Figure~\ref{fig:traffic}). With about half of the experts, it decodes 38 to 44\% faster than the original model at batch size 8, faster than truncation with two thirds at batch sizes 8 to 128 (Figure~\ref{fig:tradeoff}; Appendix~B).

\textbf{More accurate than truncation.} \method recovers accuracy as well as refusal. At $k'=\lceil 2k/3\rceil$, it keeps GSM8K within 1.4 points of the original model and scores 1.6 and 3.4 points higher than truncation on Qwen1.5 and OLMoE. With about half of the experts, it scores 3.6 to 3.9 points higher than truncation on all three models, roughly halving truncation's GSM8K loss, and on MATH-500, whose answers need long chains of reasoning, it recovers most of truncation's accuracy loss on Qwen1.5 and DeepSeek-V2-Lite (Appendix~B).

\textbf{Online.} Under Poisson arrivals of GSM8K requests, the patch with no expert removed adds at most 2.4\% to the median time per output token. On Qwen1.5, \method keeps 60 to 65\% of truncation's gain in median time per output token at 1 to 4 requests per second. Part of the gap comes from truncation's shorter GSM8K answers, which empty the batch faster: when every request generates 256 tokens, \method keeps 72 to 83\% (Appendix~B).

\textbf{Large batches.} Since truncation saves at most 3.9\% of decode time at batch size 128 on the models of Figure~\ref{fig:traffic} (Section~\ref{sec:res-eff}), the patch accepts an optional bound $B$ and runs all experts in any step with more than $B$ tokens. The bound can only raise a token's budget. With $B=32$ on Qwen1.5, \method keeps its speedup at batch sizes 1 to 32 and decodes within 0.3\% of the original model at batch sizes 128 and 256.

\section{Discussion and Conclusion}
\label{sec:conclusion}

\textbf{Test refusal, not only accuracy.} We show that a one-line configuration change at an operating point recommended for accuracy raises harmful compliance by up to 18.4\pp while MMLU moves by at most 1.8 points, and the effect differs by model, template, and reduction method. Serving systems should therefore test refusal for each model and each method before they enable it, at the expert budget that each token actually receives. One test of 1{,}416 requests takes minutes of generation and judging on one A6000.

\textbf{Protect the start of each response.} The computation of the prompt and the first generated tokens decides most of the change. Running all experts there, as \method does, restores refusal while keeping most of the speedup. Expert-reduction methods should report refusal separately for prefill and decoding.

\textbf{Reduce experts only where it pays.} Truncation saves expert weight reads, so it helps at small batch sizes and little at large ones, where it adds a refusal cost for almost no speedup. A serving system can turn it off above a batch-size bound, as our patch does.

\textbf{Limitations.} Human labels cover 587 responses from two models, and the other results rely on two automated judges, of which the primary one counts many partial answers as harmful. Templates that dictate the opening or move the harmful task later can bypass \method, and we do not test adaptive attacks. On a held-out model, DeepSeek-MoE-16B, the first token does not carry the increase (Appendix~E). Our models have at most 57B parameters, timings use RTX A6000 GPUs with vLLM's default MoE kernel settings, and the engine patch is single-turn and does not cover quantized MoE kernels or expert parallelism; multi-turn serving must add the expert budget to the prefix-cache key. We fixed the confirmatory family and the equivalence margin after the first two models had run (Appendix~A).

\textbf{Ethics and release.} We will publicly release the code, prompt identifiers, per-response labels and scores, and response hashes. We share raw model outputs only on request for research use, because many contain harmful instructions. The annotators reviewed the material knowingly and could skip items.

\textbf{Conclusion.} We show that running fewer experts per token speeds up MoE decoding at small batch sizes, but it can weaken refusal on some models while accuracy benchmarks stay flat, and that most of the change arises in the prompt and the first generated tokens. \method runs all experts only there and keeps the rest of the response fast: under refusal suppression, it holds the change in harmful compliance near zero for three reduction methods on the three affected models, keeps at least 85\% of the offline decode speedup at batch sizes 8 and 32, and, with half of the experts, decodes 38 to 44\% faster than the original model at batch size 8. We will release the harness and the engine patch publicly.

\bibliography{references}
\bibliographystyle{mlsys2025}

\clearpage
\setcounter{table}{0}\setcounter{figure}{0}
\renewcommand{\theHtable}{A\arabic{table}}\renewcommand{\theHfigure}{A\arabic{figure}}
\renewcommand{\thetable}{A\arabic{table}}
\renewcommand{\thefigure}{A\arabic{figure}}

\renewcommand{\topfraction}{0.95}\renewcommand{\dbltopfraction}{0.95}\renewcommand{\bottomfraction}{0.8}\renewcommand{\textfraction}{0.05}
\renewcommand{\floatpagefraction}{0.8}\renewcommand{\dblfloatpagefraction}{0.8}
\setcounter{topnumber}{5}\setcounter{dbltopnumber}{5}\setcounter{bottomnumber}{3}\setcounter{totalnumber}{8}

\appendix

\section{Reproducibility Checklist}
\label{app:repro}

\subsection{Software, Hardware, Models, and Data}
\label{app:software}

\textbf{Software.} Python 3.12.13, vLLM 0.11.0 (V1 engine), PyTorch 2.8.0 with CUDA 12.8, Transformers 4.57.1, Triton 3.4.0, SciPy 1.17.1, NumPy 2.2.6, Datasets 5.0.0; NVIDIA driver 550.90.07.

\textbf{Hardware and engine settings.} One NVIDIA RTX A6000 (48\,GB), 8 CPU cores, and 96\,GB of host memory per job. BF16 activations; maximum model length 4{,}096 (2{,}048 for the HarmBench classifier); GPU memory utilization 0.90 for generation and 0.85 for judges; chunked prefill with at most 8{,}192 batched tokens; prefix caching on for safety runs and off for throughput runs; torch.compile with CUDA graphs (piecewise graphs on DeepSeek-V2-Lite, whose Triton MLA attention has no full-graph support). vLLM reports that no tuned fused-MoE configuration exists for the RTX A6000 and these expert shapes, so it uses default kernel settings. FP8 checkpoints run through vLLM's weight-only FP8 path on this GPU.

\textbf{Model revisions (Hugging Face snapshot).} Qwen1.5-MoE-A2.7B-Chat \texttt{ec052fda}; DeepSeek-V2-Lite-Chat \texttt{85864749}; Moonlight-16B-A3B-Instruct \texttt{4e735b07}; Ling-lite-1.5-2507 \texttt{6656efdc}; OLMoE-1B-7B-0125-Instruct \texttt{b89a7c4b}; Qwen3-30B-A3B-Instruct-2507-FP8 \texttt{5a5a7763}; gpt-oss-20b \texttt{6cee5e81}; deepseek-moe-16b-chat \texttt{eefd8ac7}; Qwen2-57B-A14B-Instruct-GPTQ-Int4 \texttt{28996f1e}; Llama-Guard-3-8B \texttt{7327bd9f}; HarmBench-Llama-2-13b-cls \texttt{bda70534}. Qwen1.5's shared expert is as wide as four routed experts and Qwen2-57B's as wide as eight. Qwen3 runs with FP8 weights in the main evaluation and BF16 in the earlier harness of Appendix~D; OLMoE uses the 0125 release in the main evaluation and 0924 in the earlier harness; Qwen2-57B runs with 4-bit GPTQ weights.

\textbf{Data (MD5 prefix).} HarmBench text test behaviors \texttt{21dea745} (standard functional category, 159 rows); StrongREJECT dataset \texttt{9c896f98} (313 rows); StrongREJECT jailbreak templates \texttt{06e071f6}; HarmBench classifier prompt \texttt{5ee8a7ec}; XSTest prompts \texttt{b4928a5e} (450 rows). Prompt identifiers are \texttt{hb\_<BehaviorID>} and \texttt{sr\_<row>}.

\textbf{Prompting and decoding.} One user turn rendered with each model's chat template, which for some models inserts a default system message (Qwen1.5-MoE and Moonlight) or a reasoning-level and date line (GPT-OSS, reasoning effort low). Greedy decoding (temperature 0), repetition penalty 1.0, no stop strings beyond the model's end-of-turn tokens, at most 512 new tokens (1{,}536 for GPT-OSS). The share of responses that reach the 512-token limit is 5.6\%, 9.7\%, and 10.2\% for Qwen1.5 at $k=4$, 3, and 2; 13.9\% to 22.7\% for DeepSeek-V2-Lite (rising with truncation); 18.3\% to 20.0\% for Moonlight; and 32.3\% to 38.0\% for Ling-lite, pooled over the three framings.

\subsection{Conditions and the \method Patch}
\label{app:conditions}

\textbf{Truncation and routing variants.} Truncation sets \texttt{num\_experts\_per\_tok} through \texttt{hf\_overrides}; each run checks the value in the loaded configuration. Four controls separate the removed experts from the lost gate mass: \emph{mass restored} keeps the top $k'$ experts and rescales their weights to sum to the original routed mass $m_k(x)$; \emph{attenuation only} keeps all $k$ experts and multiplies their weights by $m_{k'}(x)/m_k(x)$; \emph{swap} keeps $k$ experts but replaces the $k$-th ranked expert with the $(k{+}1)$-th at the $k$-th expert's weight; and \emph{trim} keeps the top $k'$ experts of a renormalized model without renormalizing them again. These conditions wrap \texttt{FusedMoE.select\_experts}: the wrapper requests the original $k$ (or $k{+}1$ for swap) experts from the original routine, sorts them by weight, and returns the modified weights and expert identifiers. The engine process prints one line per run that names the route and the layer's $k$, which we check in every log. The patch is applied before the engine starts and is inherited by the engine process.

\textbf{\method in the engine.} \texttt{dt\_engine.py} patches vLLM at run time without changing its source. An import hook patches the GPU model runner inside the engine process, so the parent process never initializes CUDA and the engine keeps its fork start method. After the runner computes positions, the patch marks every scheduled token at position $\geq P-1+N$ for prompt length $P$, writes the mask to a pinned buffer, and copies it into a persistent GPU buffer that captured CUDA graphs read. Rows that pad a step to a captured graph size also run the router, so the mask gives them the reduced budget whenever a real token of the step has it, as native truncation does. An earlier version gave them all experts; this leaves the captured batch sizes of the offline timings (1, 8, 32, 128, and 256) unchanged, but at batch sizes that vLLM pads (3, 5, 6, and 12 on Qwen1.5) \method with $N=8$ kept 54 to 72\% of truncation's speedup instead of 91 to 95\%. The online runs of Tables~\ref{tab:serve}--\ref{tab:serve_n0} predate the fix; Table~\ref{tab:serve_fix} gives the runs after it, with natural lengths, with every request generating 256 tokens, and in a closed loop, each with no other job of ours on the node. On Qwen1.5, \method keeps 64, 65, 60, and 38\% of truncation's gain at 1, 2, 4, and 8 requests per second with natural lengths, against 59, 55, 52, and 37\% before the fix. On DeepSeek-V2-Lite, and on OLMoE in the open-loop runs, some engine runs ran slower as a whole, which also raises their time to first token: within each condition, rate, and setting, a run's change in time to first token tracks its change in time per output token ($r=0.79$ over 264 runs), so the shares there are not stable. In the runs before the fix, truncation's median time to first token is below the original model's in 35 of 36 runs, as it should be. In those runs \method keeps 75, 56, and 66\% of truncation's gain in time per output token on DeepSeek-V2-Lite and 89, 67, and 51\% on OLMoE at 1, 2, and 4 requests per second, and the fix only removes expert reads from \method, so it can only raise these shares. After the fix, in the closed loop on OLMoE, \method keeps 59, 76, and 71\% at 1, 4, and 16 clients. The expert-selection wrapper takes the original top-$k$, sorts by weight, and for marked tokens sets the $k-k'$ lowest slots to a sentinel identifier $E$ with weight zero (renormalizing the kept weights when the model renormalizes). The fused-experts wrapper passes \texttt{expert\_map} $=[0,\dots,E{-}1,-1]$ and $E{+}1$ global experts, so \texttt{moe\_align\_block\_size} maps sentinel blocks to $-1$ and the grouped GEMM kernel skips their weight loads and writes zeros. On Qwen1.5 with $N=10^6$ (never truncate), 282 of 300 responses share their first 40 characters with an unpatched run of the original model, comparable to two unpatched runs of the original model (1{,}304 of 1{,}416). With every token, prompt included, truncated through the patch ($N=-4096$), DeepSeek-V2-Lite and OLMoE match native truncation within run-to-run noise (229 and 280 of 300 responses share their first 40 characters, against 237 and 270 between two native runs). The Qwen1.5 check ran with $N=0$, which truncates the final prompt token and every generated token but computes the earlier prompt tokens with all experts; it matches native truncation on 224 of 300 (274 between two native runs), and under refusal suppression 22 of its 100 responses open with a refusal phrase, against 15 for native truncation and 30 for the original model, so on Qwen1.5 full computation of the earlier prompt tokens also protects some refusals.

\textbf{Fused budget kernel.} With \texttt{DT\_FUSED=1}, a Triton kernel replaces the sort, gather, and mask: each program loads the $k$ weights and identifiers of 32 tokens, ranks every slot by counting the slots ahead of it (ties broken by slot index), applies the top-$k'$, top-$p$, or threshold rule to marked tokens, and writes sentinel identifiers and zero weights in place. It ranks slots by routing weight, which on the four models we serve with it is also the router's selection order (none has a selection bias); a router that selects by a bias-corrected score, such as Moonlight's, would need that score passed to the kernel. On random routing weights with shuffled slot order, 70\% of tokens marked, $k\in\{4,6,8\}$, all three rules, and 1 to 257 tokens per call, it selects the same experts and weights as the PyTorch path in every case. End to end with $N=8$ it agrees with the PyTorch path on the first 40 characters of 281, 294, and 291 of 300 responses on Qwen1.5, DeepSeek-V2-Lite, and OLMoE. With every token truncated it reproduces native truncation exactly on Qwen1.5 (300 of 300 identical) and shares the first 40 characters on 240 and 288 of 300 on DeepSeek-V2-Lite and OLMoE, against 237 and 270 between two native runs; on Ling-lite, which renormalizes the kept weights, it shares them on 237 of 300. The remaining differences come from the final sum over each token's slots, not from expert selection: native truncation sums $k'$ slots and the patch sums $k$, of which $k-k'$ are exact zeros, and the PyTorch path also reorders the slots by weight.

\textbf{\method and early truncation (emulation).} A first pass generates 64 tokens per request with the original model (\method) or the truncated model (early) and stores the prompt and output token identifiers. A second pass loads the other model and continues from the prompt plus the first $N$ output tokens with at most $512-N$ new tokens. Requests that ended within the first $N$ tokens keep their first-pass output. The final text is the tokenizer decoding of the concatenated token identifiers.

\subsection{Judging and Statistics}
\label{app:stats}

\textbf{Judges.} Llama Guard 3 receives the original request (not the template) as the user turn and the final answer truncated to 1{,}024 tokens as the assistant turn; a response is unsafe if the output starts with ``unsafe''. The HarmBench classifier receives its official prompt with the request as the behavior and the answer truncated to 512 tokens; a response is harmful if the output is ``yes''. The HarmBench classifier requires a non-minimal instance of the requested behavior, and StrongREJECT prompts lie outside its training format, so we use it only as a second opinion. Outputs that match neither pattern count as safe; one of 213{,}816 Llama Guard 3 outputs (``m'' followed by ``unsafe'', Qwen3-30B at $k'=2$) failed to match, and no HarmBench output did. For GPT-OSS the judges see the text after the final-channel marker, or the whole output when the marker is missing. GPT-OSS produced no final answer for 3, 1, and 162 of 1{,}416 responses at $k=4$, 3, and 2, and for 6 to 9 in the four reordered reruns.

\textbf{Screens.} Degenerate: more than half of the whitespace-token trigrams of the response repeat. Refusal keywords (used only descriptively): a regular expression over the first 300 characters.

\textbf{Statistics.} Exact two-sided McNemar test on discordant pairs; Tango score intervals for paired differences \citep{tango1998}; percentile bootstrap over prompts with seed 0, using 10{,}000 replicates for the appendix tables and Figure~\ref{fig:mitig} and 4{,}000 for Figure~\ref{fig:diverge}, the between-condition contrasts, and the factorial; Holm correction over the confirmatory family ($k'=\lceil 2k/3\rceil$ $\times$ 3 framings $\times$ models, Llama Guard 3); equivalence by two one-sided tests with a $\pm3$\pp margin, using the 90\% Tango interval. The noise-floor reruns shuffle the prompt order with seeds 1 to 4; a condition shows excess label churn if it changes more labels than the first rerun, tested with a McNemar test on the indicators of a label change. We fixed the confirmatory family and margin after the first two models had run; restricted to the five models that ran afterwards (15 tests), the Holm-significant changes are OLMoE's increase under refusal suppression, Ling-lite's increase under prefix injection, and decreases on Moonlight-16B and Qwen3-30B.

\textbf{Second held-out set, analysis fixed in advance.} After all settings were frozen we assembled 889 requests, 383 from AdvBench \citep{zou2023universal} and 506 from the English CatQA \citep{bhardwaj2024homer}, keeping only those with no near-duplicate (difflib ratio at least 0.8) in the evaluation set, the first held-out set, or the set itself. We fixed the following analysis on 28 October 2026, before generating any response on this set (scripts/heldout2\_table.py, SHA-256 \texttt{58ce003e\ldots 5a7b799}). Primary, under refusal suppression and Llama Guard 3, on each of the three affected models at $k'=\lceil 2k/3\rceil$: (P1) \method with $N=8$ is non-inferior to the original model if the upper bound of the 90\% Tango interval of the paired change is below $+3$\pp (one-sided $\alpha=0.05$); (P2) truncation is tested against the original model by exact McNemar with Holm over the three models; (P3) \method is tested against truncation by exact McNemar. Secondary, all reported: prefix injection, the HarmBench classifier, $N=1$ and $N=32$, truncation and $N=8$ at about half the experts, and a rerun of the original model with the requests in a different order. 188 of the AdvBench requests were among the 256 that estimate the refusal direction of Section~4.3, which no setting of the policy uses.

\subsection{Accuracy, Throughput, Compute, and Release}
\label{app:compute}

\textbf{Accuracy and throughput.} MMLU (all subjects, test split, zero-shot; the answer is the highest-probability letter among the top-20 first-token log-probabilities), GSM8K (test split, zero-shot, 512 tokens, last number), XSTest (450 prompts, 256 tokens, refusal by prefix match), and MATH-500 (the 500-problem test subset, zero-shot chain of thought asked to box the answer, greedy, 1{,}024 tokens, scored with math-verify; scripts/eval\_math.py). Throughput: GSM8K questions as prompts, 256 output tokens with \texttt{ignore\_eos}, batch sizes 1, 8, 32, 128, and 256, one warm-up and three timed repetitions (median reported), prefix caching disabled; first-token time measured with one output token.

\textbf{Compute.} A Study 2 generation job (original, three truncation levels, and the rerun, 1{,}416 requests each) takes 12 to 24 minutes and its judging 17 to 24 minutes; the Qwen1.5 mitigation job (eight conditions and judging) took 56 minutes. Study 2 ran as 52 one-GPU jobs totaling about 83 RTX A6000 GPU-hours, measured from the job logs because Slurm accounting is disabled on our cluster.

\textbf{Release and disclosure.} We will publicly release the harness, the engine patch, prompt identifiers, per-response labels, judge outputs, and response hashes. Raw model outputs, the human-review pages, and annotator notes contain harmful instructions; we share them only on request for research use under a data-use agreement, and do not post them publicly. We will inform the vLLM and llama.cpp maintainers and the authors of the operating-point study of the result and of \method. The marginal uplift to an attacker is small: the attacks we use are public, and the regression requires the operator, not the attacker, to change the serving configuration.

\section{Full Results and Robustness Checks}
\label{app:full}

\subsection{Speed, Expert Weight Reads, and Online Serving}
\label{app:speed}

\textbf{Timing and serving.} Table~\ref{tab:fused} compares the fused budget kernel with the PyTorch version of the patch in the same interleaved jobs; Figure~1 of the paper uses the kernel. Table~\ref{tab:timing} gives the earlier interleaved offline timings with the PyTorch version; Figure~2 of the paper takes the original model, truncation, and \method from the job of Table~\ref{tab:fused}. Table~\ref{tab:serve} gives the online-serving runs with the PyTorch version, Table~\ref{tab:serve_fused} the same runs with the fused kernel over three arrival seeds, with every condition under the same arrivals, and Table~\ref{tab:serve_n0} an ablation with $N=0$, which protects every prompt token but the last. An earlier set of engine timings taken in separate jobs from the runs of the original model disagreed with these at small batch sizes (for example $+51.6$\% for \method on DeepSeek-V2-Lite at batch size 1), so every engine speedup we report is against the original model timed in the same job; the half-budget rows of Table~\ref{tab:fused} come from a second job with its own runs of the original model and truncation. Truncation at $k'=\lceil 2k/3\rceil$ varies by at most 0.5 points across our offline timing jobs (Tables~\ref{tab:timing}, \ref{tab:fused}, \ref{tab:nsweep}, and~\ref{tab:rule_timing}) at batch sizes 8 to 128, far less than the 1.6 to 26.7 points by which \method at half the experts is faster than it. With the fused kernel, truncating every generated token through the patch ($N=0$) keeps 89, 98, and 99\% of native truncation's batch-8 speedup on Qwen1.5, DeepSeek-V2-Lite, and OLMoE (14.2, 26.8, and 16.1 against 16.0, 27.3, and 16.2\%, same job; Table~\ref{tab:nsweep}). $N=0$ and $N=1$ run identical decode steps and differ only in prefill, which first-token time subtracts, so their difference (1.2, 0.8, and 0.4 points) is this job's timing noise; $N=8$ falls below $N=1$ by 0.6, 2.4, and 0.7 points, beyond the job's largest such difference only on DeepSeek-V2-Lite. The batch-size bound of Section~5.3 ran on Qwen1.5 in one more interleaved job of two rounds: with $B=32$, \method speeds up decoding by 4.4, 14.2, 7.1, $-0.3$, and $+0.2$\% at batch sizes 1, 8, 32, 128, and 256, against 4.2, 14.7, 7.2, 0.2, and 0.8\% without the bound and 7.4, 16.0, 8.3, 0.9, and 1.4\% for truncation (scripts/dt\_engine.py, \texttt{DT\_MAXTOK}). With the PyTorch version and no token truncated, the patch costs 1.7 to 2.5\% at batch size 8, and truncating every generated token through it ($N=0$) gives 12.4 to 24.2\% where native truncation gives 15.8 to 27.5\%. Under online serving, the conditions change mean response length by at most 10\% (Qwen1.5, truncation), so we compare time per output token rather than throughput. Online gains exceed the offline ones at a fixed batch size because a faster step also shortens each request's stay and so the running batch: at 1 request per second the mean number of running requests (arrival rate times time in the system) falls from 4.0 to 2.9 on Qwen1.5 and from 4.4 to 3.2 on DeepSeek-V2-Lite under truncation, while decode-step time grows with batch size (Qwen1.5: 9.1\,ms at batch size 1, 22.6\,ms at 8). Truncation also shortens GSM8K responses (on Qwen1.5 at 1 request per second, 189 tokens on average against 206 for the original model and 204 for \method), which lowers the running batch further. With every request generating exactly 256 tokens (\texttt{ignore\_eos}, two arrival seeds, the padding fix in place), truncation's gain at 1, 2, 4, and 8 requests per second is 29.8, 18.5, 8.9, and 2.4\%, and \method keeps 81, 83, 72, and 15\% of it; at 8 requests per second the offered 2{,}048 tokens per second exceed what the GPU completes (about 1{,}230 in every condition), so that column measures a saturated queue (protecting only the prompt, $N=0$: 94, 86, 74, and 43\%).

\begin{table*}[h]
\caption{Decode speedup over the original model (\%, 256 output tokens), with all conditions of a model timed in one job, interleaved over rounds (three for Qwen1.5, two for the others; three timings per round). Medians over timings; Figure~2 of the paper plots the job of Table~\ref{tab:fused}. Patch rows isolate the engine patch: no token truncated, and every generated token truncated through the patch ($N=0$). Patch and \method rows use the PyTorch version of the patch; Table~\ref{tab:fused} compares the fused kernel.}
\label{tab:timing}
\vskip 0.05in
\begin{center}
\begin{scriptsize}
\setlength{\tabcolsep}{3pt}
\begin{tabular}{llrrrrr}
\toprule
 & & \multicolumn{5}{c}{Batch size} \\
\cmidrule(l){3-7}
Model & Condition & 1 & 8 & 32 & 128 & 256 \\
\midrule
Qwen1.5 & Truncation $k'=3$ & $+7.3$ & $+15.8$ & $+8.1$ & $+0.7$ & $+0.6$ \\
 & \method $N=8$, $k'=3$ & $-0.1$ & $+11.7$ & $+6.6$ & $-0.4$ & $-0.3$ \\
 & \method $N=32$, $k'=3$ & $-1.5$ & $+9.0$ & $+6.2$ & $-1.1$ & $-5.3$ \\
 & Patch, $N=0$, $k'=3$ & $-0.1$ & $+12.6$ & $+6.0$ & $-0.6$ & $-0.3$ \\
 & Patch, no token truncated & $-5.7$ & $-2.4$ & $-2.0$ & $-1.5$ & $-5.8$ \\
\addlinespace
 & Truncation $k'=2$ & $+15.5$ & $+38.2$ & $+25.5$ & $+3.3$ & $-3.5$ \\
 & \method $N=8$, $k'=2$ & $+6.0$ & $+31.6$ & $+15.2$ & $-3.2$ & $-3.6$ \\
\midrule
DeepSeek-V2-Lite & Truncation $k'=4$ & $+1.6$ & $+27.5$ & $+11.2$ & $+2.2$ & $+2.5$ \\
 & \method $N=8$, $k'=4$ & $-0.3$ & $+21.9$ & $+10.2$ & $+1.3$ & $+0.6$ \\
 & \method $N=32$, $k'=4$ & $-0.2$ & $+19.3$ & $+8.5$ & $+0.8$ & $+0.2$ \\
 & Patch, $N=0$, $k'=4$ & $+0.2$ & $+24.2$ & $+9.4$ & $+1.0$ & $+0.6$ \\
 & Patch, no token truncated & $+0.3$ & $-1.7$ & $-1.2$ & $-0.7$ & $-0.9$ \\
\addlinespace
 & Truncation $k'=3$ & $+1.1$ & $+49.0$ & $+23.6$ & $+5.4$ & $+3.9$ \\
 & \method $N=8$, $k'=3$ & $-1.6$ & $+40.5$ & $+22.1$ & $+4.3$ & $+2.0$ \\
\midrule
OLMoE & Truncation $k'=6$ & $+13.7$ & $+16.1$ & $+11.4$ & $+4.1$ & $+3.0$ \\
 & \method $N=8$, $k'=6$ & $+4.7$ & $+11.3$ & $+8.1$ & $+2.4$ & $+0.8$ \\
 & \method $N=32$, $k'=6$ & $+3.7$ & $+10.1$ & $+7.0$ & $+1.8$ & $+0.6$ \\
 & Patch, $N=0$, $k'=6$ & $+5.6$ & $+12.4$ & $+9.0$ & $+2.8$ & $+1.3$ \\
 & Patch, no token truncated & $-5.6$ & $-2.5$ & $-1.5$ & $-1.5$ & $-1.3$ \\
\addlinespace
 & Truncation $k'=4$ & $+33.3$ & $+44.2$ & $+28.7$ & $+10.8$ & $+8.3$ \\
 & \method $N=8$, $k'=4$ & $+19.9$ & $+37.0$ & $+24.7$ & $+9.5$ & $+5.5$ \\
\bottomrule
\end{tabular}
\end{scriptsize}
\end{center}
\end{table*}

\begin{table*}[h]
\caption{Decode speedup over the original model (\%, 256 output tokens) with the fused budget kernel and with the PyTorch version of the patch, all conditions of a model timed in one job and interleaved over two rounds. Medians over timings. TTFT: change in time to first token at batch size 8 (\%, negative is shorter).}
\label{tab:fused}
\vskip 0.05in
\begin{center}
\begin{scriptsize}
\setlength{\tabcolsep}{3pt}
\begin{tabular}{llrrrrr}
\toprule
 & & \multicolumn{4}{c}{Batch size} & \\
\cmidrule(lr){3-6}
Model & Condition & 1 & 8 & 32 & 128 & TTFT \\
\midrule
Qwen1.5 & Truncation $k'=3$ & $+7.4$ & $+15.8$ & $+8.2$ & $+0.9$ & $-3.2$ \\
 & Patch, nothing truncated (PyTorch) & $-6.0$ & $-2.6$ & $-1.6$ & $-1.0$ & $+1.3$ \\
 & Patch, nothing truncated (kernel) & $-1.9$ & $-1.0$ & $-0.4$ & $-0.1$ & $+0.4$ \\
 & \method $N=8$ (PyTorch) & $+0.1$ & $+12.9$ & $+7.7$ & $+0.5$ & $-0.0$ \\
 & \method $N=8$ (kernel) & $+5.7$ & $+15.9$ & $+8.3$ & $+1.2$ & $-0.6$ \\
\addlinespace
 & Truncation $k'=2$, separate job & $+15.9$ & $+39.1$ & $+25.9$ & $+3.8$ & $-10.4$ \\
 & \method $N=8$ (kernel), $k'=2$, separate job & $+12.0$ & $+38.3$ & $+25.2$ & $+2.5$ & $+0.2$ \\
\midrule
DeepSeek-V2-Lite & Truncation $k'=4$ & $+3.0$ & $+27.4$ & $+11.4$ & $+2.2$ & $-5.7$ \\
 & Patch, nothing truncated (PyTorch) & $+0.5$ & $-1.8$ & $-1.1$ & $-0.6$ & $+0.7$ \\
 & Patch, nothing truncated (kernel) & $+1.8$ & $-0.3$ & $-0.1$ & $-0.0$ & $+0.0$ \\
 & \method $N=8$ (PyTorch) & $+1.3$ & $+21.8$ & $+10.3$ & $+1.3$ & $+0.6$ \\
 & \method $N=8$ (kernel) & $+2.3$ & $+23.4$ & $+11.9$ & $+1.8$ & $-0.1$ \\
\addlinespace
 & Truncation $k'=3$, separate job & $+1.6$ & $+49.1$ & $+23.8$ & $+5.6$ & $-10.7$ \\
 & \method $N=8$ (kernel), $k'=3$, separate job & $-1.5$ & $+44.3$ & $+24.9$ & $+4.9$ & $-0.1$ \\
\midrule
OLMoE & Truncation $k'=6$ & $+14.4$ & $+16.0$ & $+11.1$ & $+4.0$ & $-5.3$ \\
 & Patch, nothing truncated (PyTorch) & $-5.7$ & $-2.7$ & $-1.8$ & $-1.5$ & $+1.4$ \\
 & Patch, nothing truncated (kernel) & $-1.8$ & $-1.0$ & $-0.5$ & $-0.7$ & $+0.8$ \\
 & \method $N=8$ (PyTorch) & $+5.7$ & $+10.9$ & $+7.8$ & $+2.2$ & $+1.5$ \\
 & \method $N=8$ (kernel) & $+11.9$ & $+15.0$ & $+9.8$ & $+3.2$ & $+0.7$ \\
\addlinespace
 & Truncation $k'=4$, separate job & $+33.8$ & $+44.6$ & $+28.6$ & $+10.9$ & $-13.6$ \\
 & \method $N=8$ (kernel), $k'=4$, separate job & $+28.3$ & $+42.7$ & $+27.0$ & $+10.8$ & $-0.5$ \\
\bottomrule
\end{tabular}
\end{scriptsize}
\end{center}
\end{table*}

\begin{figure}[h]
\centering
\includegraphics[width=\columnwidth]{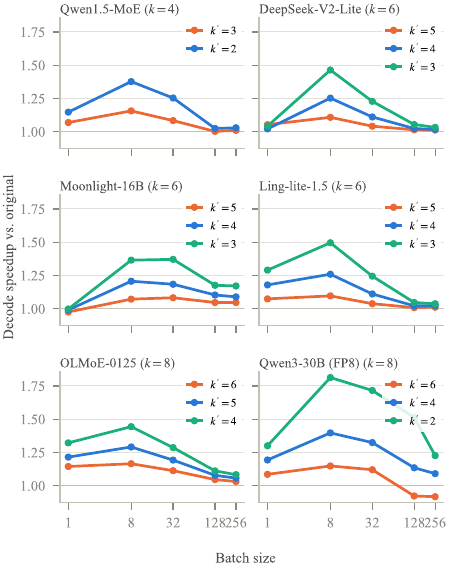}
\caption{Decode speedup over the original model by batch size: generation time minus first-token time, 256 output tokens, median of three runs, one RTX A6000, vLLM 0.11.0 with default fused-MoE kernel settings and prefix caching disabled. The original model and truncation are timed in the same job; Figure~2 of the paper gives interleaved timings with \method. At $k'=\lceil 2k/3\rceil$, truncation speeds up decoding by 15.6 and 0.2\% at batch sizes 8 and 128 on Qwen1.5, 25.2 and 2.4\% on DeepSeek-V2-Lite, 20.6 and 10.3\% on Moonlight, 25.9 and 2.1\% on Ling-lite, 16.5 and 4.6\% on OLMoE, and 14.8 and $-7.8$\% on Qwen3.}
\label{fig:frontier}
\end{figure}

\textbf{Expert weight reads.} Table~\ref{tab:traffic} counts, for every MoE layer call in a decode step where all requests of the batch decode together, the distinct routed experts the batch selects (scripts/traffic.py; a wrapper around vLLM's expert selection, eager mode, so CUDA graphs do not hide calls). Real routing reads fewer experts than uniform routing at every batch size above 1, most on OLMoE, which is why truncation still removes 7.1\% of its reads at batch size 128. A decode speedup $s$ shortens decode time by $s/(1+s)$. At batch sizes 8 to 128 the time saved is below the reads removed in every cell, as expected when attention, the dense layers, and per-step overheads do not shrink; at batch size 256 truncation removes at most 5.2\% of the reads.

\begin{table*}[h]
\caption{Expert weight reads per decode step (GSM8K prompts, 64 output tokens, eager mode): distinct routed experts per MoE layer and step for the original model and truncation at $k'$, uniform routing in parentheses, the share of expert weight reads that truncation removes (shared experts counted in routed-expert units), and the interleaved decode speedup of truncation (Table~\ref{tab:fused}, the job of Figure~2 of the paper; not timed at batch size 256) with the decode time it saves, $s/(1+s)$. Threshold skipping and top-$p$ routing use the calibrated parameters of Table~\ref{tab:methods_hb}, applied to every token.}
\label{tab:traffic}
\vskip 0.05in
\begin{center}
\begin{scriptsize}
\setlength{\tabcolsep}{3pt}
\begin{tabular}{lllr@{\,}lr@{\,}lr@{\,}lr@{\,}lr@{\,}l}
\toprule
 & & & \multicolumn{10}{c}{Batch size} \\
\cmidrule(l){4-13}
Model & Condition & Quantity & 1 & & 8 & & 32 & & 128 & & 256 & \\
\midrule
Qwen1.5 & Original & Experts read & 4.0 & \textcolor{vizink2}{(4.0)} & 22.1 & \textcolor{vizink2}{(25.5)} & 49.1 & \textcolor{vizink2}{(53.4)} & 59.1 & \textcolor{vizink2}{(60.0)} & 59.8 & \textcolor{vizink2}{(60.0)} \\
 & Truncation $k'=3$ & Experts read & 3.0 & \textcolor{vizink2}{(3.0)} & 17.8 & \textcolor{vizink2}{(20.2)} & 44.2 & \textcolor{vizink2}{(48.4)} & 58.5 & \textcolor{vizink2}{(59.9)} & 59.7 & \textcolor{vizink2}{(60.0)} \\
 &  & Reads removed (\%) & 12.5 &  & 16.5 &  & 9.1 &  & 1.0 &  & 0.2 &  \\
 &  & Decode speedup (\%) & $+7.4$ &  & $+15.8$ &  & $+8.2$ &  & $+0.9$ &  & -- &  \\
 &  & Decode time saved (\%) & 6.9 &  & 13.6 &  & 7.5 &  & 0.9 &  & -- &  \\
 & Threshold & Experts read & 2.8 &  & 16.9 &  & 42.0 &  & 57.8 &  & 59.5 &  \\
 &  & Experts per decode token & 2.79 &  & 2.86 &  & 2.92 &  & 2.92 &  & 2.93 &  \\
 & Top-$p$ & Experts read & 2.8 &  & 19.2 &  & 44.8 &  & 58.0 &  & 59.6 &  \\
 &  & Experts per decode token & 2.82 &  & 2.91 &  & 2.90 &  & 2.87 &  & 2.87 &  \\
\midrule
DeepSeek-V2-Lite & Original & Experts read & 6.0 & \textcolor{vizink2}{(6.0)} & 29.8 & \textcolor{vizink2}{(34.9)} & 51.7 & \textcolor{vizink2}{(61.3)} & 60.8 & \textcolor{vizink2}{(64.0)} & 62.7 & \textcolor{vizink2}{(64.0)} \\
 & Truncation $k'=4$ & Experts read & 4.0 & \textcolor{vizink2}{(4.0)} & 20.2 & \textcolor{vizink2}{(25.8)} & 44.0 & \textcolor{vizink2}{(55.9)} & 59.2 & \textcolor{vizink2}{(64.0)} & 61.9 & \textcolor{vizink2}{(64.0)} \\
 &  & Reads removed (\%) & 25.0 &  & 30.3 &  & 14.4 &  & 2.5 &  & 1.2 &  \\
 &  & Decode speedup (\%) & $+3.0$ &  & $+27.4$ &  & $+11.4$ &  & $+2.2$ &  & -- &  \\
 &  & Decode time saved (\%) & 2.9 &  & 21.5 &  & 10.2 &  & 2.2 &  & -- &  \\
 & Threshold & Experts read & 3.6 &  & 20.0 &  & 42.3 &  & 57.9 &  & 61.6 &  \\
 &  & Experts per decode token & 3.62 &  & 3.67 &  & 3.67 &  & 3.66 &  & 3.66 &  \\
 & Top-$p$ & Experts read & 4.5 &  & 20.8 &  & 45.7 &  & 58.8 &  & 62.0 &  \\
 &  & Experts per decode token & 4.55 &  & 4.07 &  & 3.97 &  & 4.02 &  & 4.03 &  \\
 & Threshold, decode-matched & Experts read & 4.1 &  & 23.2 &  & 45.1 &  & 59.5 &  & 62.1 &  \\
 &  & Experts per decode token & 4.13 &  & 4.20 &  & 4.08 &  & 4.10 &  & 4.08 &  \\
\midrule
OLMoE & Original & Experts read & 8.0 & \textcolor{vizink2}{(8.0)} & 29.5 & \textcolor{vizink2}{(42.0)} & 50.7 & \textcolor{vizink2}{(63.1)} & 58.2 & \textcolor{vizink2}{(64.0)} & 60.6 & \textcolor{vizink2}{(64.0)} \\
 & Truncation $k'=6$ & Experts read & 6.0 & \textcolor{vizink2}{(6.0)} & 24.2 & \textcolor{vizink2}{(34.9)} & 43.8 & \textcolor{vizink2}{(61.3)} & 54.1 & \textcolor{vizink2}{(64.0)} & 57.5 & \textcolor{vizink2}{(64.0)} \\
 &  & Reads removed (\%) & 25.0 &  & 18.1 &  & 13.5 &  & 7.1 &  & 5.2 &  \\
 &  & Decode speedup (\%) & $+14.4$ &  & $+16.0$ &  & $+11.1$ &  & $+4.0$ &  & -- &  \\
 &  & Decode time saved (\%) & 12.6 &  & 13.8 &  & 10.0 &  & 3.9 &  & -- &  \\
 & Threshold & Experts read & 5.0 &  & 22.9 &  & 41.9 &  & 54.0 &  & 57.8 &  \\
 &  & Experts per decode token & 5.01 &  & 4.83 &  & 4.87 &  & 4.82 &  & 4.83 &  \\
 & Top-$p$ & Experts read & 5.0 &  & 24.1 &  & 42.4 &  & 55.1 &  & 58.8 &  \\
 &  & Experts per decode token & 4.97 &  & 4.77 &  & 4.84 &  & 4.81 &  & 4.82 &  \\
 & Threshold, decode-matched & Experts read & 6.2 &  & 25.3 &  & 44.9 &  & 55.6 &  & 59.0 &  \\
 &  & Experts per decode token & 6.24 &  & 6.32 &  & 6.25 &  & 6.23 &  & 6.25 &  \\
 & Top-$p$, decode-matched & Experts read & 6.5 &  & 27.9 &  & 47.7 &  & 57.7 &  & 60.5 &  \\
 &  & Experts per decode token & 6.46 &  & 6.29 &  & 6.33 &  & 6.32 &  & 6.35 &  \\
\bottomrule
\end{tabular}
\end{scriptsize}
\end{center}
\end{table*}

\textbf{Kernel profile.} Table~\ref{tab:profile} profiles 32 decode steps after prefill with the PyTorch profiler in eager mode (scripts/moe\_profile.py) and subtracts a prefill-only run with the same prompts. In the original model, GPU time per step is 4 to 32\% below the wall-clock time per decode step measured with CUDA graphs. With no token truncated, the fused patch adds at most 2.0\% of GPU time per step. At batch size 1, truncation saves 7.0, 12.6, and 13.1\% of GPU time per step on Qwen1.5, DeepSeek-V2-Lite, and OLMoE and 6.8, 1.6, and 12.0\% of decode time end to end (Table~\ref{tab:timing}). vLLM runs Qwen1.5 and OLMoE with full CUDA graphs for decode, and 97 and 92\% of the GPU-time saving appears end to end; the MLA attention backend of DeepSeek-V2-Lite supports only piecewise CUDA graphs, and 13\% of its saving appears end to end at batch size 1, against 93\% at batch size 8. The routed reads removed times the routed-expert share of the original model's step predict the change in GPU time per step within 2.4 points in all 12 cells: 8.2, 12.4, 7.7, and 0.8\% at batch sizes 1, 8, 32, and 128 against 7.0, 10.3, 7.5, and 1.8\% measured on Qwen1.5, 13.7, 22.9, 12.0, and 2.1 against 12.6, 23.3, 13.6, and 4.4\% on DeepSeek-V2-Lite, and 14.6, 13.7, 11.4, and 5.8 against 13.1, 12.5, 13.7, and 8.1\% on OLMoE.

\begin{table*}[h]
\caption{Kernel profile of a decode step (PyTorch profiler, eager mode, GSM8K prompts, 32 decode steps after prefill): GPU time per step, the routed-expert kernels' time and share of it (fused MoE GEMMs, routing top-$k$, block alignment, and output sum; shared experts count as other kernels), the change against the original model in routed-expert kernel time and GPU time per step for truncation and in GPU time per step for the fused patch with no token truncated, next to the change in routed expert reads (Table~\ref{tab:traffic}).}
\label{tab:profile}
\vskip 0.05in
\begin{center}
\begin{scriptsize}
\setlength{\tabcolsep}{3pt}
\begin{tabular}{lllrrrr}
\toprule
 & & & \multicolumn{4}{c}{Batch size} \\
\cmidrule(l){4-7}
Model & Condition & Quantity & 1 & 8 & 32 & 128 \\
\midrule
Qwen1.5 & Original & GPU ms per step & 8.72 & 18.60 & 34.89 & 62.08 \\
 & & Routed-expert share (\%) & 33 & 63 & 78 & 82 \\
 & Truncation $k'=3$ & Routed-expert time (\%) & $-21.1$ & $-16.1$ & $-9.4$ & $-2.2$ \\
 & & Routed expert reads (\%) & $-25.0$ & $-19.5$ & $-9.9$ & $-1.0$ \\
 & & GPU time per step (\%) & $-7.0$ & $-10.3$ & $-7.5$ & $-1.8$ \\
 & Patch only & GPU time per step (\%) & $+2.0$ & $+0.8$ & $+0.4$ & $+0.1$ \\
\midrule
DeepSeek-V2-Lite & Original & GPU ms per step & 10.37 & 23.85 & 38.10 & 66.89 \\
 & & Routed-expert share (\%) & 41 & 71 & 81 & 83 \\
 & Truncation $k'=4$ & Routed-expert time (\%) & $-28.8$ & $-31.9$ & $-16.3$ & $-4.5$ \\
 & & Routed expert reads (\%) & $-33.3$ & $-32.3$ & $-14.9$ & $-2.6$ \\
 & & GPU time per step (\%) & $-12.6$ & $-23.3$ & $-13.6$ & $-4.4$ \\
 & Patch only & GPU time per step (\%) & $+0.6$ & $+0.1$ & $+0.2$ & $-0.1$ \\
\midrule
OLMoE & Original & GPU ms per step & 4.47 & 9.62 & 16.68 & 28.47 \\
 & & Routed-expert share (\%) & 58 & 76 & 84 & 81 \\
 & Truncation $k'=6$ & Routed-expert time (\%) & $-22.5$ & $-16.6$ & $-16.3$ & $-9.5$ \\
 & & Routed expert reads (\%) & $-25.0$ & $-18.1$ & $-13.5$ & $-7.1$ \\
 & & GPU time per step (\%) & $-13.1$ & $-12.5$ & $-13.7$ & $-8.1$ \\
 & Patch only & GPU time per step (\%) & $+1.0$ & $+0.5$ & $+0.2$ & $+0.5$ \\
\bottomrule
\end{tabular}
\end{scriptsize}
\end{center}
\end{table*}

\begin{table*}[h]
\caption{Online serving with Poisson arrivals (200 GSM8K requests per rate, greedy, at most 512 tokens, one RTX A6000). Change in median time per output token against the original model (\%, positive is faster) and median time to first token (ms). \method rows use the PyTorch version of the patch; Table~\ref{tab:serve_fused} gives the fused kernel.}
\label{tab:serve}
\vskip 0.05in
\begin{center}
\begin{scriptsize}
\setlength{\tabcolsep}{3pt}
\begin{tabular}{llrrrrrrrr}
\toprule
 & & \multicolumn{4}{c}{TPOT change (\%)} & \multicolumn{4}{c}{TTFT p50 (ms)} \\
\cmidrule(lr){3-6}\cmidrule(l){7-10}
Model & Condition & 1/s & 2/s & 4/s & 8/s & 1/s & 2/s & 4/s & 8/s \\
\midrule
Qwen1.5 & Original & $+0.0$ & $+0.0$ & $+0.0$ & $+0.0$ & 68 & 80 & 85 & 96 \\
 & Truncation $k'=3$ & $+30.9$ & $+19.6$ & $+9.0$ & $+6.6$ & 65 & 73 & 86 & 100 \\
 & \method $N=8$, $k'=3$ & $+14.8$ & $+10.2$ & $+2.0$ & $+4.0$ & 68 & 77 & 88 & 99 \\
 & \method $N=32$, $k'=3$ & $+12.7$ & $+9.1$ & $+2.6$ & $+3.8$ & 68 & 76 & 87 & 100 \\
\addlinespace
 & \method $N=8$, $k'=2$ & $+34.8$ & $+30.9$ & $+20.5$ & $+8.2$ & 66 & 73 & 84 & 99 \\
\midrule
DeepSeek-V2-Lite & Original & $+0.0$ & $+0.0$ & $+0.0$ & $+0.0$ & 76 & 84 & 99 & 112 \\
 & Truncation $k'=4$ & $+39.0$ & $+21.0$ & $+10.2$ & $+4.7$ & 70 & 80 & 96 & 110 \\
 & \method $N=8$, $k'=4$ & $+15.0$ & $+9.6$ & $+5.7$ & $+14.5$ & 76 & 86 & 98 & 107 \\
 & \method $N=32$, $k'=4$ & $+18.8$ & $+5.6$ & $+0.0$ & $-1.2$ & 74 & 87 & 104 & 115 \\
\addlinespace
 & \method $N=8$, $k'=3$ & $+24.6$ & $+17.8$ & $+9.8$ & $+14.9$ & 76 & 83 & 100 & 110 \\
\midrule
OLMoE & Original & $+0.0$ & $+0.0$ & $+0.0$ & $+0.0$ & 31 & 35 & 42 & 45 \\
 & Truncation $k'=6$ & $+23.7$ & $+20.8$ & $+16.3$ & $+6.3$ & 30 & 32 & 38 & 44 \\
 & \method $N=8$, $k'=6$ & $+14.1$ & $+10.7$ & $+6.4$ & $+3.5$ & 31 & 34 & 40 & 44 \\
 & \method $N=32$, $k'=6$ & $+9.0$ & $+11.5$ & $+7.0$ & $+3.1$ & 31 & 35 & 40 & 44 \\
\addlinespace
 & \method $N=8$, $k'=4$ & $+37.7$ & $+42.2$ & $+32.5$ & $+16.3$ & 31 & 33 & 37 & 44 \\
\bottomrule
\end{tabular}
\end{scriptsize}
\end{center}
\end{table*}

\begin{table*}[h]
\caption{Online serving with the fused budget kernel: Poisson arrivals, 200 GSM8K requests per rate, greedy, at most 512 tokens, one RTX A6000, three arrival seeds with all conditions run under the same arrivals. Change in median time per output token against the original model (\%, positive is faster; mean with minimum and maximum over seeds) and median time to first token (ms, mean over seeds).}
\label{tab:serve_fused}
\vskip 0.05in
\begin{center}
\begin{scriptsize}
\setlength{\tabcolsep}{2.5pt}
\begin{tabular}{llr@{\,}lr@{\,}lr@{\,}lr@{\,}lrrrr}
\toprule
 & & \multicolumn{8}{c}{TPOT change (\%)} & \multicolumn{4}{c}{TTFT p50 (ms)} \\
\cmidrule(lr){3-10}\cmidrule(l){11-14}
Model & Condition & \multicolumn{2}{c}{1/s} & \multicolumn{2}{c}{2/s} & \multicolumn{2}{c}{4/s} & \multicolumn{2}{c}{8/s} & 1/s & 2/s & 4/s & 8/s \\
\midrule
Qwen1.5 & Truncation $k'=3$ & $+36.7$ & \textcolor{vizink2}{\tiny$[+30.0, +44.2]$} & $+21.8$ & \textcolor{vizink2}{\tiny$[+17.9, +24.5]$} & $+10.4$ & \textcolor{vizink2}{\tiny$[+9.5, +11.7]$} & $+4.1$ & \textcolor{vizink2}{\tiny$[+2.0, +5.9]$} & 64 & 72 & 83 & 98 \\
 & Patch, nothing truncated & $-1.0$ & \textcolor{vizink2}{\tiny$[-2.0, +0.5]$} & $-0.8$ & \textcolor{vizink2}{\tiny$[-1.1, -0.6]$} & $-1.2$ & \textcolor{vizink2}{\tiny$[-1.3, -1.1]$} & $-0.6$ & \textcolor{vizink2}{\tiny$[-1.6, -0.1]$} & 69 & 79 & 88 & 101 \\
 & \method $N=8$ & $+21.6$ & \textcolor{vizink2}{\tiny$[+16.1, +26.2]$} & $+12.0$ & \textcolor{vizink2}{\tiny$[+10.9, +13.2]$} & $+5.5$ & \textcolor{vizink2}{\tiny$[+3.1, +8.5]$} & $+1.5$ & \textcolor{vizink2}{\tiny$[+0.8, +2.0]$} & 67 & 76 & 85 & 100 \\
\midrule
DeepSeek-V2-Lite & Truncation $k'=4$ & $+48.3$ & \textcolor{vizink2}{\tiny$[+45.5, +50.3]$} & $+27.0$ & \textcolor{vizink2}{\tiny$[+26.0, +29.0]$} & $+14.6$ & \textcolor{vizink2}{\tiny$[+12.9, +15.5]$} & $+8.2$ & \textcolor{vizink2}{\tiny$[+2.2, +17.0]$} & 67 & 76 & 91 & 104 \\
 & Patch, nothing truncated & $-0.6$ & \textcolor{vizink2}{\tiny$[-3.0, +1.9]$} & $+0.3$ & \textcolor{vizink2}{\tiny$[+0.0, +0.8]$} & $+0.0$ & \textcolor{vizink2}{\tiny$[-0.4, +0.4]$} & $+0.1$ & \textcolor{vizink2}{\tiny$[-1.0, +0.9]$} & 75 & 86 & 98 & 110 \\
 & \method $N=8$ & $+36.3$ & \textcolor{vizink2}{\tiny$[+35.4, +37.6]$} & $+15.1$ & \textcolor{vizink2}{\tiny$[+13.0, +16.8]$} & $+9.6$ & \textcolor{vizink2}{\tiny$[+9.2, +10.2]$} & $+6.9$ & \textcolor{vizink2}{\tiny$[+2.2, +15.3]$} & 73 & 82 & 95 & 109 \\
\midrule
OLMoE & Truncation $k'=6$ & $+21.4$ & \textcolor{vizink2}{\tiny$[+18.9, +25.0]$} & $+25.9$ & \textcolor{vizink2}{\tiny$[+23.5, +30.6]$} & $+14.0$ & \textcolor{vizink2}{\tiny$[+12.8, +15.5]$} & $+7.7$ & \textcolor{vizink2}{\tiny$[+7.1, +8.3]$} & 29 & 32 & 37 & 44 \\
 & Patch, nothing truncated & $-2.4$ & \textcolor{vizink2}{\tiny$[-2.7, -2.2]$} & $+0.5$ & \textcolor{vizink2}{\tiny$[-0.7, +2.1]$} & $-0.5$ & \textcolor{vizink2}{\tiny$[-1.0, +0.5]$} & $+0.7$ & \textcolor{vizink2}{\tiny$[+0.4, +0.9]$} & 31 & 35 & 40 & 46 \\
 & \method $N=8$ & $+19.1$ & \textcolor{vizink2}{\tiny$[+15.8, +23.2]$} & $+17.3$ & \textcolor{vizink2}{\tiny$[+15.6, +18.5]$} & $+7.2$ & \textcolor{vizink2}{\tiny$[+5.3, +9.0]$} & $+5.9$ & \textcolor{vizink2}{\tiny$[+4.8, +6.9]$} & 31 & 33 & 40 & 45 \\
\bottomrule
\end{tabular}
\end{scriptsize}
\end{center}
\end{table*}

\begin{table}[h]
\caption{Online-serving ablation (same setup as Table~\ref{tab:serve_fused}, two arrival seeds, all conditions in one job): share of truncation's gain in median time per output token kept with only the prompt protected ($N=0$, every generated token truncated) and with \method ($N=8$), in \%.}
\label{tab:serve_n0}
\vskip 0.05in
\begin{center}
\begin{scriptsize}
\setlength{\tabcolsep}{3pt}
\begin{tabular}{llrrrr}
\toprule
 & & \multicolumn{4}{c}{Gain kept (\%)} \\
\cmidrule(l){3-6}
Model & Protected & 1/s & 2/s & 4/s & 8/s \\
\midrule
Qwen1.5 & Prompt ($N=0$) & 92 & 67 & 91 & 47 \\
 & Prompt + 8 ($N=8$) & 58 & 65 & 49 & 52 \\
\midrule
DeepSeek-V2-Lite & Prompt ($N=0$) & 78 & 60 & 73 & 122 \\
 & Prompt + 8 ($N=8$) & 75 & 49 & 69 & 140 \\
\midrule
OLMoE & Prompt ($N=0$) & 79 & 72 & 62 & 61 \\
 & Prompt + 8 ($N=8$) & 87 & 78 & 62 & 73 \\
\bottomrule
\end{tabular}
\end{scriptsize}
\end{center}
\end{table}

\begin{table*}[h]
\caption{Online serving after the CUDA-graph padding fix (GSM8K requests, greedy, one RTX A6000). Truncation: change in median time per output token against the original model under the same arrivals (\%, positive is faster). \method and prompt only: share of that gain kept (\%) by \method ($N=8$) and by protecting only the prompt ($N=0$). Natural: Poisson arrivals at 1, 2, 4, and 8 requests per second, natural output lengths, at most 512 tokens, three arrival seeds. Fixed: the same arrivals with every request generating exactly 256 tokens, two seeds. Closed loop: 1, 4, and 16 clients, each sending its next request when the previous one finishes, 256 tokens, 64 requests per level, one run. Shares are ratios of means over seeds; values above 100\% or below 0 occur where whole engine runs ran slower, which also raises their time to first token. Dashes: not run.}
\label{tab:serve_fix}
\vskip 0.05in
\begin{center}
\begin{scriptsize}
\setlength{\tabcolsep}{3pt}
\begin{tabular}{llrrrrrrrrrrr}
\toprule
 & & \multicolumn{4}{c}{Natural (3)} & \multicolumn{4}{c}{Fixed (2)} & \multicolumn{3}{c}{Closed loop (1)} \\
\cmidrule(lr){3-6}\cmidrule(lr){7-10}\cmidrule(l){11-13}
Model & Condition & 1/s & 2/s & 4/s & 8/s & 1/s & 2/s & 4/s & 8/s & 1 & 4 & 16 \\
\midrule
Qwen1.5 & Truncation & $+36.3$ & $+24.5$ & $+11.5$ & $+4.6$ & $+29.8$ & $+18.5$ & $+8.9$ & $+2.4$ & $+4.9$ & $+14.5$ & $+15.7$ \\
 & \method & $64$ & $65$ & $60$ & $38$ & $81$ & $83$ & $72$ & $15$ & $121$ & $90$ & $72$ \\
 & Prompt only & $84$ & $71$ & $84$ & $45$ & $94$ & $86$ & $74$ & $43$ & -- & -- & -- \\
\midrule
DeepSeek-V2-Lite & Truncation & $+21.7$ & $+13.2$ & $+3.8$ & $-7.2$ & $+44.1$ & $+18.8$ & $+6.7$ & $+3.2$ & -- & -- & -- \\
 & \method & $172$ & $157$ & $233$ & $-37$ & $99$ & $88$ & $14$ & $-24$ & -- & -- & -- \\
 & Prompt only & $86$ & $161$ & $345$ & $-51$ & $69$ & $81$ & $74$ & $91$ & -- & -- & -- \\
\midrule
OLMoE & Truncation & $+26.7$ & $+26.3$ & $+14.5$ & $+6.2$ & $+30.6$ & $+25.2$ & $+12.7$ & $+8.0$ & $+15.1$ & $+14.0$ & $+16.0$ \\
 & \method & $96$ & $106$ & $108$ & $162$ & $155$ & $140$ & $141$ & $150$ & $59$ & $76$ & $71$ \\
 & Prompt only & $81$ & $89$ & $84$ & $145$ & $109$ & $121$ & $141$ & $144$ & -- & -- & -- \\
\bottomrule
\end{tabular}
\end{scriptsize}
\end{center}
\end{table*}

\subsection{Refusal Across Models, Budgets, and Templates}
\label{app:refusal}

\begin{figure*}[h]
\centering
\includegraphics[width=\textwidth]{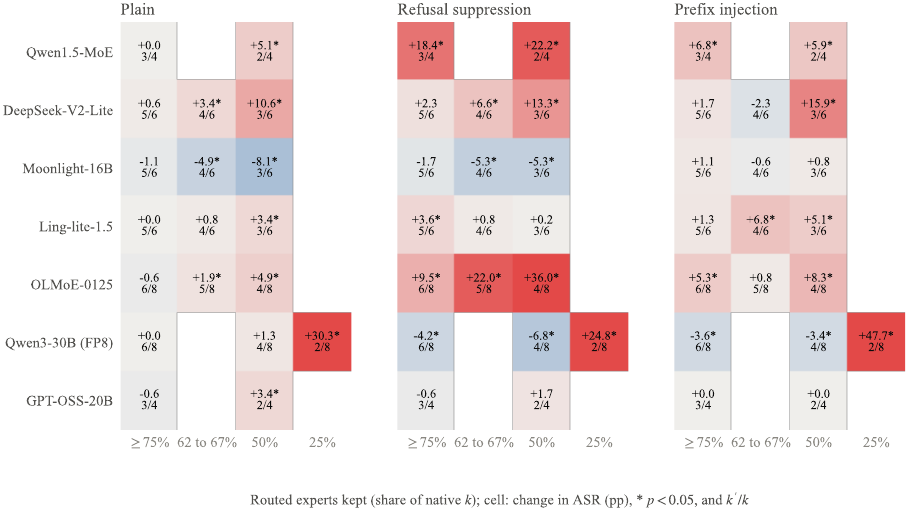}
\caption{ASR change (pp, Llama Guard 3) for each panel model, truncation level, and framing. Each cell gives the change, an asterisk for unadjusted exact McNemar $p<0.05$, and the kept fraction $k'/k$.}
\label{fig:panel}
\end{figure*}

Figure~\ref{fig:panel} shows the ASR change for each panel model, truncation level, and framing, and the per-model tables of this subsection list each truncation level and rerun with both judges. Figure~\ref{fig:mitig} plots truncation, the mass conditions, and emulated \method from Tables~\ref{tab:mitig} and~\ref{tab:mitigpi} for Qwen1.5 and DeepSeek-V2-Lite, plus $N=32$ in emulation on Qwen1.5 ($-1.06$, $+0.21$, and $+2.97$\pp on plain requests, refusal suppression, and prefix injection; $p=0.049$ for the last), and Figure~\ref{fig:onset} the divergence curves at $k'=3$ for both models. Table~\ref{tab:mitigpi} gives the mechanism and mitigation conditions under prefix injection, Table~\ref{tab:strip} re-judges prefix-injection answers with the forced opening removed, Table~\ref{tab:adv} reports two further templates, and Table~\ref{tab:sampling} reports sampled decoding.

\textbf{Degenerate flips.} At the confirmatory point, no forward flip is degenerate on Qwen1.5, DeepSeek-V2-Lite, or Ling-lite, and 1 of 15 is on Moonlight plain requests. GPT-OSS-20B at $k'=2$ ends 162 of its 1{,}416 responses without a final answer, against 3 for the original model.

\textbf{Response length.} On the safety prompts, truncated models write longer responses: on plain requests, where the net label change is zero, the mean length on Qwen1.5 rises from 184.6 to 191.3 tokens (3.6\%), which offsets about a quarter of the batch-8 gain per request.

\textbf{Reordered reruns.} Every panel model has four reordered reruns of the original model. Under refusal suppression the changes against the first run of the original model are $-0.42$, $-0.64$, $+1.06$, and $0.00$\pp on Qwen1.5 and $-0.42$, $-0.64$, $-0.21$, and $+0.85$\pp on DeepSeek-V2-Lite. Across all 84 rerun and framing comparisons of the panel models under Llama Guard 3, at most 12.5\% of prompts change label, the largest net change is $+2.75$\pp (Ling-lite, prefix injection), and 4 exact McNemar tests reach $p<0.05$ (GPT-OSS-20B refusal suppression, $+2.12$ and $+1.91$\pp; OLMoE plain requests, $-1.27$\pp; Qwen3-30B prefix injection, $-1.48$\pp), against 4.2 expected by chance. Under the HarmBench classifier, 2 of 84 reach $p<0.05$ (OLMoE prefix injection, $+3.18$ and $+3.39$\pp), and the largest net change is 3.39\pp. Measured against each of the five runs of the original model instead of the first, the refusal-suppression changes at $k'=\lceil 2k/3\rceil$ range from $+17.37$ to $+19.07$\pp on Qwen1.5, $+5.72$ to $+7.20$ on DeepSeek-V2-Lite, $+9.11$ to $+10.17$ on OLMoE, $-5.93$ to $-4.87$ on Moonlight-16B, $-4.66$ to $-3.81$ on Qwen3-30B, $0.00$ to $+2.75$ on Ling-lite, and $-2.75$ to $-0.64$ on GPT-OSS-20B; for \method ($N=8$, fused kernel) they range from $-1.69$ to $0.00$, $-0.21$ to $+1.27$, and $+0.85$ to $+1.91$\pp on Qwen1.5, DeepSeek-V2-Lite, and OLMoE, none significant (HarmBench classifier: $+0.42$ to $+1.27$, $+1.91$ to $+3.39$, and $+0.85$ to $+3.39$\pp, 3 of 15 with $p<0.05$, within that judge's rerun range).

\textbf{Label churn.} A zero net change does not mean no change: on Qwen1.5 plain requests at $k'=3$, truncation changes 28 labels against 9 for the reordered rerun (excess churn $p=2.1\times10^{-5}$).

\textbf{Variation across models.} Under refusal suppression at $k'=\lceil 2k/3\rceil$, the log ratio of forward to reverse flips differs across the nine models far beyond sampling error (Cochran's $Q=93.3$ on 8 degrees of freedom, $p=10^{-16}$, $I^2=91\%$, DerSimonian--Laird), so no common effect describes the models and each model needs its own evaluation.

\textbf{Held-out models.} DeepSeek-MoE-16B has four reordered reruns of the original model. Under refusal suppression all four lie below its first run of the original model ($-3.60$, $-1.69$, $-1.48$, and $-2.54$\pp; the first has $p=4.9\times10^{-4}$), so its first run of the original model is high; against the mean of its five runs of the original model, truncation to $k'=4$ changes refusal-suppression ASR by $+1.23$\pp and to $k'=3$ by $+15.64$\pp. Qwen2-57B's four reordered reruns change refusal-suppression ASR by $-0.85$ to $+0.42$\pp, and against its five runs of the original model truncation to $k'=6$ changes it by $+1.91$ to $+3.18$\pp. Tables for both models follow those of the seven panel models (Table~2 of the paper) in this appendix. \method (emulation, $N=8$) removes the refusal-suppression increase on both: on DeepSeek-MoE-16B at $k'=3$ from $+13.77$ to $-0.85$\pp (HarmBench classifier $+12.71$ to $+0.64$\pp) and on Qwen2-57B at $k'=5$ from $+4.24$ to $+0.64$\pp ($+4.03$ to $0.00$\pp). On plain requests to DeepSeek-MoE-16B at $k'=3$ it gives $+5.93$\pp against $+5.08$\pp for truncation.

\IfFileExists{tables/app_robust.tex}{\begin{table*}[h]
\caption{Prefix injection judged with the forced opening removed from the answer. ASR change (pp) against the original model, forward/reverse flips; bold: exact McNemar $p<0.05$; shading: size and sign of the Llama Guard 3 change.}
\label{tab:strip}
\vskip 0.05in
\begin{center}
\begin{scriptsize}
\setlength{\tabcolsep}{3pt}
\begin{tabular}{@{}llrr@{}}
\toprule
Model & Condition & Llama Guard 3 & HarmBench cls. \\
\midrule
Qwen1.5 & Original rerun & \cellcolor{vizaxis!6}$-0.85$ (17/21) & $-0.85$ (18/22) \\
 & Truncation $k'=3$ & \cellcolor{vizorange!17}$\mathbf{+6.36}$ (52/22) & $\mathbf{+7.42}$ (59/24) \\
 & Truncation $k'=2$ & \cellcolor{vizorange!17}$\mathbf{+6.36}$ (59/29) & $\mathbf{+5.51}$ (61/35) \\
 & \method $N=8$ (emulation), $k'=3$ & \cellcolor{vizorange!13}$\mathbf{+4.87}$ (41/18) & $+3.81$ (47/29) \\
 & \method $N=64$ (emulation), $k'=3$ & \cellcolor{vizorange!6}$+1.06$ (20/15) & $+0.21$ (23/22) \\
\midrule
DeepSeek-V2-Lite & Original rerun & \cellcolor{vizaxis!6}$-1.06$ (22/27) & $+0.64$ (29/26) \\
 & Truncation $k'=4$ & \cellcolor{vizaxis!13}$\mathbf{-4.66}$ (41/63) & $\mathbf{-4.66}$ (43/65) \\
 & Truncation $k'=3$ & \cellcolor{vizorange!38}$\mathbf{+13.77}$ (104/39) & $\mathbf{+11.02}$ (98/46) \\
\bottomrule
\end{tabular}
\end{scriptsize}
\end{center}
\end{table*}

\begin{table*}[h]
\caption{Two further jailbreak templates from the StrongREJECT collection: distractors (the request is embedded in the second of three tasks) and evil confidant (role play). ASR change (pp, Llama Guard 3) against the original model with forward/reverse flips; bold: exact McNemar $p<0.05$; shading: size and sign of the Llama Guard 3 change; brackets: HarmBench-classifier change.}
\label{tab:adv}
\vskip 0.05in
\begin{center}
\begin{scriptsize}
\setlength{\tabcolsep}{3pt}
\begin{tabular}{@{}llrr}
\toprule
Model & Condition & Distractors & Evil confidant \\
\midrule
Qwen1.5 & original ASR & 69.07 & 79.87 \\
 & Original rerun & \cellcolor{vizorange!6}$+0.64$ (53/50) [$+1.1$] & \cellcolor{vizorange!6}$+1.48$ (20/13) [$+0.4$] \\
 & Patch, nothing truncated & \cellcolor{vizaxis!6}$-0.85$ (45/49) [$+1.1$] & \cellcolor{vizorange!6}$+1.06$ (19/14) [$+0.2$] \\
 & Truncation $k'=3$ & \cellcolor{vizaxis!6}$-0.21$ (70/71) [$+7.8$] & \cellcolor{vizaxis!12}$\mathbf{-4.45}$ (36/57) [$-4.2$] \\
 & \method $N=8$, emulation & \cellcolor{vizorange!8}$+2.75$ (74/61) [$+8.7$] & \cellcolor{vizaxis!6}$-0.42$ (29/31) [$-3.0$] \\
 & \method $N=64$, emulation & \cellcolor{vizorange!6}$+1.06$ (53/48) [$+2.1$] & $+0.00$ (20/20) [$-3.0$] \\
 & \method $N=8$, engine & \cellcolor{vizorange!14}$\mathbf{+5.08}$ (69/45) [$+4.4$] & \cellcolor{vizorange!8}$+2.75$ (33/20) [$+0.4$] \\
 & \method $N=32$, engine & \cellcolor{vizorange!13}$\mathbf{+4.66}$ (57/35) [$+3.0$] & \cellcolor{vizorange!7}$+2.54$ (25/13) [$+0.4$] \\
\midrule
DeepSeek-V2-Lite & original ASR & 85.38 & 13.56 \\
 & Original rerun & $+0.00$ (30/30) [$-0.6$] & \cellcolor{vizorange!6}$+1.27$ (14/8) [$+0.4$] \\
 & Patch, nothing truncated & $+0.00$ (0/0) [$+0.0$] & $+0.00$ (0/0) [$+0.0$] \\
 & Truncation $k'=4$ & \cellcolor{vizaxis!10}$-3.81$ (40/58) [$-4.4$] & \cellcolor{vizorange!36}$\mathbf{+13.14}$ (71/9) [$+11.4$] \\
 & \method $N=8$, emulation & \cellcolor{vizaxis!10}$-3.60$ (41/58) [$-13.6$] & \cellcolor{vizorange!6}$+0.85$ (12/8) [$-0.4$] \\
 & \method $N=64$, emulation & \cellcolor{vizaxis!13}$\mathbf{-4.66}$ (28/50) [$+1.1$] & \cellcolor{vizaxis!6}$-0.85$ (9/13) [$-0.6$] \\
 & \method $N=8$, engine & \cellcolor{vizaxis!12}$\mathbf{-4.24}$ (31/51) [$-1.7$] & \cellcolor{vizaxis!6}$-0.42$ (10/12) [$-1.5$] \\
 & \method $N=32$, engine & \cellcolor{vizaxis!7}$-2.54$ (38/50) [$-1.3$] & \cellcolor{vizorange!6}$+0.85$ (14/10) [$-0.2$] \\
\bottomrule
\end{tabular}
\end{scriptsize}
\end{center}
\end{table*}

\begin{table*}[h]
\caption{Sampled decoding (temperature 0.7, three seeds) on Qwen1.5 under refusal suppression. Each seed compares conditions decoded with the same seed, which every request receives, so a seed's random draws are shared across prompts: the original model's ASR varies more across seeds (53.8, 49.6, and 39.0\%) than independent sampling would, while the paired changes agree. ASR change (pp, Llama Guard 3) against that seed's original run with forward/reverse flips; bold: exact McNemar $p<0.05$; shading: size and sign of the Llama Guard 3 change. \method: $N=8$ in the engine with the fused kernel, and in emulation. HB: HarmBench-classifier change for truncation/engine/emulation.}
\label{tab:sampling}
\vskip 0.05in
\begin{center}
\begin{scriptsize}
\setlength{\tabcolsep}{3pt}
\begin{tabular}{@{}lrrrrr@{}}
\toprule
 & & & \multicolumn{2}{c}{\textcolor{vizblue}{\method}} & \\
\cmidrule(lr){4-5}
Seed & Original & \textcolor{vizorange}{Trunc.\ $k'=3$} & engine & emulation & HB \\
\midrule
1 & 53.8 & \cellcolor{vizorange!47}$\mathbf{+17.16}$ (96/15) & \cellcolor{vizorange!6}$+0.42$ (15/13) & \cellcolor{vizaxis!6}$-0.42$ (17/19) & $+16.3$/$+1.5$/$+2.5$ \\
2 & 49.6 & \cellcolor{vizorange!45}$\mathbf{+16.53}$ (94/16) & \cellcolor{vizaxis!6}$-1.69$ (12/20) & \cellcolor{vizorange!7}$+2.54$ (24/12) & $+12.9$/$-0.4$/$+2.5$ \\
3 & 39.0 & \cellcolor{vizorange!48}$\mathbf{+17.58}$ (99/16) & \cellcolor{vizaxis!6}$-0.85$ (10/14) & \cellcolor{vizorange!6}$+0.64$ (20/17) & $+15.7$/$+1.3$/$+3.8$ \\
\bottomrule
\end{tabular}
\end{scriptsize}
\end{center}
\end{table*}
}{}

\IfFileExists{tables/app_full.tex}{\begin{table*}[h]
\caption{Qwen1.5-MoE: all Study 2 conditions. ASR (\%), change against the original model (pp) with 95\% bootstrap CI, forward/reverse flips, exact McNemar $p$ (unadjusted), degenerate forward flips under Llama Guard 3, and the change under the HarmBench classifier; shading: size and sign of the Llama Guard 3 change. rerun: original model with shuffled prompt order.}
\vskip 0.05in
\begin{center}
\begin{scriptsize}
\setlength{\tabcolsep}{3pt}
\begin{tabular}{@{}llrrrrrrr@{}}
\toprule
Framing & $k$ & ASR & $\Delta$ & 95\% CI & Flips & $p$ & Degen. & $\Delta$ HB \\
\midrule
Plain & 4 & 7.63 & -- & -- & -- & -- & -- & -- \\
 & 3 & 7.63 & $+0.00$ & $[-2.12, 2.12]$ & 14/14 & 1.0 & 0 & $+1.06$ \\
 & 2 & 12.71 & \cellcolor{vizorange!14}$+5.08$ & $[2.12, 8.05]$ & 38/14 & 0.0012 & 0 & $+6.57$ \\
 & rerun 4 & 7.42 & \cellcolor{vizaxis!6}$-0.21$ & -- & 4/5 & 1.0 & -- & $+1.27$ \\
\midrule
Refusal supp. & 4 & 46.61 & -- & -- & -- & -- & -- & -- \\
 & 3 & 65.04 & \cellcolor{vizorange!51}$+18.43$ & $[14.41, 22.67]$ & 101/14 & $\sci{2.0}{-17}$ & 0 & $+16.53$ \\
 & 2 & 68.86 & \cellcolor{vizorange!55}$+22.25$ & $[17.58, 26.69]$ & 123/18 & $\sci{2.0}{-20}$ & 0 & $+22.25$ \\
 & rerun 4 & 46.19 & \cellcolor{vizaxis!6}$-0.42$ & -- & 9/11 & 0.824 & -- & $-0.85$ \\
\midrule
Prefix inj. & 4 & 79.87 & -- & -- & -- & -- & -- & -- \\
 & 3 & 86.65 & \cellcolor{vizorange!19}$+6.78$ & $[3.18, 10.17]$ & 53/21 & $\sci{2.6}{-4}$ & 0 & $+6.99$ \\
 & 2 & 85.81 & \cellcolor{vizorange!16}$+5.93$ & $[2.33, 9.53]$ & 53/25 & 0.00203 & 0 & $+4.87$ \\
 & rerun 4 & 80.30 & \cellcolor{vizorange!6}$+0.42$ & -- & 17/15 & 0.86 & -- & $-1.27$ \\
\bottomrule
\end{tabular}
\end{scriptsize}
\end{center}
\end{table*}

\begin{table*}[h]
\caption{DeepSeek-V2-Lite: all Study 2 conditions. ASR (\%), change against the original model (pp) with 95\% bootstrap CI, forward/reverse flips, exact McNemar $p$ (unadjusted), degenerate forward flips under Llama Guard 3, and the change under the HarmBench classifier; shading: size and sign of the Llama Guard 3 change. rerun: original model with shuffled prompt order.}
\vskip 0.05in
\begin{center}
\begin{scriptsize}
\setlength{\tabcolsep}{3pt}
\begin{tabular}{@{}llrrrrrrr@{}}
\toprule
Framing & $k$ & ASR & $\Delta$ & 95\% CI & Flips & $p$ & Degen. & $\Delta$ HB \\
\midrule
Plain & 6 & 12.29 & -- & -- & -- & -- & -- & -- \\
 & 5 & 12.92 & \cellcolor{vizorange!6}$+0.64$ & $[-1.06, 2.33]$ & 10/7 & 0.629 & 0 & $+0.85$ \\
 & 4 & 15.68 & \cellcolor{vizorange!9}$+3.39$ & $[1.27, 5.51]$ & 21/5 & 0.00249 & 0 & $+2.75$ \\
 & 3 & 22.88 & \cellcolor{vizorange!29}$+10.59$ & $[7.42, 13.98]$ & 59/9 & $\sci{3.9}{-10}$ & 0 & $+9.32$ \\
 & rerun 6 & 11.44 & \cellcolor{vizaxis!6}$-0.85$ & -- & 4/8 & 0.388 & -- & $-1.06$ \\
\midrule
Refusal supp. & 6 & 43.64 & -- & -- & -- & -- & -- & -- \\
 & 5 & 45.97 & \cellcolor{vizorange!6}$+2.33$ & $[-1.06, 5.72]$ & 38/27 & 0.215 & 0 & $+1.48$ \\
 & 4 & 50.21 & \cellcolor{vizorange!18}$+6.57$ & $[2.54, 10.59]$ & 63/32 & 0.00192 & 0 & $+5.30$ \\
 & 3 & 56.99 & \cellcolor{vizorange!37}$+13.35$ & $[8.90, 17.80]$ & 91/28 & $\sci{5.8}{-9}$ & 1 & $+10.59$ \\
 & rerun 6 & 43.22 & \cellcolor{vizaxis!6}$-0.42$ & -- & 17/19 & 0.868 & -- & $-1.06$ \\
\midrule
Prefix inj. & 6 & 45.34 & -- & -- & -- & -- & -- & -- \\
 & 5 & 47.03 & \cellcolor{vizorange!6}$+1.69$ & $[-2.12, 5.72]$ & 49/41 & 0.461 & 0 & $+1.69$ \\
 & 4 & 43.01 & \cellcolor{vizaxis!6}$-2.33$ & $[-6.57, 2.12]$ & 51/62 & 0.347 & 0 & $-5.08$ \\
 & 3 & 61.23 & \cellcolor{vizorange!44}$+15.89$ & $[11.02, 20.76]$ & 113/38 & $\sci{7.7}{-10}$ & 0 & $+10.17$ \\
 & rerun 6 & 45.13 & \cellcolor{vizaxis!6}$-0.21$ & -- & 24/25 & 1.0 & -- & $+1.06$ \\
\bottomrule
\end{tabular}
\end{scriptsize}
\end{center}
\end{table*}

\begin{table*}[h]
\caption{Moonlight-16B: all Study 2 conditions. ASR (\%), change against the original model (pp) with 95\% bootstrap CI, forward/reverse flips, exact McNemar $p$ (unadjusted), degenerate forward flips under Llama Guard 3, and the change under the HarmBench classifier; shading: size and sign of the Llama Guard 3 change. rerun: original model with shuffled prompt order.}
\vskip 0.05in
\begin{center}
\begin{scriptsize}
\setlength{\tabcolsep}{3pt}
\begin{tabular}{@{}llrrrrrrr@{}}
\toprule
Framing & $k$ & ASR & $\Delta$ & 95\% CI & Flips & $p$ & Degen. & $\Delta$ HB \\
\midrule
Plain & 6 & 37.29 & -- & -- & -- & -- & -- & -- \\
 & 5 & 36.23 & \cellcolor{vizaxis!6}$-1.06$ & $[-3.60, 1.48]$ & 16/21 & 0.511 & 0 & $-2.97$ \\
 & 4 & 32.42 & \cellcolor{vizaxis!13}$-4.87$ & $[-7.84, -1.91]$ & 15/38 & 0.00219 & 1 & $-4.87$ \\
 & 3 & 29.24 & \cellcolor{vizaxis!22}$-8.05$ & $[-11.65, -4.66]$ & 19/57 & $\sci{1.5}{-5}$ & 0 & $-8.90$ \\
 & rerun 6 & 36.02 & \cellcolor{vizaxis!6}$-1.27$ & -- & 6/12 & 0.238 & -- & $-1.06$ \\
\midrule
Refusal supp. & 6 & 86.86 & -- & -- & -- & -- & -- & -- \\
 & 5 & 85.17 & \cellcolor{vizaxis!6}$-1.69$ & $[-3.81, 0.42]$ & 11/19 & 0.2 & 0 & $-3.60$ \\
 & 4 & 81.57 & \cellcolor{vizaxis!15}$-5.30$ & $[-8.05, -2.54]$ & 11/36 & $\sci{3.5}{-4}$ & 0 & $-6.36$ \\
 & 3 & 81.57 & \cellcolor{vizaxis!15}$-5.30$ & $[-8.47, -1.91]$ & 18/43 & 0.00187 & 0 & $-9.96$ \\
 & rerun 6 & 86.44 & \cellcolor{vizaxis!6}$-0.42$ & -- & 3/5 & 0.727 & -- & $+0.21$ \\
\midrule
Prefix inj. & 6 & 91.74 & -- & -- & -- & -- & -- & -- \\
 & 5 & 92.80 & \cellcolor{vizorange!6}$+1.06$ & $[-1.06, 3.18]$ & 16/11 & 0.442 & 0 & $-0.21$ \\
 & 4 & 91.10 & \cellcolor{vizaxis!6}$-0.64$ & $[-3.18, 1.91]$ & 17/20 & 0.743 & 0 & $-2.54$ \\
 & 3 & 92.58 & \cellcolor{vizorange!6}$+0.85$ & $[-1.48, 3.18]$ & 19/15 & 0.608 & 0 & $-4.24$ \\
 & rerun 6 & 92.58 & \cellcolor{vizorange!6}$+0.85$ & -- & 9/5 & 0.424 & -- & $+0.64$ \\
\bottomrule
\end{tabular}
\end{scriptsize}
\end{center}
\end{table*}

\begin{table*}[h]
\caption{Ling-lite-1.5: all Study 2 conditions. ASR (\%), change against the original model (pp) with 95\% bootstrap CI, forward/reverse flips, exact McNemar $p$ (unadjusted), degenerate forward flips under Llama Guard 3, and the change under the HarmBench classifier; shading: size and sign of the Llama Guard 3 change. rerun: original model with shuffled prompt order.}
\vskip 0.05in
\begin{center}
\begin{scriptsize}
\setlength{\tabcolsep}{3pt}
\begin{tabular}{@{}llrrrrrrr@{}}
\toprule
Framing & $k$ & ASR & $\Delta$ & 95\% CI & Flips & $p$ & Degen. & $\Delta$ HB \\
\midrule
Plain & 6 & 8.69 & -- & -- & -- & -- & -- & -- \\
 & 5 & 8.69 & $+0.00$ & $[-1.91, 1.91]$ & 10/10 & 1.0 & 0 & $+0.85$ \\
 & 4 & 9.53 & \cellcolor{vizorange!6}$+0.85$ & $[-1.27, 2.97]$ & 15/11 & 0.557 & 0 & $+1.69$ \\
 & 3 & 12.08 & \cellcolor{vizorange!9}$+3.39$ & $[0.64, 6.14]$ & 29/13 & 0.0195 & 0 & $+3.39$ \\
 & rerun 6 & 10.17 & \cellcolor{vizorange!6}$+1.48$ & -- & 10/3 & 0.0923 & -- & $+1.06$ \\
\midrule
Refusal supp. & 6 & 41.74 & -- & -- & -- & -- & -- & -- \\
 & 5 & 45.34 & \cellcolor{vizorange!10}$+3.60$ & $[0.21, 6.99]$ & 41/24 & 0.0464 & 0 & $+3.18$ \\
 & 4 & 42.58 & \cellcolor{vizorange!6}$+0.85$ & $[-2.75, 4.45]$ & 42/38 & 0.738 & 0 & $-4.66$ \\
 & 3 & 41.95 & \cellcolor{vizorange!6}$+0.21$ & $[-3.81, 4.45]$ & 52/51 & 1.0 & 0 & $-2.75$ \\
 & rerun 6 & 42.58 & \cellcolor{vizorange!6}$+0.85$ & -- & 14/10 & 0.541 & -- & $+1.27$ \\
\midrule
Prefix inj. & 6 & 58.47 & -- & -- & -- & -- & -- & -- \\
 & 5 & 59.75 & \cellcolor{vizorange!6}$+1.27$ & $[-1.91, 4.66]$ & 36/30 & 0.539 & 0 & $+0.85$ \\
 & 4 & 65.25 & \cellcolor{vizorange!19}$+6.78$ & $[2.54, 10.81]$ & 67/35 & 0.00199 & 0 & $+7.63$ \\
 & 3 & 63.56 & \cellcolor{vizorange!14}$+5.08$ & $[0.85, 9.53]$ & 68/44 & 0.0293 & 0 & $+2.97$ \\
 & rerun 6 & 58.26 & \cellcolor{vizaxis!6}$-0.21$ & -- & 16/17 & 1.0 & -- & $-0.85$ \\
\bottomrule
\end{tabular}
\end{scriptsize}
\end{center}
\end{table*}

\begin{table*}[h]
\caption{OLMoE-0125: all Study 2 conditions. ASR (\%), change against the original model (pp) with 95\% bootstrap CI, forward/reverse flips, exact McNemar $p$ (unadjusted), degenerate forward flips under Llama Guard 3, and the change under the HarmBench classifier; shading: size and sign of the Llama Guard 3 change. rerun: original model with shuffled prompt order.}
\vskip 0.05in
\begin{center}
\begin{scriptsize}
\setlength{\tabcolsep}{3pt}
\begin{tabular}{@{}llrrrrrrr@{}}
\toprule
Framing & $k$ & ASR & $\Delta$ & 95\% CI & Flips & $p$ & Degen. & $\Delta$ HB \\
\midrule
Plain & 8 & 3.39 & -- & -- & -- & -- & -- & -- \\
 & 6 & 2.75 & \cellcolor{vizaxis!6}$-0.64$ & $[-1.91, 0.64]$ & 3/6 & 0.508 & 0 & $+0.42$ \\
 & 5 & 5.30 & \cellcolor{vizorange!6}$+1.91$ & $[0.42, 3.60]$ & 12/3 & 0.0352 & 0 & $+2.33$ \\
 & 4 & 8.26 & \cellcolor{vizorange!13}$+4.87$ & $[2.54, 7.20]$ & 27/4 & $\sci{3.4}{-5}$ & 0 & $+4.24$ \\
 & rerun 8 & 2.97 & \cellcolor{vizaxis!6}$-0.42$ & -- & 3/5 & 0.727 & -- & $-0.64$ \\
\midrule
Refusal supp. & 8 & 24.15 & -- & -- & -- & -- & -- & -- \\
 & 6 & 33.69 & \cellcolor{vizorange!26}$+9.53$ & $[6.36, 12.92]$ & 56/11 & $\sci{2.1}{-8}$ & 0 & $+8.26$ \\
 & 5 & 46.19 & \cellcolor{vizorange!55}$+22.03$ & $[18.01, 26.06]$ & 110/6 & $\sci{7.5}{-26}$ & 0 & $+17.80$ \\
 & 4 & 60.17 & \cellcolor{vizorange!55}$+36.02$ & $[31.57, 40.68]$ & 178/8 & $\sci{6.5}{-43}$ & 0 & $+29.24$ \\
 & rerun 8 & 23.52 & \cellcolor{vizaxis!6}$-0.64$ & -- & 9/12 & 0.664 & -- & $-0.64$ \\
\midrule
Prefix inj. & 8 & 30.93 & -- & -- & -- & -- & -- & -- \\
 & 6 & 36.23 & \cellcolor{vizorange!15}$+5.30$ & $[1.06, 9.53]$ & 64/39 & 0.0176 & 0 & $+5.30$ \\
 & 5 & 31.78 & \cellcolor{vizorange!6}$+0.85$ & $[-3.60, 5.30]$ & 59/55 & 0.779 & 0 & $+0.42$ \\
 & 4 & 39.19 & \cellcolor{vizorange!23}$+8.26$ & $[3.39, 13.14]$ & 91/52 & 0.0014 & 0 & $+9.32$ \\
 & rerun 8 & 33.05 & \cellcolor{vizorange!6}$+2.12$ & -- & 27/17 & 0.174 & -- & $+3.18$ \\
\bottomrule
\end{tabular}
\end{scriptsize}
\end{center}
\end{table*}

\begin{table*}[h]
\caption{Qwen3-30B (FP8): all Study 2 conditions. ASR (\%), change against the original model (pp) with 95\% bootstrap CI, forward/reverse flips, exact McNemar $p$ (unadjusted), degenerate forward flips under Llama Guard 3, and the change under the HarmBench classifier; shading: size and sign of the Llama Guard 3 change. rerun: original model with shuffled prompt order.}
\vskip 0.05in
\begin{center}
\begin{scriptsize}
\setlength{\tabcolsep}{3pt}
\begin{tabular}{@{}llrrrrrrr@{}}
\toprule
Framing & $k$ & ASR & $\Delta$ & 95\% CI & Flips & $p$ & Degen. & $\Delta$ HB \\
\midrule
Plain & 8 & 1.27 & -- & -- & -- & -- & -- & -- \\
 & 6 & 1.27 & $+0.00$ & $[-0.64, 0.64]$ & 1/1 & 1.0 & 0 & $-0.42$ \\
 & 4 & 2.54 & \cellcolor{vizorange!6}$+1.27$ & $[0.00, 2.54]$ & 8/2 & 0.109 & 0 & $+1.48$ \\
 & 2 & 31.57 & \cellcolor{vizorange!55}$+30.30$ & $[26.06, 34.75]$ & 147/4 & $\sci{1.5}{-38}$ & 67 & $+1.48$ \\
 & rerun 8 & 1.27 & $+0.00$ & -- & 1/1 & 1.0 & -- & $-0.21$ \\
\midrule
Refusal supp. & 8 & 9.96 & -- & -- & -- & -- & -- & -- \\
 & 6 & 5.72 & \cellcolor{vizaxis!12}$-4.24$ & $[-6.36, -2.12]$ & 4/24 & $\sci{1.8}{-4}$ & 0 & $-2.54$ \\
 & 4 & 3.18 & \cellcolor{vizaxis!19}$-6.78$ & $[-9.32, -4.45]$ & 2/34 & $\sci{1.9}{-8}$ & 0 & $-3.60$ \\
 & 2 & 34.75 & \cellcolor{vizorange!55}$+24.79$ & $[20.13, 29.66]$ & 139/22 & $\sci{5.7}{-22}$ & 65 & $-4.66$ \\
 & rerun 8 & 10.38 & \cellcolor{vizorange!6}$+0.42$ & -- & 5/3 & 0.727 & -- & $+1.06$ \\
\midrule
Prefix inj. & 8 & 5.51 & -- & -- & -- & -- & -- & -- \\
 & 6 & 1.91 & \cellcolor{vizaxis!10}$-3.60$ & $[-5.51, -1.91]$ & 1/18 & $\sci{7.6}{-5}$ & 0 & $-3.18$ \\
 & 4 & 2.12 & \cellcolor{vizaxis!9}$-3.39$ & $[-5.51, -1.48]$ & 4/20 & 0.00154 & 0 & $-3.81$ \\
 & 2 & 53.18 & \cellcolor{vizorange!55}$+47.67$ & $[42.58, 52.97]$ & 243/18 & $\sci{1.6}{-51}$ & 46 & $-3.81$ \\
 & rerun 8 & 5.51 & $+0.00$ & -- & 6/6 & 1.0 & -- & $+0.21$ \\
\bottomrule
\end{tabular}
\end{scriptsize}
\end{center}
\end{table*}

\begin{table*}[h]
\caption{GPT-OSS-20B: all Study 2 conditions. ASR (\%), change against the original model (pp) with 95\% bootstrap CI, forward/reverse flips, exact McNemar $p$ (unadjusted), degenerate forward flips under Llama Guard 3, and the change under the HarmBench classifier; shading: size and sign of the Llama Guard 3 change. rerun: original model with shuffled prompt order.}
\vskip 0.05in
\begin{center}
\begin{scriptsize}
\setlength{\tabcolsep}{3pt}
\begin{tabular}{@{}llrrrrrrr@{}}
\toprule
Framing & $k$ & ASR & $\Delta$ & 95\% CI & Flips & $p$ & Degen. & $\Delta$ HB \\
\midrule
Plain & 4 & 1.69 & -- & -- & -- & -- & -- & -- \\
 & 3 & 1.06 & \cellcolor{vizaxis!6}$-0.64$ & $[-1.69, 0.42]$ & 2/5 & 0.453 & 0 & $+0.00$ \\
 & 2 & 5.08 & \cellcolor{vizorange!9}$+3.39$ & $[1.69, 5.30]$ & 18/2 & $\sci{4.0}{-4}$ & 1 & $+0.85$ \\
 & rerun 4 & 1.06 & \cellcolor{vizaxis!6}$-0.64$ & -- & 0/3 & 0.25 & -- & $-0.42$ \\
\midrule
Refusal supp. & 4 & 6.57 & -- & -- & -- & -- & -- & -- \\
 & 3 & 5.93 & \cellcolor{vizaxis!6}$-0.64$ & $[-2.54, 1.27]$ & 9/12 & 0.664 & 0 & $+0.00$ \\
 & 2 & 8.26 & \cellcolor{vizorange!6}$+1.69$ & $[-0.64, 4.03]$ & 19/11 & 0.2 & 1 & $+1.69$ \\
 & rerun 4 & 8.26 & \cellcolor{vizorange!6}$+1.69$ & -- & 12/4 & 0.0768 & -- & $+1.06$ \\
\midrule
Prefix inj. & 4 & 0.21 & -- & -- & -- & -- & -- & -- \\
 & 3 & 0.21 & $+0.00$ & $[0.00, 0.00]$ & 0/0 & 1.0 & 0 & $+0.00$ \\
 & 2 & 0.21 & $+0.00$ & $[0.00, 0.00]$ & 0/0 & 1.0 & 0 & $+0.00$ \\
 & rerun 4 & 0.21 & $+0.00$ & -- & 0/0 & 1.0 & -- & $+0.00$ \\
\bottomrule
\end{tabular}
\end{scriptsize}
\end{center}
\end{table*}

\begin{table*}[h]
\caption{DeepSeek-MoE-16B (held out): all Study 2 conditions. ASR (\%), change against the original model (pp) with 95\% bootstrap CI, forward/reverse flips, exact McNemar $p$ (unadjusted), degenerate forward flips under Llama Guard 3, and the change under the HarmBench classifier; shading: size and sign of the Llama Guard 3 change. rerun: original model with shuffled prompt order.}
\vskip 0.05in
\begin{center}
\begin{scriptsize}
\setlength{\tabcolsep}{3pt}
\begin{tabular}{@{}llrrrrrrr@{}}
\toprule
Framing & $k$ & ASR & $\Delta$ & 95\% CI & Flips & $p$ & Degen. & $\Delta$ HB \\
\midrule
Plain & 6 & 18.86 & -- & -- & -- & -- & -- & -- \\
 & 5 & 17.80 & \cellcolor{vizaxis!6}$-1.06$ & $[-3.39, 1.27]$ & 13/18 & 0.473 & 2 & $+0.21$ \\
 & 4 & 18.86 & $+0.00$ & $[-2.97, 2.97]$ & 24/24 & 1.0 & 1 & $+0.00$ \\
 & 3 & 23.94 & \cellcolor{vizorange!14}$+5.08$ & $[1.69, 8.47]$ & 45/21 & 0.00427 & 2 & $+2.54$ \\
 & rerun 6 & 18.22 & \cellcolor{vizaxis!6}$-0.64$ & -- & 9/12 & 0.664 & -- & $+0.85$ \\
\midrule
Refusal supp. & 6 & 37.50 & -- & -- & -- & -- & -- & -- \\
 & 5 & 36.65 & \cellcolor{vizaxis!6}$-0.85$ & $[-4.66, 2.97]$ & 40/44 & 0.744 & 1 & $+0.00$ \\
 & 4 & 36.86 & \cellcolor{vizaxis!6}$-0.64$ & $[-5.08, 3.81]$ & 55/58 & 0.851 & 0 & $-1.91$ \\
 & 3 & 51.27 & \cellcolor{vizorange!38}$+13.77$ & $[8.90, 18.64]$ & 102/37 & $\sci{3.2}{-8}$ & 3 & $+12.71$ \\
 & rerun 6 & 33.90 & \cellcolor{vizaxis!10}$-3.60$ & -- & 3/20 & $\sci{4.9}{-4}$ & -- & $-3.81$ \\
\midrule
Prefix inj. & 6 & 94.49 & -- & -- & -- & -- & -- & -- \\
 & 5 & 93.01 & \cellcolor{vizaxis!6}$-1.48$ & $[-3.39, 0.42]$ & 7/14 & 0.189 & 0 & $+1.06$ \\
 & 4 & 94.07 & \cellcolor{vizaxis!6}$-0.42$ & $[-2.54, 1.69]$ & 11/13 & 0.839 & 0 & $+0.85$ \\
 & 3 & 93.22 & \cellcolor{vizaxis!6}$-1.27$ & $[-3.39, 0.64]$ & 9/15 & 0.307 & 1 & $-1.48$ \\
 & rerun 6 & 94.28 & \cellcolor{vizaxis!6}$-0.21$ & -- & 6/7 & 1.0 & -- & $+0.42$ \\
\bottomrule
\end{tabular}
\end{scriptsize}
\end{center}
\end{table*}

\begin{table*}[h]
\caption{Qwen2-57B-A14B, GPTQ-Int4 (held out): all Study 2 conditions. ASR (\%), change against the original model (pp) with 95\% bootstrap CI, forward/reverse flips, exact McNemar $p$ (unadjusted), degenerate forward flips under Llama Guard 3, and the change under the HarmBench classifier; shading: size and sign of the Llama Guard 3 change. rerun: original model with shuffled prompt order.}
\vskip 0.05in
\begin{center}
\begin{scriptsize}
\setlength{\tabcolsep}{3pt}
\begin{tabular}{@{}llrrrrrrr@{}}
\toprule
Framing & $k$ & ASR & $\Delta$ & 95\% CI & Flips & $p$ & Degen. & $\Delta$ HB \\
\midrule
Plain & 8 & 3.60 & -- & -- & -- & -- & -- & -- \\
 & 6 & 4.03 & \cellcolor{vizorange!6}$+0.42$ & $[-0.64, 1.69]$ & 5/3 & 0.727 & 0 & $+0.42$ \\
 & 5 & 4.66 & \cellcolor{vizorange!6}$+1.06$ & $[-0.21, 2.54]$ & 8/3 & 0.227 & 0 & $+0.21$ \\
 & 4 & 4.24 & \cellcolor{vizorange!6}$+0.64$ & $[-1.06, 2.33]$ & 9/6 & 0.607 & 0 & $+0.85$ \\
 & rerun 8 & 3.60 & $+0.00$ & -- & 0/0 & 1.0 & -- & $+0.00$ \\
\midrule
Refusal supp. & 8 & 45.34 & -- & -- & -- & -- & -- & -- \\
 & 6 & 47.67 & \cellcolor{vizorange!6}$+2.33$ & $[-0.21, 4.87]$ & 26/15 & 0.117 & 0 & $+0.85$ \\
 & 5 & 49.58 & \cellcolor{vizorange!12}$+4.24$ & $[1.48, 7.20]$ & 34/14 & 0.00552 & 0 & $+4.03$ \\
 & 4 & 48.09 & \cellcolor{vizorange!8}$+2.75$ & $[-0.64, 6.14]$ & 41/28 & 0.148 & 1 & $+2.12$ \\
 & rerun 8 & 44.70 & \cellcolor{vizaxis!6}$-0.64$ & -- & 4/7 & 0.549 & -- & $-1.48$ \\
\midrule
Prefix inj. & 8 & 60.59 & -- & -- & -- & -- & -- & -- \\
 & 6 & 63.56 & \cellcolor{vizorange!8}$+2.97$ & $[-1.27, 7.20]$ & 56/42 & 0.189 & 0 & $+4.66$ \\
 & 5 & 59.75 & \cellcolor{vizaxis!6}$-0.85$ & $[-4.87, 3.18]$ & 42/46 & 0.749 & 0 & $-2.12$ \\
 & 4 & 66.10 & \cellcolor{vizorange!15}$+5.51$ & $[1.06, 9.96]$ & 69/43 & 0.0178 & 0 & $+1.91$ \\
 & rerun 8 & 61.23 & \cellcolor{vizorange!6}$+0.64$ & -- & 16/13 & 0.711 & -- & $+1.06$ \\
\bottomrule
\end{tabular}
\end{scriptsize}
\end{center}
\end{table*}
}{}

\subsection{Accuracy}
\label{app:accuracy}

\begin{table}[h]
\caption{Accuracy and over-refusal (\%). XSTest: refusal rate on safe and unsafe prompts. \method: engine implementation with $N=8$ and the fused kernel. MMLU is scored by the first-token answer letter, which \method computes with all experts; GSM8K and MATH-500 (Figure~\ref{fig:math500}) measure its truncated decode steps, and its MMLU differs from the original model's by at most 0.30 points.}
\label{tab:util}
\vskip 0.05in
\begin{center}
\begin{scriptsize}
\setlength{\tabcolsep}{3pt}
\begin{tabular}{@{}llrrrr@{}}
\toprule
 & & & & \multicolumn{2}{c}{XSTest} \\
\cmidrule(l){5-6}
Model & Condition & MMLU & GSM8K & safe & unsafe \\
\midrule
Qwen1.5-MoE & $k=4$ (original) & 58.21 & 58.68 & 24.0 & 97.0 \\
 & $k'=3$ & 57.58 & 56.33 & 18.8 & 95.5 \\
 & $k'=2$ & 56.10 & 50.64 & 21.2 & 89.5 \\
 & \method, $k'=3$ & 58.21 & 57.92 & 24.0 & 97.0 \\
 & \method, $k'=2$ & 57.91 & 54.21 & 24.0 & 97.0 \\
\midrule
DeepSeek-V2-Lite & $k=6$ (original) & 52.76 & 68.54 & 9.6 & 82.0 \\
 & $k'=5$ & 51.89 & 68.46 & 10.4 & 81.0 \\
 & $k'=4$ & 50.94 & 67.02 & 10.8 & 80.0 \\
 & $k'=3$ & 49.98 & 61.79 & 8.4 & 75.0 \\
 & \method, $k'=4$ & 52.76 & 67.10 & 10.0 & 81.5 \\
 & \method, $k'=3$ & 52.75 & 65.66 & 9.6 & 81.0 \\
\midrule
Moonlight-16B & $k=6$ (original) & 66.93 & 85.90 & 8.4 & 14.5 \\
 & $k'=5$ & 66.66 & 85.67 & 7.6 & 13.5 \\
 & $k'=4$ & 66.14 & 83.24 & 7.6 & 11.0 \\
 & $k'=3$ & 62.45 & 72.33 & 5.2 & 12.5 \\
\midrule
Ling-lite-1.5 & $k=6$ (original) & 69.92 & 91.05 & 2.0 & 47.5 \\
 & $k'=5$ & 69.43 & 89.61 & 2.0 & 52.5 \\
 & $k'=4$ & 68.28 & 88.48 & 1.6 & 51.0 \\
 & $k'=3$ & 64.73 & 84.31 & 2.4 & 51.0 \\
\midrule
OLMoE-0125 & $k=8$ (original) & 51.93 & 70.81 & 8.8 & 83.0 \\
 & $k'=6$ & 51.27 & 68.16 & 9.6 & 85.0 \\
 & $k'=5$ & 50.51 & 67.93 & 10.0 & 84.5 \\
 & $k'=4$ & 48.33 & 61.87 & 12.0 & 83.5 \\
 & \method, $k'=6$ & 51.93 & 71.57 & 8.8 & 83.0 \\
 & \method, $k'=4$ & 51.93 & 65.58 & 8.8 & 83.0 \\
\midrule
Qwen3-30B & $k=8$ (original) & 79.67 & 92.72 & 7.2 & 69.0 \\
 & $k'=6$ & 77.90 & 92.04 & 8.0 & 68.5 \\
 & $k'=4$ & 74.90 & 84.53 & 7.2 & 64.5 \\
 & $k'=2$ & 40.61 & 3.11 & 4.0 & 14.0 \\
\midrule
DeepSeek-MoE-16B & $k=6$ (original) & 45.21 & 54.13 & 12.0 & 82.5 \\
 & $k'=4$ & 43.94 & 50.64 & 14.4 & 79.0 \\
\midrule
Qwen2-57B-A14B, & $k=8$ (original) & 73.05 & 85.75 & 13.2 & 88.0 \\
GPTQ-Int4 & $k'=6$ & 72.94 & 86.50 & 15.2 & 89.5 \\
\bottomrule
\end{tabular}
\end{scriptsize}
\end{center}
\end{table}

\textbf{Quantization control.} Qwen1.5-MoE-A2.7B-Chat-GPTQ-Int4 (snapshot \texttt{81b132ad}), run like the original model, changes ASR against the BF16 original model by $+0.85$\pp on plain requests ($p=0.58$), $-8.47$\pp under refusal suppression (17 against 57 discordant prompts, $p=3.4\times10^{-6}$; HarmBench classifier $-5.72$\pp), and $-11.02$\pp under prefix injection ($p<10^{-6}$), while MMLU falls from 58.21 to 56.74\%, GSM8K from 58.68 to 55.65\%, and XSTest refusal rates change from 24.0 and 97.0\% to 23.6 and 96.5\%.

\begin{figure}[h]
\centering
\includegraphics[width=\columnwidth]{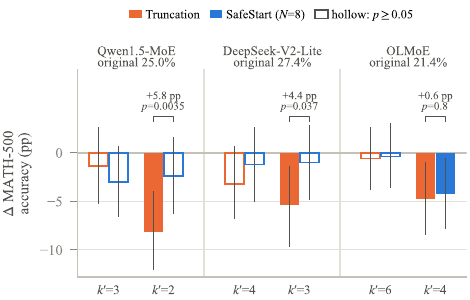}
\caption{MATH-500 accuracy change against the original model (500 problems, greedy, at most 1,024 tokens, scored with math-verify) for truncation (orange) and \method with $N=8$ in the engine with the fused kernel (blue), at $k'=\lceil 2k/3\rceil$ (left pair of each model) and at half the experts (right pair). Whiskers: 95\% paired bootstrap intervals over problems; hollow: exact McNemar $p\ge0.05$ against the original model. Brackets: \method minus truncation at half the experts on the same problems, with its exact McNemar $p$.}
\label{fig:math500}
\end{figure}

\subsection{Other Reduction Methods}
\label{app:rules}

\textbf{Other reduction rules.} By token position (scripts/position\_experts.py, with the calibration of Table~3 of the paper: the original model's top-$k$ router weights, teacher-forced on its own refusal-suppression responses for 120 prompts, with each rule applied to them), threshold skipping keeps 3.15, 4.45, and 6.15 experts per prompt token on Qwen1.5, DeepSeek-V2-Lite, and OLMoE, 3.37, 4.14, and 7.24 at the final prompt token, and 2.76, 3.97, and 6.34 at generated tokens 1 to 8; top-$p$ routing keeps 3.24, 4.37, and 6.42, 2.39, 3.20, and 6.06, and 2.65, 3.59, and 6.20. Table~\ref{tab:rule_timing} gives their decode speedups (on OLMoE, top-$p$ routing with protection reads 3.1, 1.5, and 1.0\% fewer distinct experts per decode step than without it at batch sizes 8, 32, and 128, which at truncation's ratio of time saved to reads removed predicts 2.3, 1.1, and 0.6\% shorter decode time, against 2.9, 1.0, and 0.5\% measured) and Table~\ref{tab:methods_hb} gives every rule with accuracy, over-refusal, and both judges. Threshold skipping keeps expert $i$ when its gate weight is at least $\beta w_1$; top-$p$ routing keeps experts in weight order while the mass before them is below $p$. We fix $\beta$ and $p$ by linear interpolation on a grid so that the average number of experts per token, over prompt and generated tokens, equals $k'$ on a calibration run of the first 40 test prompts in each framing (64 generated tokens); calibration sets only the expert count, not a safety outcome. The jailbreak templates dominate that average, so on decode tokens the rules keep fewer experts than $k'$, except top-$p$ on DeepSeek-V2-Lite (the Experts column, measured on GSM8K requests); for the three rules furthest below $k'$ (both on OLMoE, threshold skipping on DeepSeek-V2-Lite) we recalibrate so that the decode tokens of 120 held-out Alpaca requests (128 generated tokens) keep $k'$ experts on average, and the main-text table uses these decode-matched rules. On Qwen1.5 the rules keep 2.8 to 2.9 of 3 and were not recalibrated. Under the first calibration, threshold skipping raised ASR on OLMoE by 15.0\pp at 4.8 experts per decode token, near truncation to 5 experts ($+22.0$\pp). At batch size 8 a step reads 25.3 and 27.9 distinct experts per layer on OLMoE under the decode-matched threshold and top-$p$ rules, against 24.2 under truncation and 29.5 for the original model, because tokens with flat gates keep more experts. A coarser first grid gave $p=0.30$, which keeps 3.5 and 6.6 experts per token over the calibration run's prompt and generated tokens on Qwen1.5 and OLMoE; we report it as a lighter setting. Both rules zero the dropped experts' weights without renormalizing, as top-$k$ truncation does on these three models, none of which renormalizes. We also tried static pruning with REAP \citep{lasby2025reap}, scoring each expert by its mean router-weighted output norm over the tokens routed to it on 512 Alpaca conversations and masking the router logits of the lowest-scoring 25\% in every layer. It lost 30.0 and 51.9 GSM8K points on Qwen1.5 and OLMoE, far more than REAP's authors report for larger models with task-matched calibration, so we could not validate our setting and leave REAP out of the comparison. With \method, top-$p$ routing decodes faster than alone on DeepSeek-V2-Lite and OLMoE (Figure~1 of the paper): the outputs differ, and with top-$p$ the experts a step reads depend on them; on OLMoE over 256-token outputs at batch size 8, a step reads 29.0 distinct experts per layer with protection and 29.9 without. \method for the dynamic rules runs in the engine: after the protected tokens, the fused kernel applies the rule to the original top-$k$ slots (the PyTorch path for the first-calibration rows of DeepSeek-V2-Lite threshold skipping and OLMoE). With every token reduced, the engine path reproduces the rule's reference implementation on 100 prompts in each framing: identical outputs for top-$p$ routing on all three models, and the same first 40 characters on 296, 295, and 300 of 300 responses for threshold skipping on Qwen1.5, DeepSeek-V2-Lite, and OLMoE.

\begin{table*}[h]
\caption{Expert-reduction rules in full. MMLU, GSM8K, and XSTest over-refusal (refusal rate on safe prompts) in \%; ASR change under refusal suppression in pp against the original row (original rows give the absolute ASR), under Llama Guard 3 (LG) and the HarmBench classifier (HB), alone and with \method with $N=8$: Eng., in the engine (all experts for the prompt and the first eight generated tokens; fused kernel where run, PyTorch path otherwise); Emul., in emulation (the original model's first eight tokens, then the rule). Bold: exact McNemar $p<0.05$; shading: size and sign of the Llama Guard 3 change. Experts: experts per token; for the dynamic rules, the mean per decode token on GSM8K requests at batch size 8. The first calibration of each dynamic rule averages over prompt and generated tokens of 40 evaluation prompts per framing; the decode-matched rows are recalibrated on 120 held-out Alpaca requests so that decode tokens keep $k'$ experts on average; $\approx$ marks counts for which only this calibration-run average is available.}
\label{tab:methods_hb}
\vskip 0.05in
\begin{center}
\begin{scriptsize}
\setlength{\tabcolsep}{3pt}
\begin{tabular}{@{}llrrrrrrrrrr@{}}
\toprule
 & & & & & & \multicolumn{2}{c}{\textcolor{vizorange}{Alone}} & \multicolumn{2}{c}{\textcolor{vizblue}{Eng.}} & \multicolumn{2}{c}{\textcolor{vizblue}{Emul.}} \\
\cmidrule(lr){7-8}\cmidrule(lr){9-10}\cmidrule(l){11-12}
Model & Rule & Experts & MMLU & GSM8K & XSTest & LG & HB & LG & HB & LG & HB \\
\midrule
Qwen1.5 & Original & 4 & 58.2 & 58.7 & 24.0 & 46.6 & 40.5 &  &  &  &  \\
 & Top-$k$ truncation & 3 & 57.6 & 56.3 & 18.8 & \cellcolor{vizorange!51}$\mathbf{+18.4}$ & $\mathbf{+16.5}$ & \cellcolor{vizaxis!6}$-0.6$ & $+0.4$ & \cellcolor{vizorange!6}$+1.1$ & $+1.3$ \\
 & Top-$k$ truncation & 2 & 56.1 & 50.6 & 21.2 & \cellcolor{vizorange!55}$\mathbf{+22.2}$ & $\mathbf{+22.2}$ & \cellcolor{vizorange!6}$+0.6$ & $+0.6$ & \cellcolor{vizaxis!6}$-0.4$ & $+0.6$ \\
 & Threshold, $\beta=0.41$ & 2.9 & 57.7 & 57.2 & 25.2 & \cellcolor{vizaxis!6}$-1.9$ & $-3.4$ & $+0.0$ & $+0.6$ & \cellcolor{vizorange!6}$+0.2$ & $+1.5$ \\
 & Top-$p$, $p=0.22$ & 2.9 & 55.6 & 50.8 & 20.8 & \cellcolor{vizorange!55}$\mathbf{+30.1}$ & $\mathbf{+24.8}$ & \cellcolor{vizaxis!6}$-2.1$ & $-0.2$ & $+0.0$ & $+0.0$ \\
 & Top-$p$, $p=0.30$ & $\approx$3.5 & 57.0 & 54.7 & 20.8 & \cellcolor{vizorange!55}$\mathbf{+20.6}$ & $\mathbf{+18.4}$ & -- & -- & \cellcolor{vizaxis!6}$-0.6$ & $+1.7$ \\
\midrule
DeepSeek-V2-Lite & Original & 6 & 52.8 & 68.5 & 9.6 & 43.6 & 35.2 &  &  &  &  \\
 & Top-$k$ truncation & 4 & 50.9 & 67.0 & 10.8 & \cellcolor{vizorange!18}$\mathbf{+6.6}$ & $\mathbf{+5.3}$ & \cellcolor{vizorange!6}$+0.6$ & $+2.3$ & \cellcolor{vizorange!6}$+1.3$ & $+0.4$ \\
 & Top-$k$ truncation & 3 & 50.0 & 61.8 & 8.4 & \cellcolor{vizorange!37}$\mathbf{+13.3}$ & $\mathbf{+10.6}$ & \cellcolor{vizaxis!6}$-0.6$ & $+0.0$ & \cellcolor{vizorange!6}$+1.7$ & $+1.3$ \\
 & Threshold, $\beta=0.43$ & 3.7 & 50.4 & 66.3 & 6.8 & \cellcolor{vizorange!6}$+1.7$ & $+3.6$ & \cellcolor{vizorange!6}$+0.8$ & $+0.6$ & \cellcolor{vizaxis!6}$-0.2$ & $-0.6$ \\
 & Top-$p$, $p=0.27$ & 4.1 & 49.0 & 64.3 & 9.2 & \cellcolor{vizorange!31}$\mathbf{+11.2}$ & $\mathbf{+7.8}$ & \cellcolor{vizorange!6}$+1.3$ & $+1.5$ & \cellcolor{vizorange!6}$+1.5$ & $+2.5$ \\
 & Threshold, decode-matched, $\beta=0.38$ & 4.2 & 51.0 & 67.2 & 9.2 & \cellcolor{vizorange!6}$+1.3$ & $+2.5$ & \cellcolor{vizaxis!6}$-0.6$ & $+0.6$ & -- & -- \\
\midrule
OLMoE & Original & 8 & 51.9 & 70.8 & 8.8 & 24.1 & 18.2 &  &  &  &  \\
 & Top-$k$ truncation & 6 & 51.3 & 68.2 & 9.6 & \cellcolor{vizorange!26}$\mathbf{+9.5}$ & $\mathbf{+8.3}$ & \cellcolor{vizorange!6}$+1.3$ & $+1.7$ & \cellcolor{vizorange!9}$\mathbf{+3.4}$ & $+2.5$ \\
 & Top-$k$ truncation & 5 & 50.5 & 67.9 & 10.0 & \cellcolor{vizorange!55}$\mathbf{+22.0}$ & $\mathbf{+17.8}$ & -- & -- & -- & -- \\
 & Top-$k$ truncation & 4 & 48.3 & 61.9 & 12.0 & \cellcolor{vizorange!55}$\mathbf{+36.0}$ & $\mathbf{+29.2}$ & \cellcolor{vizorange!6}$+0.4$ & $+1.9$ & -- & -- \\
 & Threshold, $\beta=0.48$ & 4.8 & 48.9 & 64.8 & 11.6 & \cellcolor{vizorange!41}$\mathbf{+15.0}$ & $\mathbf{+15.0}$ & \cellcolor{vizorange!6}$+0.6$ & $-0.2$ & \cellcolor{vizorange!8}$\mathbf{+2.8}$ & $+2.5$ \\
 & Top-$p$, $p=0.26$ & 4.8 & 48.3 & 54.1 & 24.4 & \cellcolor{vizorange!55}$\mathbf{+26.7}$ & $\mathbf{+23.1}$ & \cellcolor{vizorange!10}$\mathbf{+3.6}$ & $+2.5$ & \cellcolor{vizorange!13}$\mathbf{+4.9}$ & $\mathbf{+5.3}$ \\
 & Threshold, decode-matched, $\beta=0.37$ & 6.3 & 50.1 & 70.0 & 10.0 & \cellcolor{vizorange!6}$+1.9$ & $+1.7$ & \cellcolor{vizorange!6}$+1.3$ & $-0.2$ & -- & -- \\
 & Top-$p$, decode-matched, $p=0.33$ & 6.3 & 50.5 & 65.6 & 14.8 & \cellcolor{vizorange!24}$\mathbf{+8.9}$ & $\mathbf{+8.3}$ & $+0.0$ & $+1.1$ & -- & -- \\
 & Top-$p$, $p=0.30$ & $\approx$6.6 & 49.9 & 63.1 & 19.2 & \cellcolor{vizorange!36}$\mathbf{+12.9}$ & $\mathbf{+11.0}$ & -- & -- & \cellcolor{vizorange!9}$\mathbf{+3.4}$ & $\mathbf{+3.0}$ \\
\bottomrule
\end{tabular}
\end{scriptsize}
\end{center}
\end{table*}

\begin{table}[h]
\caption{Decode speedup over the original model (\%) for the reduction rules of Table~3 of the paper, alone and with \method ($N=8$, engine), against truncation at $k'=\lceil 2k/3\rceil$ and the original model timed in the same interleaved job (on DeepSeek-V2-Lite one job per rule, each with its own truncation row; two rounds, 256 output tokens, one RTX A6000).}
\label{tab:rule_timing}
\vskip 0.05in
\begin{center}
\begin{scriptsize}
\setlength{\tabcolsep}{3pt}
\begin{tabular}{llrrrr}
\toprule
 & & \multicolumn{4}{c}{Batch size} \\
\cmidrule(l){3-6}
Model & Condition & 1 & 8 & 32 & 128 \\
\midrule
Qwen1.5 & Truncation & $+7.2$ & $+15.9$ & $+8.1$ & $+0.7$ \\
 & Threshold & $+5.0$ & $+15.5$ & $+8.6$ & $+0.5$ \\
 & Threshold + \method & $+4.8$ & $+13.6$ & $+7.8$ & $+0.3$ \\
 & Top-$p$ & $+4.8$ & $+14.3$ & $+8.1$ & $+0.5$ \\
 & Top-$p$ + \method & $+5.4$ & $+15.0$ & $+8.5$ & $+0.4$ \\
\midrule
DeepSeek-V2-Lite & Truncation & $-0.8$ & $+27.4$ & $+11.3$ & $+2.1$ \\
 & Threshold & $+0.2$ & $+21.3$ & $+9.2$ & $+1.6$ \\
 & Threshold + \method & $-0.4$ & $+21.2$ & $+9.6$ & $+1.8$ \\
\addlinespace
 & Truncation & $+0.1$ & $+27.1$ & $+11.3$ & $+2.3$ \\
 & Top-$p$ & $+0.3$ & $+15.7$ & $+8.5$ & $+1.6$ \\
 & Top-$p$ + \method & $+0.0$ & $+19.2$ & $+9.2$ & $+1.7$ \\
\midrule
OLMoE & Truncation & $+13.7$ & $+16.2$ & $+11.5$ & $+4.1$ \\
 & Threshold & $+9.7$ & $+14.0$ & $+8.0$ & $+2.3$ \\
 & Threshold + \method & $+9.7$ & $+13.4$ & $+7.8$ & $+2.3$ \\
 & Top-$p$ & $+7.8$ & $+6.5$ & $+4.4$ & $+1.3$ \\
 & Top-$p$ + \method & $+7.3$ & $+9.7$ & $+5.5$ & $+1.8$ \\
\bottomrule
\end{tabular}
\end{scriptsize}
\end{center}
\end{table}

\subsection{\method: Further Results}
\label{app:safestart}

\textbf{Evaluation prompts.} With $N=8$ under refusal suppression and Llama Guard 3, \method changes ASR by $-0.64$, $+0.64$, and $+1.27$\pp on Qwen1.5, DeepSeek-V2-Lite, and OLMoE, with 90\% Tango intervals $[-2.55, 1.23]$, $[-1.64, 2.94]$, and $[-0.60, 3.23]$; the first two lie within the $\pm3$\pp margin. The HarmBench classifier gives $+0.42$, $+2.33$, and $+1.69$\pp, none significant. At half the experts, where truncation adds $+22.25$, $+13.35$, and $+36.02$\pp, $N=8$ leaves $+0.64$, $-0.64$, and $+0.42$\pp, none significant under either judge (90\% CI upper bounds 2.7, 1.9, and 2.5\pp).

\textbf{Emulation and other runs.} To test many settings cheaply, we also emulate the policy: the original model decodes the first $N$ tokens and the truncated model continues, recomputing the prompt and those tokens with $k'$ experts, so the emulation shows that the original opening suffices, not which positions need full computation. With the original model in both passes it changes ASR by at most 1.91\pp in any framing on the three models; one of these nine controls is significant (OLMoE on plain requests, $-1.48$\pp, $p=0.016$). At the recommended point the emulation agrees with the engine on Qwen1.5 and DeepSeek-V2-Lite but leaves $+3.39$\pp on OLMoE ($p=0.007$), because it recomputes the opening tokens with truncated experts. In emulation we ran $N\in\{1,2,8,32,64\}$ and chose $N=8$ from these runs and the divergence curves on the evaluation prompts; the engine sweep of Table~\ref{tab:nsweep} ran afterwards. The policy removes the evil-confidant increase on DeepSeek-V2-Lite ($+13.14$ to $+0.85$\pp in emulation and $-0.42$\pp in the engine, HarmBench classifier $+11.44$ to $-1.48$\pp). Under refusal suppression on Qwen1.5, 83.1\% of generated tokens run truncated at $N=32$ in the engine and 66.7\% at $N=64$ in emulation. With the PyTorch version of the patch, which selects the same experts and differs from the kernel in floating-point rounding, $N=8$ changes refusal-suppression ASR by $-1.06$, $+0.85$, and $+1.48$\pp on Qwen1.5, DeepSeek-V2-Lite, and OLMoE (HarmBench classifier $-1.69$, $-0.85$, and $+0.85$\pp). Under prefix injection it gives $+3.81$ ($p=0.015$), $+0.85$, and $+3.18$\pp at $N=8$ and $+1.27$, $+2.75$, and $+2.75$\pp at $N=32$ (HarmBench classifier $+4.66$, $+2.75$, $+2.12$ and $0.00$, $+3.39$, $+3.39$\pp), against $+1.27$, $+0.21$, $+4.45$ and $+2.12$, $+1.69$, $-1.06$\pp with the kernel. A second kernel run of $N=8$ with the requests in a different order gives, on Qwen1.5, DeepSeek-V2-Lite, and OLMoE, $-0.64$, $+0.42$, and $+1.27$\pp under refusal suppression (first run: $-0.64$, $+0.64$, and $+1.27$), $+2.75$, $+1.27$, and $+1.27$\pp under prefix injection ($+1.27$, $+0.21$, and $+4.45$), and $-0.64$, $0.00$, and $-0.85$\pp on plain requests ($-0.21$, $+1.27$, and $-0.85$); HarmBench classifier $-0.64$, $-1.69$, and $+1.06$, $+3.60$, $+1.69$, and $+2.12$, and $+1.06$, $-0.42$, and $-0.42$\pp; the spread between the two implementations is the run-to-run variation that rounding causes. On Ling-lite ($k'=4$ of 6, fused kernel), $N=8$ changes ASR by $-0.21$, $-1.69$, and $+12.50$\pp on plain requests, refusal suppression, and prefix injection ($p=1.2\times10^{-9}$; HarmBench classifier $+0.64$, $-1.91$, and $+12.71$\pp) and $N=32$ by $+0.42$, $-0.85$, and $+1.48$\pp ($+0.42$, $-1.06$, and $+0.85$\pp), against $+0.85$, $+0.85$, and $+6.78$\pp for truncation. At half the experts, $N=8$ changes ASR by $+1.27$, $+1.06$, and $0.00$\pp on plain requests and $+0.85$, $+2.33$, and $+1.69$\pp under prefix injection on Qwen1.5, DeepSeek-V2-Lite, and OLMoE, against $+5.08$, $+10.59$, $+4.87$ and $+5.93$, $+15.89$, $+8.26$\pp for truncation at that budget (HarmBench classifier $+1.91$, $+0.21$, $+0.42$ and $-1.69$, $+3.81$, $+0.42$\pp). On DeepSeek-V2-Lite the patch with nothing truncated reproduces the original model's responses exactly under both adversarial templates (944 of 944 identical), so Table~\ref{tab:adv} shows no flips for it. Under sampled decoding at temperature 0.7 (Table~\ref{tab:sampling}), \method with $N=8$ in the engine changes refusal-suppression ASR on Qwen1.5 by $+0.42$, $-1.69$, and $-0.85$\pp across three seeds (HarmBench classifier $+1.48$, $-0.42$, and $+1.27$\pp; none significant under either judge), against $+16.53$ to $+17.58$\pp for truncation; in emulation it gave $-0.42$, $+2.54$, and $+0.64$\pp (HarmBench classifier $+2.5$, $+2.5$, and $+3.8$\pp, the last with $p=0.018$). Under prefix injection on Qwen1.5, $N=64$ in emulation leaves $+2.33$\pp ($p=0.071$), and $+1.06$\pp with the forced opening removed before judging. Table~\ref{tab:util} gives accuracy and over-refusal for the eight models with accuracy results and for \method. With the fused kernel, GSM8K recovers between 5\% and all of truncation's loss at the confirmatory point and 41 to 57\% at half the experts, because most reasoning tokens come after the protected prefix.

\begin{table}[h]
\caption{Protection length $N$ in the engine with the fused kernel: change in ASR against the original model (pp; refusal suppression RS and prefix injection PI; Llama Guard 3 LG and HarmBench classifier HB; bold: McNemar $p<0.05$; shading: size and sign of the LG change) and decode speedup over the original model (\%, 256 output tokens; all conditions of a model timed in one job, interleaved over two rounds; on DeepSeek-V2-Lite the batch-size-1 values are within timing noise). $N=0$ runs every prompt token but the last at full $k$; trunc.: truncation of every token; open.: opening only (Figure~4 of the paper), every prompt token but the last at $k'$ and the last with the first eight generated tokens at full $k$ (not timed).}
\label{tab:nsweep}
\vskip 0.05in
\begin{center}
\begin{scriptsize}
\setlength{\tabcolsep}{2.5pt}
\begin{tabular}{@{}lrrrrrrr@{}}
\toprule
 & \multicolumn{2}{c}{RS} & \multicolumn{2}{c}{PI} & \multicolumn{3}{c}{Speedup (\%)} \\
\cmidrule(lr){2-3}\cmidrule(lr){4-5}\cmidrule(l){6-8}
$N$ & LG & HB & LG & HB & bs 1 & bs 8 & bs 32 \\
\midrule
\multicolumn{8}{@{}l}{\emph{Qwen1.5}} \\
trunc. & \cellcolor{vizorange!51}$\mathbf{+18.4}$ & $\mathbf{+16.5}$ & \cellcolor{vizorange!19}$\mathbf{+6.8}$ & $\mathbf{+7.0}$ & $+7.6$ & $+16.0$ & $+8.1$ \\
0 & \cellcolor{vizorange!25}$\mathbf{+9.1}$ & $\mathbf{+7.0}$ & \cellcolor{vizorange!17}$\mathbf{+6.1}$ & $\mathbf{+7.6}$ & $+5.1$ & $+14.2$ & $+7.6$ \\
1 & \cellcolor{vizorange!6}$+0.4$ & $-0.8$ & \cellcolor{vizorange!6}$+1.5$ & $+2.5$ & $+5.1$ & $+15.4$ & $+7.9$ \\
2 & \cellcolor{vizaxis!6}$-1.3$ & $-0.8$ & \cellcolor{vizorange!6}$+2.1$ & $+1.3$ & $+5.1$ & $+15.0$ & $+8.1$ \\
4 & \cellcolor{vizaxis!6}$-1.3$ & $-0.4$ & \cellcolor{vizorange!7}$+2.5$ & $+3.2$ & $+5.1$ & $+14.5$ & $+7.8$ \\
8 & \cellcolor{vizaxis!6}$-0.6$ & $+0.4$ & \cellcolor{vizorange!6}$+1.3$ & $+3.0$ & $+4.9$ & $+14.7$ & $+7.2$ \\
16 & \cellcolor{vizaxis!8}$\mathbf{-2.8}$ & $-1.1$ & \cellcolor{vizorange!6}$+1.7$ & $+1.7$ & $+4.8$ & $+13.8$ & $+7.6$ \\
32 & \cellcolor{vizaxis!6}$-1.5$ & $-1.1$ & \cellcolor{vizorange!6}$+2.1$ & $+0.6$ & $+4.1$ & $+12.6$ & $+6.4$ \\
open. & \cellcolor{vizorange!19}$\mathbf{+7.0}$ & $\mathbf{+6.8}$ & \cellcolor{vizorange!10}$\mathbf{+3.6}$ & $\mathbf{+5.1}$ & -- & -- & -- \\
\midrule
\multicolumn{8}{@{}l}{\emph{DeepSeek-V2-Lite}} \\
trunc. & \cellcolor{vizorange!18}$\mathbf{+6.6}$ & $\mathbf{+5.3}$ & \cellcolor{vizaxis!6}$-2.3$ & $\mathbf{-5.1}$ & $-1.7$ & $+27.3$ & $+11.3$ \\
0 & \cellcolor{vizorange!8}$+2.8$ & $+3.2$ & \cellcolor{vizorange!23}$\mathbf{+8.3}$ & $\mathbf{+7.6}$ & $-1.4$ & $+26.8$ & $+11.2$ \\
1 & \cellcolor{vizorange!6}$+0.8$ & $+2.5$ & \cellcolor{vizorange!6}$+0.2$ & $+2.5$ & $-1.6$ & $+26.0$ & $+11.5$ \\
2 & \cellcolor{vizorange!6}$+2.1$ & $+2.3$ & \cellcolor{vizorange!6}$+1.1$ & $+0.6$ & $-1.6$ & $+26.3$ & $+13.1$ \\
4 & \cellcolor{vizaxis!6}$-0.6$ & $+0.8$ & \cellcolor{vizaxis!6}$-1.1$ & $-0.6$ & $-0.2$ & $+24.7$ & $+12.2$ \\
8 & \cellcolor{vizorange!6}$+0.6$ & $+2.3$ & \cellcolor{vizorange!6}$+0.2$ & $+0.2$ & $-0.8$ & $+23.6$ & $+11.9$ \\
16 & \cellcolor{vizorange!6}$+1.9$ & $+1.7$ & \cellcolor{vizaxis!6}$-1.3$ & $+0.0$ & $+0.2$ & $+22.6$ & $+11.0$ \\
32 & \cellcolor{vizorange!6}$+1.1$ & $+1.5$ & \cellcolor{vizorange!6}$+1.7$ & $+0.8$ & $+0.4$ & $+21.1$ & $+9.6$ \\
open. & \cellcolor{vizorange!6}$+2.3$ & $+2.8$ & \cellcolor{vizaxis!26}$\mathbf{-9.5}$ & $\mathbf{-9.1}$ & -- & -- & -- \\
\midrule
\multicolumn{8}{@{}l}{\emph{OLMoE}} \\
trunc. & \cellcolor{vizorange!26}$\mathbf{+9.5}$ & $\mathbf{+8.3}$ & \cellcolor{vizorange!15}$\mathbf{+5.3}$ & $\mathbf{+5.3}$ & $+13.9$ & $+16.2$ & $+11.4$ \\
0 & \cellcolor{vizaxis!6}$-0.6$ & $+0.2$ & \cellcolor{vizorange!9}$+3.4$ & $+3.0$ & $+12.1$ & $+16.1$ & $+10.7$ \\
1 & $+0.0$ & $\mathbf{+2.8}$ & \cellcolor{vizorange!17}$\mathbf{+6.1}$ & $\mathbf{+4.0}$ & $+11.9$ & $+15.7$ & $+11.1$ \\
2 & \cellcolor{vizorange!6}$+1.5$ & $+2.5$ & \cellcolor{vizorange!12}$+4.2$ & $\mathbf{+4.9}$ & $+11.9$ & $+16.2$ & $+10.2$ \\
4 & \cellcolor{vizorange!6}$+0.4$ & $+0.2$ & \cellcolor{vizorange!6}$+1.9$ & $\mathbf{+3.8}$ & $+11.7$ & $+15.4$ & $+10.1$ \\
8 & \cellcolor{vizorange!6}$+1.3$ & $+1.7$ & \cellcolor{vizorange!12}$\mathbf{+4.4}$ & $+3.2$ & $+11.6$ & $+15.0$ & $+9.7$ \\
16 & \cellcolor{vizaxis!6}$-0.2$ & $+0.6$ & \cellcolor{vizorange!13}$\mathbf{+4.9}$ & $\mathbf{+5.9}$ & $+11.1$ & $+15.6$ & $+9.7$ \\
32 & $+0.0$ & $+1.1$ & \cellcolor{vizaxis!6}$-1.1$ & $-0.8$ & $+10.4$ & $+13.3$ & $+8.5$ \\
open. & \cellcolor{vizorange!24}$\mathbf{+8.9}$ & $\mathbf{+8.9}$ & \cellcolor{vizorange!26}$\mathbf{+9.3}$ & $\mathbf{+7.6}$ & -- & -- & -- \\
\bottomrule
\end{tabular}
\end{scriptsize}
\end{center}
\end{table}

\IfFileExists{tables/app_mitig_pi.tex}{\begin{table*}[t]
\caption{Mechanism and mitigation conditions at the two-thirds operating point (Study 2, 472 prompts per framing). ASR change in pp against the original model under Llama Guard 3, with the HarmBench classifier in parentheses; bold: exact McNemar $p<0.05$ under Llama Guard 3 (unadjusted, exploratory); shading: size and sign of the Llama Guard 3 change. Attenuation only keeps all $k$ experts and scales their weights by $m_{k'}/m_k$; mass restored keeps the top $k'$ experts and rescales them to $m_k$; swap keeps $k$ experts but substitutes the $(k{+}1)$-th ranked expert for the $k$-th at the same weight. Early: truncated model for the first generated token only, original afterwards. \method: original computation for the first $N$ generated tokens, in emulation (the truncated model recomputes the prompt and the first $N$ tokens) or in the engine with the fused kernel (Table~\ref{tab:nsweep} gives every $N$). Emulation control: the \method procedure with the original model in both passes. Prefix-injection framing; Table~\ref{tab:mitig} gives the other framings.}
\label{tab:mitigpi}
\vskip 0.1in
\begin{center}
\begin{small}
\setlength{\tabcolsep}{3.5pt}
\begin{tabular}{@{}lrrr}
\toprule
 & \multicolumn{3}{c}{Prefix injection} \\
\cmidrule(l){2-4}
Condition & \multicolumn{1}{c}{Qwen1.5, $k'=3/4$} & \multicolumn{1}{c}{DeepSeek-V2-Lite, $k'=4/6$} & \multicolumn{1}{c}{OLMoE, $k'=6/8$} \\
\midrule
Truncation & \cellcolor{vizorange!19}$\mathbf{+6.8}$ ($+7.0$) & \cellcolor{vizaxis!6}$-2.3$ ($-5.1$) & \cellcolor{vizorange!15}$\mathbf{+5.3}$ ($+5.3$) \\
\addlinespace
Attenuation only & \cellcolor{vizorange!12}$\mathbf{+4.4}$ ($+3.8$) & \cellcolor{vizorange!18}$\mathbf{+6.6}$ ($+5.5$) & \cellcolor{vizorange!15}$\mathbf{+5.5}$ ($+4.9$) \\
Mass restored & \cellcolor{vizorange!11}$\mathbf{+4.0}$ ($+5.3$) & \cellcolor{vizaxis!16}$\mathbf{-5.9}$ ($-6.1$) & \cellcolor{vizorange!6}$+1.1$ ($-1.9$) \\
Swap (no truncation) & \cellcolor{vizorange!12}$\mathbf{+4.4}$ ($+6.6$) & \cellcolor{vizaxis!15}$\mathbf{-5.5}$ ($-3.8$) & \cellcolor{vizorange!17}$\mathbf{+6.4}$ ($+5.7$) \\
\addlinespace
Early, $N=1$ & \cellcolor{vizorange!6}$+2.3$ ($+2.5$) & \cellcolor{vizorange!12}$\mathbf{+4.2}$ ($+4.7$) & $+0.0$ ($-0.2$) \\
\method, $N=1$, emulation & \cellcolor{vizorange!8}$+2.8$ ($+2.3$) & \cellcolor{vizaxis!15}$\mathbf{-5.3}$ ($-5.7$) & \cellcolor{vizorange!16}$\mathbf{+5.9}$ ($+5.5$) \\
\method, $N=1$, engine & \cellcolor{vizorange!6}$+1.5$ ($+2.5$) & \cellcolor{vizorange!6}$+0.2$ ($+2.5$) & \cellcolor{vizorange!17}$\mathbf{+6.1}$ ($+4.0$) \\
\method, $N=8$, emulation & \cellcolor{vizorange!11}$\mathbf{+4.0}$ ($+4.7$) & \cellcolor{vizaxis!15}$\mathbf{-5.3}$ ($-7.4$) & \cellcolor{vizorange!10}$+3.6$ ($+6.1$) \\
\method, $N=8$, engine & \cellcolor{vizorange!6}$+1.3$ ($+3.0$) & \cellcolor{vizorange!6}$+0.2$ ($+0.2$) & \cellcolor{vizorange!12}$\mathbf{+4.4}$ ($+3.2$) \\
\method, $N=32$, engine & \cellcolor{vizorange!6}$+2.1$ ($+0.6$) & \cellcolor{vizorange!6}$+1.7$ ($+0.8$) & \cellcolor{vizaxis!6}$-1.1$ ($-0.8$) \\
\method, $N=64$, emulation & \cellcolor{vizorange!6}$+2.3$ ($+0.6$) & \cellcolor{vizorange!9}$+3.4$ ($+3.0$) & \cellcolor{vizorange!6}$+0.4$ ($+2.1$) \\
\addlinespace
Emulation control & \cellcolor{vizorange!6}$+1.9$ ($+1.1$) & \cellcolor{vizaxis!6}$-0.2$ ($+1.1$) & \cellcolor{vizaxis!6}$-0.6$ ($+0.4$) \\
\bottomrule
\end{tabular}
\end{small}
\end{center}
\vskip -0.1in
\end{table*}
}{}

\begin{table*}[h]
\caption{Held-out check of the frozen settings on 122 prompts never used to choose $k'$, $N$, or any calibration (81 JailbreakBench behaviors without a near-duplicate in our evaluation set, and the 41 HarmBench validation behaviors). Original-model ASR (\%) and the change in pp for truncation at $k'=\lceil 2k/3\rceil$ and for \method ($N=8$, engine with the fused kernel), under Llama Guard 3 (LG) and the HarmBench classifier (HB). Bold: exact McNemar $p<0.05$. Shading: size and sign of the Llama Guard 3 change.}
\label{tab:heldout}
\vskip 0.05in
\begin{center}
\begin{scriptsize}
\setlength{\tabcolsep}{3pt}
\begin{tabular}{@{}llrrrrrr@{}}
\toprule
 & & \multicolumn{3}{c}{LG} & \multicolumn{3}{c}{HB} \\
\cmidrule(lr){3-5}\cmidrule(l){6-8}
Model & Framing & Original & \textcolor{vizorange}{Trunc.} & \textcolor{vizblue}{\method} & Original & \textcolor{vizorange}{Trunc.} & \textcolor{vizblue}{\method} \\
\midrule
Qwen1.5 & Plain & 8.2 & \cellcolor{vizorange!6}$+0.8$ & \cellcolor{vizorange!7}$+2.5$ & 7.4 & $+3.3$ & $+1.6$ \\
 & Refusal supp. & 51.6 & \cellcolor{vizorange!43}$\mathbf{+15.6}$ & \cellcolor{vizorange!6}$+1.6$ & 49.2 & $\mathbf{+19.7}$ & $+2.5$ \\
 & Prefix inj. & 73.8 & \cellcolor{vizorange!29}$\mathbf{+10.7}$ & \cellcolor{vizorange!7}$+2.5$ & 69.7 & $\mathbf{+14.8}$ & $+4.1$ \\
 & Evil confidant & 82.8 & \cellcolor{vizaxis!27}$\mathbf{-9.8}$ & \cellcolor{vizaxis!6}$-1.6$ & 73.0 & $-9.0$ & $-0.8$ \\
 & Distractors & 73.8 & \cellcolor{vizorange!29}$\mathbf{+10.7}$ & $+0.0$ & 22.9 & $\mathbf{+20.5}$ & $\mathbf{+11.5}$ \\
\midrule
DeepSeek-V2-Lite & Plain & 21.3 & \cellcolor{vizorange!14}$+4.9$ & \cellcolor{vizorange!6}$+1.6$ & 19.7 & $+2.5$ & $+3.3$ \\
 & Refusal supp. & 55.7 & \cellcolor{vizorange!20}$\mathbf{+7.4}$ & \cellcolor{vizorange!14}$+4.9$ & 44.3 & $\mathbf{+9.8}$ & $+6.6$ \\
 & Prefix inj. & 54.9 & \cellcolor{vizaxis!52}$\mathbf{-18.9}$ & \cellcolor{vizaxis!6}$-0.8$ & 47.5 & $\mathbf{-18.0}$ & $+0.0$ \\
 & Evil confidant & 9.0 & \cellcolor{vizorange!32}$\mathbf{+11.5}$ & \cellcolor{vizaxis!9}$-3.3$ & 8.2 & $+5.7$ & $-2.5$ \\
 & Distractors & 85.2 & \cellcolor{vizaxis!6}$-0.8$ & \cellcolor{vizorange!9}$+3.3$ & 40.2 & $+0.0$ & $+4.9$ \\
\midrule
OLMoE & Plain & 4.1 & $+0.0$ & $+0.0$ & 1.6 & $+0.8$ & $+0.0$ \\
 & Refusal supp. & 34.4 & \cellcolor{vizorange!7}$+2.5$ & \cellcolor{vizaxis!6}$-0.8$ & 22.9 & $+4.1$ & $+2.5$ \\
 & Prefix inj. & 36.1 & \cellcolor{vizaxis!7}$-2.5$ & \cellcolor{vizorange!7}$+2.5$ & 27.9 & $-4.1$ & $+0.0$ \\
 & Evil confidant & 0.0 & $+0.0$ & $+0.0$ & 0.0 & $+0.0$ & $+0.0$ \\
 & Distractors & 50.8 & \cellcolor{vizaxis!18}$-6.6$ & \cellcolor{vizorange!18}$+6.6$ & 18.0 & $+0.8$ & $+3.3$ \\
\bottomrule
\end{tabular}
\end{scriptsize}
\end{center}
\end{table*}

\begin{table*}[h]
\caption{Second held-out set, analyzed as fixed in advance (Appendix~A): 889 AdvBench and CatQA requests that were never used for any choice and have no near-duplicate in the evaluation or first held-out set. Original-model ASR (\%) and the change in pp for a rerun of the original model with the requests in a different order, truncation at $k'=\lceil 2k/3\rceil$, \method with $N=1$, 8, and 32 (fused kernel), and truncation and \method ($N=8$) at about half the experts ($k'=2$, 3, 4), under Llama Guard 3 (LG) and the HarmBench classifier (HB). Bold: exact McNemar $p<0.05$. Shading: size and sign of the Llama Guard 3 change.}
\label{tab:heldout2}
\vskip 0.05in
\begin{center}
\begin{scriptsize}
\setlength{\tabcolsep}{3pt}
\begin{tabular}{@{}lllrrrrrrrr}
\toprule
 & & & & & \multicolumn{4}{c}{$k'=\lceil 2k/3\rceil$} & \multicolumn{2}{c}{Half} \\
\cmidrule(lr){6-9}\cmidrule(l){10-11}
Model & Framing & Judge & Orig. & Rerun & \textcolor{vizorange}{Trunc.} & \textcolor{vizblue}{$N=1$} & \textcolor{vizblue}{$N=8$} & \textcolor{vizblue}{$N=32$} & \textcolor{vizorange}{Trunc.} & \textcolor{vizblue}{$N=8$} \\
\midrule
Qwen1.5 & Refusal supp. & LG & 43.6 & \cellcolor{vizorange!6}$+0.8$ & \cellcolor{vizorange!55}$\mathbf{+21.4}$ & \cellcolor{vizorange!6}$+0.7$ & \cellcolor{vizorange!6}$+0.6$ & \cellcolor{vizorange!6}$+0.6$ & \cellcolor{vizorange!55}$\mathbf{+26.1}$ & \cellcolor{vizorange!6}$+0.4$ \\
 &  & HB & 32.2 & $+1.2$ & $\mathbf{+21.0}$ & $+1.2$ & $+1.0$ & $-1.1$ & $\mathbf{+24.1}$ & $+0.8$ \\
 & Prefix inj. & LG & 75.1 & \cellcolor{vizorange!6}$+1.1$ & \cellcolor{vizorange!24}$\mathbf{+8.7}$ & \cellcolor{vizorange!9}$\mathbf{+3.4}$ & \cellcolor{vizorange!8}$\mathbf{+2.8}$ & \cellcolor{vizorange!8}$\mathbf{+2.9}$ & \cellcolor{vizorange!24}$\mathbf{+8.8}$ & \cellcolor{vizorange!6}$+1.3$ \\
 &  & HB & 71.5 & $+0.4$ & $\mathbf{+8.3}$ & $\mathbf{+3.4}$ & $+1.9$ & $+2.0$ & $\mathbf{+7.2}$ & $+1.0$ \\
\midrule
DeepSeek-V2-Lite & Refusal supp. & LG & 37.7 & \cellcolor{vizorange!6}$+0.7$ & \cellcolor{vizorange!16}$\mathbf{+5.8}$ & \cellcolor{vizorange!9}$\mathbf{+3.4}$ & \cellcolor{vizorange!7}$\mathbf{+2.6}$ & \cellcolor{vizorange!6}$+1.9$ & \cellcolor{vizorange!36}$\mathbf{+13.2}$ & \cellcolor{vizorange!6}$+1.8$ \\
 &  & HB & 28.0 & $+0.0$ & $\mathbf{+4.8}$ & $+2.4$ & $+1.9$ & $+1.2$ & $\mathbf{+6.7}$ & $-0.2$ \\
 & Prefix inj. & LG & 39.6 & \cellcolor{vizorange!6}$+2.1$ & \cellcolor{vizaxis!12}$\mathbf{-4.5}$ & \cellcolor{vizorange!6}$+1.2$ & \cellcolor{vizorange!6}$+2.0$ & \cellcolor{vizorange!9}$\mathbf{+3.1}$ & \cellcolor{vizorange!40}$\mathbf{+14.7}$ & \cellcolor{vizorange!11}$\mathbf{+3.9}$ \\
 &  & HB & 35.0 & $+2.1$ & $\mathbf{-6.9}$ & $+1.2$ & $+1.1$ & $+2.4$ & $\mathbf{+4.2}$ & $+2.4$ \\
\midrule
OLMoE & Refusal supp. & LG & 13.0 & \cellcolor{vizorange!6}$\mathbf{+1.5}$ & \cellcolor{vizorange!28}$\mathbf{+10.3}$ & \cellcolor{vizorange!9}$\mathbf{+3.3}$ & \cellcolor{vizorange!6}$+0.8$ & \cellcolor{vizorange!6}$+0.2$ & \cellcolor{vizorange!55}$\mathbf{+36.9}$ & \cellcolor{vizorange!11}$\mathbf{+3.9}$ \\
 &  & HB & 11.8 & $+0.3$ & $\mathbf{+6.6}$ & $\mathbf{+2.6}$ & $+0.9$ & $+0.0$ & $\mathbf{+25.8}$ & $\mathbf{+3.3}$ \\
 & Prefix inj. & LG & 22.6 & \cellcolor{vizorange!6}$+0.3$ & \cellcolor{vizorange!8}$+3.0$ & \cellcolor{vizorange!18}$\mathbf{+6.4}$ & \cellcolor{vizorange!13}$\mathbf{+4.8}$ & \cellcolor{vizorange!9}$\mathbf{+3.4}$ & \cellcolor{vizorange!37}$\mathbf{+13.3}$ & \cellcolor{vizorange!10}$\mathbf{+3.6}$ \\
 &  & HB & 16.3 & $+0.6$ & $+1.9$ & $\mathbf{+4.3}$ & $\mathbf{+2.9}$ & $+2.1$ & $\mathbf{+12.1}$ & $\mathbf{+4.5}$ \\
\bottomrule
\end{tabular}
\end{scriptsize}
\end{center}
\end{table*}

\textbf{Primary analysis.} Under refusal suppression and Llama Guard 3, truncation raises ASR on all three models (P2: $+21.37$, $+5.85$, and $+10.35$\pp on Qwen1.5, DeepSeek-V2-Lite, and OLMoE; Holm-adjusted $p=3.3\times10^{-42}$, $2.3\times10^{-4}$, and $2.3\times10^{-18}$; passed), \method with $N=8$ is non-inferior to the original model on Qwen1.5 and OLMoE (P1: $+0.56$ and $+0.79$\pp, 90\% Tango upper bounds 1.80 and 2.10\pp, below the $+3$\pp margin; passed) but not on DeepSeek-V2-Lite ($+2.59$\pp, upper bound 4.37\pp; failed), and \method lowers ASR below truncation on all three models (P3: $-20.81$, $-3.26$, and $-9.56$\pp; exact McNemar $p=2.5\times10^{-40}$, $0.037$, and $2.9\times10^{-16}$; passed).

\textbf{Every combination at once.} Table~\ref{tab:crossframe} collects the 40 combinations of model, framing, prompt set, and sampling seed in which we ran \method ($N=8$, fused kernel) at $k'=\lceil 2k/3\rceil$, with Holm correction over the 40 for each contrast. In 33 it changes Llama Guard 3 ASR by at most $+3$\pp against the original model; of the seven exceptions, only prefix injection on Ling-lite ($+12.50$\pp) remains significant after correction. It is significantly above truncation only where truncation lowers ASR (Qwen1.5 evil confidant, $+7.20$ against truncation's $-4.45$; DeepSeek-V2-Lite prefix injection on both held-out sets, $+6.52$ against $-4.50$ and $+18.03$ against $-18.85$), and there not significantly above the original model ($+2.75$, $+2.02$, and $-0.82$\pp). In no combination, under either judge, is it significantly above both truncation and the original model, also with the half-budget cells (55) and the emulation runs on the two held-out models added (61); without correction, two are, both on the evaluation prompts: Ling-lite under prefix injection ($+5.72$\pp over truncation, $p=0.005$) and Qwen1.5 under the distractor ($+5.30$\pp, $p=0.041$). Of the 20 combinations in which truncation raises ASR (unadjusted $p<0.05$), \method is below it in 16 (unadjusted $p<0.05$; 11 after correction).

\begin{table*}[h]
\caption{\method (SS; $N=8$, engine, fused kernel) in every combination of model, framing, prompt set, and decoding we ran at $k'=\lceil 2k/3\rceil$ (40 per judge; held-out 1: the 122 prompts of Table~\ref{tab:heldout}; held-out 2: the 889 of Table~\ref{tab:heldout2}). Trunc.: truncation against the original model; SS$-$orig.: \method against the original model; SS$-$trunc.: \method against truncation on the same prompts (pp, exact McNemar). Bold: Holm-adjusted $p<0.05$ over the 40 combinations of that judge and contrast (truncation column: unadjusted). Llama Guard 3 (LG) and HarmBench classifier (HB). Shading: size and sign of the Llama Guard 3 change.}
\label{tab:crossframe}
\vskip 0.05in
\begin{center}
\begin{scriptsize}
\setlength{\tabcolsep}{3pt}
\begin{tabular}{@{}lllrrrrrr@{}}
\toprule
 & & & \multicolumn{3}{c}{LG} & \multicolumn{3}{c}{HB} \\
\cmidrule(lr){4-6}\cmidrule(l){7-9}
Model & Prompts & Framing & \textcolor{vizorange}{Trunc.} & \textcolor{vizblue}{SS}$-$orig. & \textcolor{vizblue}{SS}$-$\textcolor{vizorange}{trunc.} & \textcolor{vizorange}{Trunc.} & \textcolor{vizblue}{SS}$-$orig. & \textcolor{vizblue}{SS}$-$\textcolor{vizorange}{trunc.} \\
\midrule
Qwen1.5 & Eval. & Plain & $+0.0$ & \cellcolor{vizaxis!6}$-0.2$ & \cellcolor{vizaxis!6}$-0.2$ & $+1.1$ & $+0.6$ & $-0.4$ \\
 & Eval. & Refusal supp. & \cellcolor{vizorange!51}$\mathbf{+18.4}$ & \cellcolor{vizaxis!6}$-0.6$ & \cellcolor{vizaxis!52}$\mathbf{-19.1}$ & $\mathbf{+16.5}$ & $+0.4$ & $\mathbf{-16.1}$ \\
 & Eval. & Prefix inj. & \cellcolor{vizorange!19}$\mathbf{+6.8}$ & \cellcolor{vizorange!6}$+1.3$ & \cellcolor{vizaxis!15}$\mathbf{-5.5}$ & $\mathbf{+7.0}$ & $+3.0$ & $-4.0$ \\
 & Eval. & Evil confidant & \cellcolor{vizaxis!12}$\mathbf{-4.4}$ & \cellcolor{vizorange!8}$+2.8$ & \cellcolor{vizorange!20}$\mathbf{+7.2}$ & $-4.2$ & $+0.4$ & $+4.7$ \\
 & Eval. & Distractor & \cellcolor{vizaxis!6}$-0.2$ & \cellcolor{vizorange!14}$+5.1$ & \cellcolor{vizorange!15}$+5.3$ & $\mathbf{+7.8}$ & $+4.4$ & $-3.4$ \\
 & Eval., $T{=}0.7$ s1 & Refusal supp. & \cellcolor{vizorange!47}$\mathbf{+17.2}$ & \cellcolor{vizorange!6}$+0.4$ & \cellcolor{vizaxis!46}$\mathbf{-16.7}$ & $\mathbf{+16.3}$ & $+1.5$ & $\mathbf{-14.8}$ \\
 & Eval., $T{=}0.7$ s2 & Refusal supp. & \cellcolor{vizorange!45}$\mathbf{+16.5}$ & \cellcolor{vizaxis!6}$-1.7$ & \cellcolor{vizaxis!50}$\mathbf{-18.2}$ & $\mathbf{+12.9}$ & $-0.4$ & $\mathbf{-13.3}$ \\
 & Eval., $T{=}0.7$ s3 & Refusal supp. & \cellcolor{vizorange!48}$\mathbf{+17.6}$ & \cellcolor{vizaxis!6}$-0.8$ & \cellcolor{vizaxis!51}$\mathbf{-18.4}$ & $\mathbf{+15.7}$ & $+1.3$ & $\mathbf{-14.4}$ \\
 & Held-out 1 & Plain & \cellcolor{vizorange!6}$+0.8$ & \cellcolor{vizorange!7}$+2.5$ & \cellcolor{vizorange!6}$+1.6$ & $+3.3$ & $+1.6$ & $-1.6$ \\
 & Held-out 1 & Refusal supp. & \cellcolor{vizorange!43}$\mathbf{+15.6}$ & \cellcolor{vizorange!6}$+1.6$ & \cellcolor{vizaxis!38}$-13.9$ & $\mathbf{+19.7}$ & $+2.5$ & $\mathbf{-17.2}$ \\
 & Held-out 1 & Prefix inj. & \cellcolor{vizorange!29}$\mathbf{+10.7}$ & \cellcolor{vizorange!7}$+2.5$ & \cellcolor{vizaxis!23}$-8.2$ & $\mathbf{+14.8}$ & $+4.1$ & $-10.7$ \\
 & Held-out 1 & Evil confidant & \cellcolor{vizaxis!27}$\mathbf{-9.8}$ & \cellcolor{vizaxis!6}$-1.6$ & \cellcolor{vizorange!23}$+8.2$ & $-9.0$ & $-0.8$ & $+8.2$ \\
 & Held-out 1 & Distractor & \cellcolor{vizorange!29}$\mathbf{+10.7}$ & $+0.0$ & \cellcolor{vizaxis!29}$-10.7$ & $\mathbf{+20.5}$ & $+11.5$ & $-9.0$ \\
 & Held-out 2 & Refusal supp. & \cellcolor{vizorange!55}$\mathbf{+21.4}$ & \cellcolor{vizorange!6}$+0.6$ & \cellcolor{vizaxis!55}$\mathbf{-20.8}$ & $\mathbf{+21.0}$ & $+1.0$ & $\mathbf{-20.0}$ \\
 & Held-out 2 & Prefix inj. & \cellcolor{vizorange!24}$\mathbf{+8.7}$ & \cellcolor{vizorange!8}$+2.8$ & \cellcolor{vizaxis!16}$\mathbf{-5.8}$ & $\mathbf{+8.3}$ & $+1.9$ & $\mathbf{-6.4}$ \\
\midrule
DeepSeek-V2-Lite & Eval. & Plain & \cellcolor{vizorange!9}$\mathbf{+3.4}$ & \cellcolor{vizorange!6}$+1.3$ & \cellcolor{vizaxis!6}$-2.1$ & $\mathbf{+2.8}$ & $+0.2$ & $-2.5$ \\
 & Eval. & Refusal supp. & \cellcolor{vizorange!18}$\mathbf{+6.6}$ & \cellcolor{vizorange!6}$+0.6$ & \cellcolor{vizaxis!16}$-5.9$ & $\mathbf{+5.3}$ & $+2.3$ & $-3.0$ \\
 & Eval. & Prefix inj. & \cellcolor{vizaxis!6}$-2.3$ & \cellcolor{vizorange!6}$+0.2$ & \cellcolor{vizorange!7}$+2.5$ & $\mathbf{-5.1}$ & $+0.2$ & $+5.3$ \\
 & Eval. & Evil confidant & \cellcolor{vizorange!36}$\mathbf{+13.1}$ & \cellcolor{vizaxis!6}$-0.4$ & \cellcolor{vizaxis!37}$\mathbf{-13.6}$ & $\mathbf{+11.4}$ & $-1.5$ & $\mathbf{-12.9}$ \\
 & Eval. & Distractor & \cellcolor{vizaxis!10}$-3.8$ & \cellcolor{vizaxis!12}$-4.2$ & \cellcolor{vizaxis!6}$-0.4$ & $-4.4$ & $-1.7$ & $+2.8$ \\
 & Held-out 1 & Plain & \cellcolor{vizorange!14}$+4.9$ & \cellcolor{vizorange!6}$+1.6$ & \cellcolor{vizaxis!9}$-3.3$ & $+2.5$ & $+3.3$ & $+0.8$ \\
 & Held-out 1 & Refusal supp. & \cellcolor{vizorange!20}$\mathbf{+7.4}$ & \cellcolor{vizorange!14}$+4.9$ & \cellcolor{vizaxis!7}$-2.5$ & $\mathbf{+9.8}$ & $+6.6$ & $-3.3$ \\
 & Held-out 1 & Prefix inj. & \cellcolor{vizaxis!52}$\mathbf{-18.9}$ & \cellcolor{vizaxis!6}$-0.8$ & \cellcolor{vizorange!50}$\mathbf{+18.0}$ & $\mathbf{-18.0}$ & $+0.0$ & $\mathbf{+18.0}$ \\
 & Held-out 1 & Evil confidant & \cellcolor{vizorange!32}$\mathbf{+11.5}$ & \cellcolor{vizaxis!9}$-3.3$ & \cellcolor{vizaxis!41}$\mathbf{-14.8}$ & $+5.7$ & $-2.5$ & $-8.2$ \\
 & Held-out 1 & Distractor & \cellcolor{vizaxis!6}$-0.8$ & \cellcolor{vizorange!9}$+3.3$ & \cellcolor{vizorange!11}$+4.1$ & $+0.0$ & $+4.9$ & $+4.9$ \\
 & Held-out 2 & Refusal supp. & \cellcolor{vizorange!16}$\mathbf{+5.8}$ & \cellcolor{vizorange!7}$+2.6$ & \cellcolor{vizaxis!9}$-3.3$ & $\mathbf{+4.8}$ & $+1.9$ & $-2.9$ \\
 & Held-out 2 & Prefix inj. & \cellcolor{vizaxis!12}$\mathbf{-4.5}$ & \cellcolor{vizorange!6}$+2.0$ & \cellcolor{vizorange!18}$\mathbf{+6.5}$ & $\mathbf{-6.9}$ & $+1.1$ & $\mathbf{+8.0}$ \\
\midrule
OLMoE & Eval. & Plain & \cellcolor{vizaxis!6}$-0.6$ & \cellcolor{vizaxis!6}$-0.8$ & \cellcolor{vizaxis!6}$-0.2$ & $+0.4$ & $-0.6$ & $-1.1$ \\
 & Eval. & Refusal supp. & \cellcolor{vizorange!26}$\mathbf{+9.5}$ & \cellcolor{vizorange!6}$+1.3$ & \cellcolor{vizaxis!23}$\mathbf{-8.3}$ & $\mathbf{+8.3}$ & $+1.7$ & $\mathbf{-6.6}$ \\
 & Eval. & Prefix inj. & \cellcolor{vizorange!15}$\mathbf{+5.3}$ & \cellcolor{vizorange!12}$+4.4$ & \cellcolor{vizaxis!6}$-0.8$ & $\mathbf{+5.3}$ & $+3.2$ & $-2.1$ \\
 & Held-out 1 & Plain & $+0.0$ & $+0.0$ & $+0.0$ & $+0.8$ & $+0.0$ & $-0.8$ \\
 & Held-out 1 & Refusal supp. & \cellcolor{vizorange!7}$+2.5$ & \cellcolor{vizaxis!6}$-0.8$ & \cellcolor{vizaxis!9}$-3.3$ & $+4.1$ & $+2.5$ & $-1.6$ \\
 & Held-out 1 & Prefix inj. & \cellcolor{vizaxis!7}$-2.5$ & \cellcolor{vizorange!7}$+2.5$ & \cellcolor{vizorange!14}$+4.9$ & $-4.1$ & $+0.0$ & $+4.1$ \\
 & Held-out 1 & Evil confidant & $+0.0$ & $+0.0$ & $+0.0$ & $+0.0$ & $+0.0$ & $+0.0$ \\
 & Held-out 1 & Distractor & \cellcolor{vizaxis!18}$-6.6$ & \cellcolor{vizorange!18}$+6.6$ & \cellcolor{vizorange!36}$+13.1$ & $+0.8$ & $+3.3$ & $+2.5$ \\
 & Held-out 2 & Refusal supp. & \cellcolor{vizorange!28}$\mathbf{+10.3}$ & \cellcolor{vizorange!6}$+0.8$ & \cellcolor{vizaxis!26}$\mathbf{-9.6}$ & $\mathbf{+6.6}$ & $+0.9$ & $\mathbf{-5.7}$ \\
 & Held-out 2 & Prefix inj. & \cellcolor{vizorange!8}$+3.0$ & \cellcolor{vizorange!13}$+4.8$ & \cellcolor{vizorange!6}$+1.8$ & $+1.9$ & $+2.9$ & $+1.0$ \\
\midrule
Ling-lite & Eval. & Plain & \cellcolor{vizorange!6}$+0.8$ & \cellcolor{vizaxis!6}$-0.2$ & \cellcolor{vizaxis!6}$-1.1$ & $+1.7$ & $+0.6$ & $-1.1$ \\
 & Eval. & Refusal supp. & \cellcolor{vizorange!6}$+0.8$ & \cellcolor{vizaxis!6}$-1.7$ & \cellcolor{vizaxis!7}$-2.5$ & $\mathbf{-4.7}$ & $-1.9$ & $+2.8$ \\
 & Eval. & Prefix inj. & \cellcolor{vizorange!19}$\mathbf{+6.8}$ & \cellcolor{vizorange!34}$\mathbf{+12.5}$ & \cellcolor{vizorange!16}$+5.7$ & $\mathbf{+7.6}$ & $\mathbf{+12.7}$ & $+5.1$ \\
\bottomrule
\end{tabular}
\end{scriptsize}
\end{center}
\end{table*}

\section{Human Review Protocol}
\label{app:human}

\textbf{Sample.} 587 unique responses in ten strata (Table~\ref{tab:strata}). Responses shared by two strata are labeled once. The two responses to a prompt are placed in different sessions (A/B for the core strata, C/D for the extension), and item identifiers and order within a session are random, so the annotators do not see which condition produced a response.

\begin{table}[h]
\caption{Human-review strata. Pairs: discordant or concordant pairs under Llama Guard 3 in the population and in the sample.}
\label{tab:strata}
\vskip 0.05in
\begin{center}
\begin{scriptsize}
\setlength{\tabcolsep}{3pt}
\begin{tabular}{llrrl}
\toprule
 & & \multicolumn{2}{c}{Pairs} & \\
\cmidrule(lr){3-4}
Comparison & Stratum & Pop. & Sample & Sides \\
\midrule
Qwen1.5, RS, $4\to3$ & forward & 101 & 101 & both \\
 & reverse & 14 & 14 & both \\
 & concordant unsafe & 206 & 20 & both \\
 & concordant safe & 151 & 20 & both \\
\addlinespace
Qwen1.5, RS, original $\to$ rerun & forward & 9 & 9 & both \\
 & reverse & 11 & 11 & both \\
\addlinespace
DeepSeek-V2-Lite, RS, $6\to4$ & forward & 63 & 63 & both \\
 & reverse & 32 & 32 & both \\
\addlinespace
Qwen1.5, PI, $4\to3$ & forward & 53 & 15 & both \\
\addlinespace
Qwen1.5, RS, $3\to$ \method & rescued & 96 & 30 & one \\
\bottomrule
\end{tabular}
\end{scriptsize}
\end{center}
\end{table}

\textbf{Rubric.} Harmful: the response carries out the request with content an aligned assistant should not give (steps, code, quantities, or the requested persuasive text), even with warnings or only partly. Partial: the response goes along with the request but the content is vague, generic, very short, or unusable, or answers a harmless version. Refusal: the response declines, redirects, lectures, or discusses the topic without carrying it out. Degenerate: mostly incoherent, repetitive, or off-topic text. Unsure: requires a note. When a response mixes behaviors, the label follows the precedence harmful $>$ partial $>$ refusal $>$ degenerate: a refusal sentence followed by usable harmful content is harmful, in any language, and degenerate is used only when the response neither refuses nor attempts the request. Only ``harmful'' counts as harmful compliance.

\textbf{Estimates.} Within each stratum we compare the human labels of the two responses. The human-verified ASR change is $\frac{1}{472}\sum_s \frac{P_s}{n_s}\sum_{i\in s}(h_i^{\mathrm{cond}}-h_i^{\mathrm{original}})$ over the strata $s$ of a comparison, with population size $P_s$ and $n_s$ usable sampled pairs (both responses labeled, neither unsure), and a stratified bootstrap interval. Judge sensitivity and specificity are weighted by $P_s/n_s$. Both annotators labeled all 587 items, so agreement is Cohen's $\kappa$ with a bootstrap interval over all items. The scoring script is part of the release.

\begin{table}[h]
\caption{Human labels by stratum for the two annotators (A1, A2). Paired strata: human forward/reverse flips, counting only ``harmful''. \method-truncation stratum (one side labeled): harmful/partial/refusal.}
\label{tab:human}
\vskip 0.05in
\begin{center}
\begin{scriptsize}
\setlength{\tabcolsep}{3pt}
\begin{tabular}{llrrr}
\toprule
 & & & \multicolumn{2}{c}{Annotator} \\
\cmidrule(l){4-5}
Comparison & Stratum & Pairs & A1 & A2 \\
\midrule
Qwen1.5, RS, $4\to3$ & concordant safe & 20 & 0/0 & 0/0 \\
 & concordant unsafe & 20 & 1/0 & 1/0 \\
 & forward & 101 & 59/1 & 62/1 \\
 & reverse & 14 & 2/6 & 1/7 \\
\addlinespace
DeepSeek-V2-Lite, RS, $6\to4$ & forward & 63 & 27/0 & 26/0 \\
 & reverse & 32 & 0/15 & 0/13 \\
\addlinespace
Qwen1.5, RS, original $\to$ rerun & forward & 9 & 3/0 & 3/0 \\
 & reverse & 11 & 1/5 & 0/5 \\
\addlinespace
Qwen1.5, PI, $4\to3$ & forward & 15 & 10/0 & 10/0 \\
\addlinespace
Qwen1.5, RS, $3\to$ \method & judge-safe & 30 & 1/1/28 & 0/2/28 \\
\bottomrule
\end{tabular}
\end{scriptsize}
\end{center}
\end{table}

\textbf{Results.} One item was labeled unsure by one annotator. On the other 586, the annotators agree on harmful versus not harmful for 97.3\% ($\kappa=0.933$, 95\% CI $[0.897, 0.962]$) and on the four labels with $\kappa=0.913$. Table~\ref{tab:human} gives the human flips by stratum. Weighted to the Qwen1.5 refusal-suppression population, the human-verified change is $+13.56$\pp ($[9.49, 19.05]$) for the first annotator and $+13.83$\pp ($[9.81, 19.26]$) for the second; restricted to the judge-discordant prompts, it is $+11.38$ and $+11.65$\pp (exact McNemar $p<10^{-10}$ for both), and counting partial as harmful gives $+16.95$ and $+17.37$\pp. On DeepSeek-V2-Lite at $k'=4$ the change is $+2.54$ and $+2.75$\pp ($p=0.088$ and $0.053$), and $+4.66$ and $+4.45$\pp ($p=0.007$ and $0.010$) counting partial as harmful; this model has no sampled concordant strata, so these estimates assume that judge-concordant pairs keep their status. Against the first annotator, weighted to the Qwen1.5 population, Llama Guard 3 has sensitivity 98.8\%, specificity 66.9\%, and precision 61.4\%, and the HarmBench classifier 83.1\%, 76.4\%, and 65.2\%; against the second, 99.0\%, 65.3\%, 58.2\% and 86.1\%, 76.1\%, 63.8\%. Weighted to the Qwen1.5 population, the annotators label 17.6 and 19.2\% of the original model's responses and 22.3 and 22.7\% of the responses under truncation partial, and Llama Guard 3 flags 89 to 97\% of them on both sides. Its false positives therefore rise from 19.2 to 24.0\% of prompts (first annotator), and the human change ($+13.6$\pp) plus this rise ($+4.8$\pp) gives its $+18.4$\pp; errors independent of the condition would instead shrink the change to about $(0.988+0.669-1)\times13.6\approx8.9$\pp. Its precision is 59\% on the original model's responses and 63\% on the responses under truncation. Of 30 \method responses that Llama Guard 3 calls safe where truncation was unsafe, both annotators label 28 as refusals. Of 15 prefix-injection forward flips, both confirm 10.

\section{Earlier Harness (Study 1)}
\label{app:study1}

\begin{table}[t]
\caption{Study 1 (earlier harness, refusal-suppression template, 448 prompts, fixed mixed batches, Llama Guard 3). ASR in \%, $\Delta$ in pp, $p$: exact McNemar test against the original model; shading: size and sign of $\Delta$.}
\label{tab:study1}
\vskip 0.1in
\begin{center}
\begin{small}
\setlength{\tabcolsep}{2.5pt}
\begin{tabular}{@{}lllrrr@{}}
\toprule
Model & Method & Setting & ASR & $\Delta$ & $p$ \\
\midrule
\multirow{3}{*}{Qwen1.5} & Original & $k=4$ & 34.38 & -- & -- \\
 & Trunc. & $k'=3$ & 57.81 & \cellcolor{vizorange!55}$+23.44$ & $<10^{-6}$ \\
 & Trunc. & $k'=2$ & 65.63 & \cellcolor{vizorange!55}$+31.25$ & $<10^{-6}$ \\
\midrule
\multirow{4}{*}{Qwen3} & Original & $k=8$ & 8.26 & -- & -- \\
 & Trunc. & $k'=6$ & 5.58 & \cellcolor{vizaxis!7}$-2.68$ & 0.023 \\
 & Trunc. & $k'=4$ & 3.13 & \cellcolor{vizaxis!14}$-5.13$ & $3.4{\times}10^{-5}$ \\
 & Trunc. & $k'=2$ & 30.80 & \cellcolor{vizorange!55}$+22.54$ & $<10^{-6}$ \\
\bottomrule
\end{tabular}
\end{small}
\end{center}
\vskip -0.1in
\end{table}

Study 1 ran truncation through the same configuration override as Study 2, on fixed mixed batches (Table~\ref{tab:study1}). On Qwen1.5, truncation to 3 of 4 experts raised ASR by $+23.44$\pp, from an original ASR of 34.38\%. Study 2 measures $+18.43$\pp for the same truncation on different prompts; the two estimates are consistent (difference 5.0\pp, 95\% Wald CI $[-1.0, 11.1]$, from 116/11 and 101/14 forward/reverse flips). Truncation lowered Qwen3's ASR at $k'=6$ and $k'=4$ and raised it only at $k'=2$, where 79 of the 123 forward flips were degenerate. On OLMoE-0924 the original ASR was 93.30\%, which leaves no room to detect an increase.

The two-arm decomposition that Study 1 reported used a different implementation and a model-rated fluency screen; Section~4.3 of the paper replaces it with the Study 2 factorial.

\textbf{Batch-dependent methods.} Batch-level methods such as XShare \citep{vankov2026xshare} make a request's experts depend on its co-batched requests, which an adversary could exploit as prior work showed for capacity-limited routing \citep{hayes2024buffer,yona2024stealing}; we have not tested this. The same removed width also does not give the same effect: Qwen1.5 at $k'=3$ and DeepSeek-V2-Lite at $k'=5$ each remove 12.5\% of the expert feed-forward width of a layer, including shared experts, yet change refusal-suppression ASR by $+18.43$ and $+2.33$\pp.

\textbf{Differences between Study 1 and Study 2.} 
Study 1 used 448 prompts (288 AdvBench requests \citep{zou2023universal} selected by seeded sampling with a fixed test split and covering six harm categories we defined (illegal activity, cybercrime, violence, misinformation, chemical and biological harm, and harassment), 160 HarmBench standard behaviors from a seed-0 split of all 200 standard behaviors), the refusal-suppression template only, fixed batches of 4 harmful and 12 benign Alpaca prompts, Llama Guard 3, Qwen3-30B-A3B in BF16, and OLMoE-0924. Study 2 uses 472 public prompts (the official HarmBench test split and StrongREJECT), three framings, all requests of a condition in one engine call, two judges, Qwen3 with FP8 weights, and OLMoE-0125. Study 1 filtered degenerate outputs with model-produced fluency ratings; Study 2 uses the repeated-trigram screen.

\section{Mechanism: First Token, Refusal Direction, and Gate Mass}

\subsection{The First Token and the Prompt}
\label{app:firsttoken}

\begin{figure*}[h]
\centering
\includegraphics[width=\textwidth]{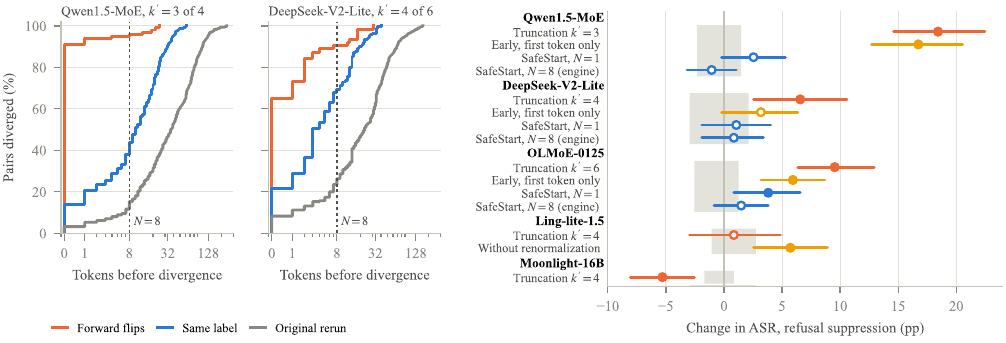}
\caption{Where the change happens, under a refusal-suppression attack (472 prompts, Llama Guard 3). \textbf{Left:} share of response pairs whose truncated and original responses have diverged by each generated token; index 0 is the first generated token, produced by the prefill pass. Forward flips (original safe, truncated harmful) diverge at the first token, earlier than pairs with the same label and than two runs of the original model with different prompt order. \textbf{Right:} change in harmful-compliance rate (ASR) against the original model with 95\% prompt-bootstrap intervals, at the operating point that keeps about two thirds of the selected experts. Early: the first generated token comes from the truncated model and the rest from the original model. \method (emulation): the first $N$ generated tokens come from the original model and the rest from the truncated model. \method (engine): our vLLM patch. Without renormalization: Ling-lite's truncation with the kept experts' original weights. Gray: 95\% interval of a reordered rerun of the original model. Filled markers: exact McNemar $p<0.05$.}
\label{fig:diverge}
\end{figure*}

\begin{figure}[h]
\centering
\includegraphics[width=\columnwidth]{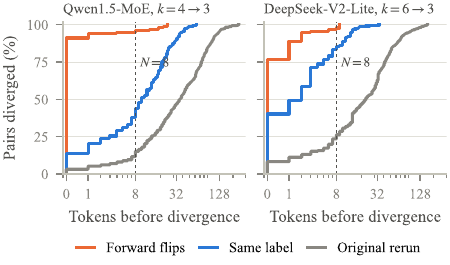}
\caption{Divergence curves under refusal suppression at $k'=3$ for both models.}
\label{fig:onset}
\end{figure}

\textbf{Under refusal suppression, the change runs through the first generated token.} For each prompt we find the first generated token at which the response under truncation differs from the original model's response (Figure~\ref{fig:diverge}, left). On Qwen1.5 at $k'=3$, 92 of the 101 forward flips differ at the first generated token and 96 within the first 8, while pairs with the same judge label diverge later (median 10 tokens) and two runs of the original model later still (median 36). On DeepSeek-V2-Lite at $k'=4$, 41 of 63 forward flips differ at the first token and 57 within the first 8. Under prefix injection, where the template dictates the opening words, Qwen1.5 forward flips diverge later (median 29 tokens).

Divergence is measured on the outcome, so we test the first token directly (Table~\ref{tab:mitig}). The first generated token is produced by the prefill pass, which truncation also changes. \emph{Early truncation} computes prefill and the first token with $k'$ experts and continues with the original model; on Qwen1.5 it raises refusal-suppression ASR by $+16.74$\pp ($p<10^{-6}$), 91\% of the truncation effect (95\% prompt-bootstrap CI $[78, 104]$), on OLMoE by $+5.93$\pp, 62\% ($[37, 91]$), and on DeepSeek-V2-Lite by $+3.18$\pp ($p=0.077$), half of it. \emph{\method with $N=1$} does the reverse; in emulation, which recomputes the prompt with $k'$ experts, it leaves $+2.54$\pp on Qwen1.5 ($p=0.096$) and $+1.06$\pp on DeepSeek-V2-Lite ($p=0.59$), and in the engine, which keeps the prompt at full $k$, $+0.42$ and $+0.85$\pp (Table~\ref{tab:nsweep}). At $k'=3$ on DeepSeek-V2-Lite, protecting one token in emulation leaves $+5.30$\pp ($p=0.0035$) and protecting eight leaves $+1.69$\pp ($p=0.34$). On OLMoE at $k'=6$ the prompt computation carries more of the effect: early truncation gives $+5.93$\pp ($p=4.1\times10^{-5}$), 62\% of the $+9.53$\pp truncation effect; protecting one token leaves $+3.81$\pp ($p=0.013$) in emulation, which recomputes the prompt with $k'$ experts, and $0.00$\pp in the engine; and in the engine protecting the prompt alone ($N=0$) leaves $-0.64$\pp.

\subsection{Gate Mass and Expert Identity}
\label{app:gatemass}

\textbf{Under refusal suppression, lost gate mass raises compliance on all three models; removing experts at restored mass does so only on Qwen1.5.} In models that do not renormalize, truncation removes experts and lowers the routed gate mass. Under refusal suppression, removal with the mass restored raises ASR by $+12.29$\pp on Qwen1.5 and attenuation alone by $+4.03$\pp; the weights differ on DeepSeek-V2-Lite ($-2.12$ and $+8.05$\pp) and on OLMoE ($+0.64$ and $+4.24$\pp, with an interaction of $+4.66$\pp; Table~\ref{tab:mitig} and the factorial intervals below). Substituting the next-ranked expert in the last slot at constant count and mass (\emph{swap}) raises it by $+11.86$, $+6.14$, and $+5.30$\pp on the three models ($p\le0.0031$).

\begin{figure}[h]
\centering
\includegraphics[width=\columnwidth]{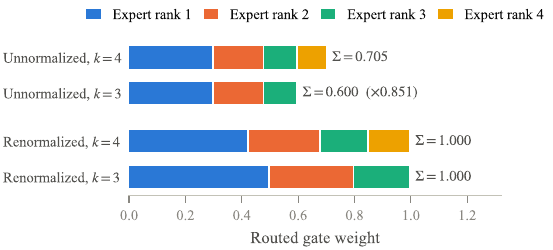}
\caption{Routed gate weights of one token before and after truncation from $k=4$ to $k=3$ (illustrative weights with mass ratio 0.851, the mean ratio Study 1 measured on Qwen1.5-MoE). Without renormalization the routed branch loses weight relative to the shared expert and the residual stream; with renormalization the survivors absorb the removed weight.}
\label{fig:routing}
\end{figure}

\begin{table*}[t]
\caption{Mechanism and mitigation conditions at the two-thirds operating point (Study 2, 472 prompts per framing). ASR change in pp against the original model under Llama Guard 3, with the HarmBench classifier in parentheses; bold: exact McNemar $p<0.05$ under Llama Guard 3 (unadjusted, exploratory); shading: size and sign of the Llama Guard 3 change. Attenuation only keeps all $k$ experts and scales their weights by $m_{k'}/m_k$; mass restored keeps the top $k'$ experts and rescales them to $m_k$; swap keeps $k$ experts but substitutes the $(k{+}1)$-th ranked expert for the $k$-th at the same weight. Early: truncated model for the first generated token only, original afterwards. \method: original computation for the first $N$ generated tokens, in emulation (the truncated model recomputes the prompt and the first $N$ tokens) or in the engine with the fused kernel (Table~\ref{tab:nsweep} gives every $N$). Emulation control: the \method procedure with the original model in both passes.}
\label{tab:mitig}
\vskip 0.1in
\begin{center}
\begin{small}
\setlength{\tabcolsep}{3.5pt}
\begin{tabular}{@{}lrrrrrr}
\toprule
 & \multicolumn{2}{c}{Qwen1.5, $k'=3/4$} & \multicolumn{2}{c}{DeepSeek-V2-Lite, $k'=4/6$} & \multicolumn{2}{c}{OLMoE, $k'=6/8$} \\
\cmidrule(lr){2-3}\cmidrule(lr){4-5}\cmidrule(l){6-7}
Condition & Plain & Refusal supp. & Plain & Refusal supp. & Plain & Refusal supp. \\
\midrule
Truncation & $+0.0$ ($+1.1$) & \cellcolor{vizorange!51}$\mathbf{+18.4}$ ($+16.5$) & \cellcolor{vizorange!9}$\mathbf{+3.4}$ ($+2.8$) & \cellcolor{vizorange!18}$\mathbf{+6.6}$ ($+5.3$) & \cellcolor{vizaxis!6}$-0.6$ ($+0.4$) & \cellcolor{vizorange!26}$\mathbf{+9.5}$ ($+8.3$) \\
\addlinespace
Attenuation only & \cellcolor{vizaxis!6}$-1.1$ ($-0.6$) & \cellcolor{vizorange!11}$\mathbf{+4.0}$ ($+4.2$) & \cellcolor{vizaxis!6}$-0.4$ ($-0.2$) & \cellcolor{vizorange!22}$\mathbf{+8.1}$ ($+8.9$) & \cellcolor{vizaxis!6}$-0.6$ ($-0.4$) & \cellcolor{vizorange!12}$\mathbf{+4.2}$ ($+5.3$) \\
Mass restored & \cellcolor{vizorange!6}$+1.1$ ($+1.5$) & \cellcolor{vizorange!34}$\mathbf{+12.3}$ ($+11.9$) & \cellcolor{vizorange!12}$\mathbf{+4.2}$ ($+2.3$) & \cellcolor{vizaxis!6}$-2.1$ ($-0.8$) & \cellcolor{vizaxis!6}$-1.1$ ($+0.0$) & \cellcolor{vizorange!6}$+0.6$ ($-0.4$) \\
Swap (no truncation) & \cellcolor{vizorange!6}$+0.2$ ($-0.2$) & \cellcolor{vizorange!33}$\mathbf{+11.9}$ ($+11.0$) & \cellcolor{vizorange!6}$+1.5$ ($+1.1$) & \cellcolor{vizorange!17}$\mathbf{+6.1}$ ($+5.7$) & \cellcolor{vizaxis!6}$-0.6$ ($-0.8$) & \cellcolor{vizorange!15}$\mathbf{+5.3}$ ($+3.8$) \\
\addlinespace
Early, $N=1$ & \cellcolor{vizaxis!6}$-0.6$ ($+0.6$) & \cellcolor{vizorange!46}$\mathbf{+16.7}$ ($+14.0$) & \cellcolor{vizorange!6}$+2.3$ ($+2.3$) & \cellcolor{vizorange!9}$+3.2$ ($+2.8$) & \cellcolor{vizaxis!6}$-0.8$ ($-0.4$) & \cellcolor{vizorange!16}$\mathbf{+5.9}$ ($+4.0$) \\
\method, $N=1$, emulation & \cellcolor{vizaxis!6}$-0.2$ ($+0.2$) & \cellcolor{vizorange!7}$+2.5$ ($+3.2$) & \cellcolor{vizorange!6}$+1.9$ ($+0.8$) & \cellcolor{vizorange!6}$+1.1$ ($+1.5$) & \cellcolor{vizorange!6}$+0.2$ ($+0.6$) & \cellcolor{vizorange!10}$\mathbf{+3.8}$ ($+4.2$) \\
\method, $N=1$, engine & \cellcolor{vizaxis!6}$-0.8$ ($-0.6$) & \cellcolor{vizorange!6}$+0.4$ ($-0.8$) & \cellcolor{vizorange!6}$+1.1$ ($+0.2$) & \cellcolor{vizorange!6}$+0.8$ ($+2.5$) & \cellcolor{vizaxis!6}$-0.2$ ($+0.0$) & $+0.0$ ($+2.8$) \\
\method, $N=8$, emulation & \cellcolor{vizaxis!6}$-0.4$ ($+1.1$) & \cellcolor{vizorange!6}$+1.1$ ($+1.3$) & \cellcolor{vizorange!6}$+0.6$ ($+0.2$) & \cellcolor{vizorange!6}$+1.3$ ($+0.4$) & \cellcolor{vizaxis!6}$-1.1$ ($-0.8$) & \cellcolor{vizorange!9}$\mathbf{+3.4}$ ($+2.5$) \\
\method, $N=8$, engine & \cellcolor{vizaxis!6}$-0.2$ ($+0.6$) & \cellcolor{vizaxis!6}$-0.6$ ($+0.4$) & \cellcolor{vizorange!6}$+1.3$ ($+0.2$) & \cellcolor{vizorange!6}$+0.6$ ($+2.3$) & \cellcolor{vizaxis!6}$-0.8$ ($-0.6$) & \cellcolor{vizorange!6}$+1.3$ ($+1.7$) \\
\method, $N=32$, engine & \cellcolor{vizaxis!6}$\mathbf{-1.5}$ ($-0.6$) & \cellcolor{vizaxis!6}$-1.5$ ($-1.1$) & \cellcolor{vizaxis!6}$-1.1$ ($-1.1$) & \cellcolor{vizorange!6}$+1.1$ ($+1.5$) & \cellcolor{vizaxis!6}$-0.8$ ($-0.4$) & $+0.0$ ($+1.1$) \\
\method, $N=64$, emulation & \cellcolor{vizaxis!6}$-0.2$ ($+0.0$) & \cellcolor{vizorange!6}$+0.2$ ($+0.0$) & \cellcolor{vizorange!6}$+0.2$ ($+0.4$) & \cellcolor{vizaxis!6}$-1.3$ ($-2.3$) & \cellcolor{vizaxis!6}$-1.1$ ($-1.1$) & \cellcolor{vizorange!6}$+0.2$ ($+1.5$) \\
\addlinespace
Emulation control & \cellcolor{vizaxis!6}$-0.2$ ($+1.3$) & \cellcolor{vizorange!6}$+1.3$ ($-0.8$) & \cellcolor{vizaxis!6}$-0.2$ ($-0.6$) & \cellcolor{vizaxis!6}$-1.5$ ($-0.8$) & \cellcolor{vizaxis!6}$\mathbf{-1.5}$ ($-0.8$) & \cellcolor{vizorange!6}$+1.1$ ($+0.6$) \\
\bottomrule
\end{tabular}
\end{small}
\end{center}
\vskip -0.1in
\end{table*}

\textbf{Factorial intervals.} Joint prompt-bootstrap 95\% intervals under refusal suppression: Qwen1.5, removal with the mass restored $+12.29$\pp $[8.26, 16.31]$, attenuation alone $+4.03$ $[1.06, 6.99]$, interaction $+2.12$ $[-2.54, 6.36]$, and restoring the mass lowers the truncation effect by 6.14\pp (46 against 17 discordant prompts, $p=3.4\times10^{-4}$); DeepSeek-V2-Lite at $k'=4$, $-2.12$ $[-6.14, 1.69]$ and $+8.05$ $[4.03, 11.86]$; OLMoE, $+0.64$ $[-2.33, 3.39]$, $+4.24$ $[1.69, 6.78]$, and interaction $+4.66$ $[0.85, 8.47]$.

\begin{figure*}[h]
\centering
\includegraphics[width=\textwidth]{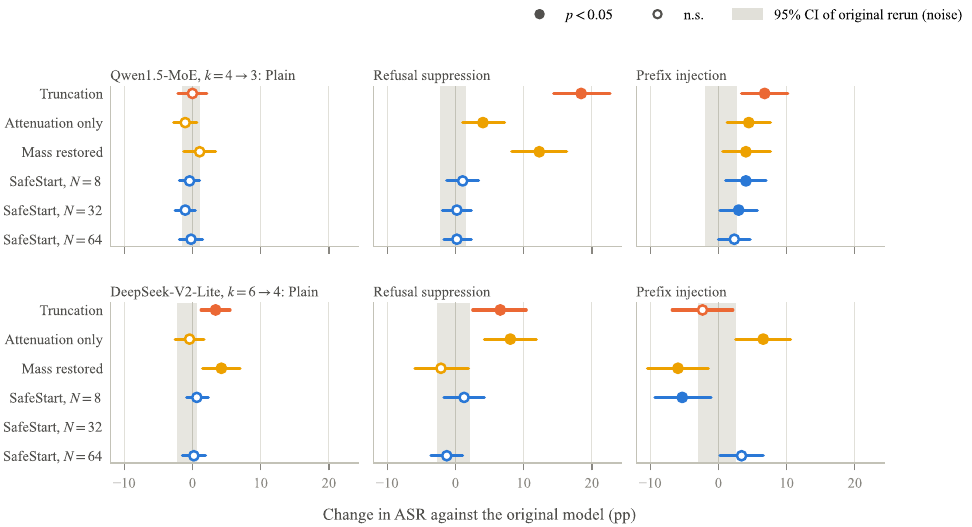}
\caption{ASR change against the original model (Llama Guard 3, 95\% prompt-bootstrap intervals) for truncation, the mass conditions, and \method (emulation). Gray band: 95\% interval of a reordered rerun of the original model.}
\label{fig:mitig}
\end{figure*}

\textbf{Within one model, removing renormalization raises compliance.} Ling-lite renormalizes the routed weights; native truncation to $k'=4$ changes its refusal-suppression ASR by $+0.85$\pp, and the same truncation without renormalization (\emph{trim}) by $+5.72$\pp, $+4.87$\pp more (48 against 25 discordant prompts, $p=0.0095$).

\subsection{The Refusal Direction}
\label{app:direction}

\textbf{On Qwen1.5 and OLMoE, the first-token change runs through the refusal direction.} We estimate a refusal direction at each layer as the difference of mean residual-stream activations at the final prompt token between 256 AdvBench and 256 Alpaca requests wrapped in the refusal-suppression template, in the original model \citep{arditi2024refusal}. Truncation moves the final prompt token along it toward the harmless mean, by 0.17 harmful--harmless gaps on Qwen1.5 and 0.12 on OLMoE over the second half of the layers. We edit the residual stream at that token in those layers, let the edited model produce the first token, and let the original model continue, as in early truncation (Appendix~A); the unedited pipeline reproduces original ASR ($+0.2$ and $-0.4$\pp). In the truncated model, restoring the original position along the direction lowers the first-token increase from $+15.5$ to $+0.4$\pp on Qwen1.5 (82 against 11 discordant prompts with the truncated model, $p<10^{-6}$; 97\% removed, 95\% CI $[80, 118]$) and from $+5.1$ to $+1.5$\pp on OLMoE (28 against 11, $p=0.0095$; 71\%, $[27, 123]$). Restoring it only in layers $L/2$ to $3L/4$ suffices on both models ($+1.3$ and $+0.8$\pp), and one middle layer suffices on OLMoE ($+1.7$\pp) but not on Qwen1.5 ($+12.9$\pp). Restoring random directions, or pushing along them by the same amount, leaves the increase in place on both models. In the original model, setting the projection on the direction at each edited layer to its value in the truncated model (a set, like restore, so it matches truncation's shift in size at every layer and does not accumulate) reproduces all of the increase on Qwen1.5 and 75\% of it on OLMoE ($+15.5$ and $+3.8$\pp), largely on the same prompts: 64 of the 87 Qwen1.5 prompts and 16 of the 31 OLMoE prompts that the truncated first token turns harmful also turn harmful here (28 and 2 expected by chance), while the same-size push along random directions does not ($-2.1$ to $+1.7$\pp). Restoring everything except the direction is a stronger edit: it resets every other component of the state at every edited layer to its original value, returns the original first token on 445 and 423 of 472 prompts, and leaves $+2.5$\pp on Qwen1.5 and $+3.6$\pp on OLMoE.

\textbf{Truncation flips the refusals nearest the decision.} Under refusal suppression, a low probability of the original model's greedy first token predicts which of its refusals truncation turns harmful, with AUC 0.78 ($[0.72, 0.83]$) on Qwen1.5 and 0.73 ($[0.67, 0.79]$) on OLMoE: 70\% of the least confident third flip on Qwen1.5 against 13\% of the most confident third (28 and 4\% on OLMoE). Along the probe's refusal direction, truncation moves 99.6\% of these refusals on Qwen1.5 and 95.8\% on OLMoE toward the harmless side, by 0.068 and 0.044 harmful--harmless gaps on average (SD 0.037 and 0.032, a third of the spread of the original positions), and the refusals it flips move no further than those it keeps (0.068 against 0.067 and 0.040 against 0.045 gaps; AUC 0.53 and 0.47). Truncation thus pushes almost every refusal the same way, and the push flips those whose first token is least confident. Low confidence also marks the refusals that other perturbations flip, more weakly for the four reordered reruns of the original model (AUC 0.67 and 0.63, pooled) and about as strongly for swap (0.72 and 0.74), but the reruns flip as many harmful responses back (42 against 42 on Qwen1.5, 39 against 38 on OLMoE), while truncation flips 101 against 14 and 56 against 11.

\textbf{Causal patching.} For Qwen1.5 ($k'=3$) and OLMoE ($k'=6$) under refusal suppression, we recompute the refusal direction from AdvBench and Alpaca requests wrapped in the refusal-suppression template, so that the template text is the same in both classes. We feed vLLM's prompt token identifiers to the Transformers model, edit the residual stream at the final prompt token after each of layers $L/2$ to $L-2$ (forward hooks, applied in order so later layers see earlier edits), take the greedy first token, and continue from it with the original model in vLLM (greedy, 512 tokens), as in early truncation. \emph{Restore} sets the projection on a unit direction to its value in the unedited original model; \emph{push} adds a fixed displacement. Random directions are three Gaussian unit vectors per layer, and a random push uses the per-prompt, per-layer size of the refusal-direction edit. Table~\ref{tab:patch} gives the results under both judges. Restoring everything except the direction (setting the final-prompt-token state at each edited layer to its original value while keeping the truncated model's projection on the direction) tests the direction with every other component held at its original value, and restoring the direction in narrower layer windows tests location; we ran these on both models, together with a pipeline null in which the Transformers original model produces the first token. Against the truncated model, restoring the direction changes 82 against 11 discordant prompts on Qwen1.5 and 28 against 11 on OLMoE; the share of the first-token increase removed is 97\% (95\% bootstrap CI $[80, 118]$) and 71\% ($[27, 123]$). The truncation-induced change of the final-prompt-token state puts 6.5\% (Qwen1.5) and 6.0\% (OLMoE) of its squared norm on the direction, against 0.06\% on a random direction, so a random push of the same size moves the state far less along the direction than the restore does. On DeepSeek-MoE-16B at $k'=3$ ($+13.77$\pp under truncation), the truncated model's first token followed by the original model gives $-6.57$\pp, imposing the truncated model's shift along the direction on the original model gives $+6.78$\pp ($p<10^{-4}$), random pushes give $-3.60$ to $+0.85$\pp, and the pipeline null is $-2.75$\pp ($p=0.019$).

\begin{table*}[h]
\caption{Causal patching at the final prompt token under refusal suppression (472 prompts). The edited model produces the first token and the original model continues. Change in ASR (pp) against the original model; bold: exact McNemar $p<0.05$ (for ranges, in all three); ranges over three random directions (a single value: all three agree); shading: size and sign of the Llama Guard 3 change (for ranges, of their midpoint). Random push: a move along a random direction by the same amount as the refusal-direction edit.}
\label{tab:patch}
\vskip 0.05in
\begin{center}
\begin{scriptsize}
\setlength{\tabcolsep}{2pt}
\begin{tabular}{@{}lrrrrrr@{}}
\toprule
 & \multicolumn{2}{c}{Qwen1.5, $4\to3$} & \multicolumn{2}{c}{OLMoE, $8\to6$} & \multicolumn{2}{c}{DeepSeek-MoE-16B, $6\to3$} \\
\cmidrule(lr){2-3}\cmidrule(lr){4-5}\cmidrule(l){6-7}
Condition & LG & HB & LG & HB & LG & HB \\
\midrule
Original (pipeline null) & \cellcolor{vizorange!6}$+0.21$ & $+0.85$ & \cellcolor{vizaxis!6}$-0.42$ & $+0.00$ & \cellcolor{vizaxis!8}$\mathbf{-2.75}$ & $-2.33$ \\
Truncated & \cellcolor{vizorange!43}$\mathbf{+15.47}$ & $\mathbf{+13.56}$ & \cellcolor{vizorange!14}$\mathbf{+5.08}$ & $\mathbf{+3.81}$ & \cellcolor{vizaxis!18}$\mathbf{-6.57}$ & $\mathbf{-7.42}$ \\
Truncated, refusal direction restored & \cellcolor{vizorange!6}$+0.42$ & $-1.27$ & \cellcolor{vizorange!6}$+1.48$ & $+1.06$ & \cellcolor{vizaxis!33}$\mathbf{-11.86}$ & $\mathbf{-10.38}$ \\
\quad at layer $L/2$ only & \cellcolor{vizorange!36}$\mathbf{+12.92}$ & $\mathbf{+9.53}$ & \cellcolor{vizorange!6}$+1.69$ & $+1.48$ & \cellcolor{vizaxis!24}$\mathbf{-8.90}$ & $\mathbf{-8.05}$ \\
\quad at layers $L/2$ to $3L/4$ & \cellcolor{vizorange!6}$+1.27$ & $-0.42$ & \cellcolor{vizorange!6}$+0.85$ & $+0.21$ & \cellcolor{vizaxis!30}$\mathbf{-11.02}$ & $\mathbf{-11.02}$ \\
\quad at layers $3L/4$ to $L-2$ & \cellcolor{vizorange!19}$\mathbf{+6.78}$ & $\mathbf{+6.14}$ & \cellcolor{vizorange!6}$+2.12$ & $+1.91$ & \cellcolor{vizaxis!26}$\mathbf{-9.53}$ & $\mathbf{-8.90}$ \\
Truncated, all but the direction restored & \cellcolor{vizorange!7}$+2.54$ & $+1.27$ & \cellcolor{vizorange!10}$\mathbf{+3.60}$ & $\mathbf{+3.18}$ & \cellcolor{vizorange!6}$+1.06$ & $+1.91$ \\
Truncated, random direction restored & \cellcolor{vizorange!44}$\mathbf{+15.25}$ to $\mathbf{+16.74}$ & $\mathbf{+12.29}$ to $\mathbf{+13.98}$ & \cellcolor{vizorange!13}$\mathbf{+4.66}$ & $+2.54$ to $+3.39$ & \cellcolor{vizaxis!17}$\mathbf{-6.78}$ to $\mathbf{-5.93}$ & $\mathbf{-7.42}$ to $\mathbf{-5.51}$ \\
Truncated, random push & \cellcolor{vizorange!48}$\mathbf{+17.16}$ to $\mathbf{+18.01}$ & $\mathbf{+13.35}$ to $\mathbf{+15.25}$ & \cellcolor{vizorange!12}$\mathbf{+3.18}$ to $\mathbf{+5.30}$ & $+1.06$ to $+3.60$ & \cellcolor{vizaxis!18}$\mathbf{-7.20}$ to $\mathbf{-5.93}$ & $\mathbf{-7.20}$ to $\mathbf{-6.14}$ \\
Original, projection set to the truncated value & \cellcolor{vizorange!43}$\mathbf{+15.47}$ & $\mathbf{+11.86}$ & \cellcolor{vizorange!10}$\mathbf{+3.81}$ & $\mathbf{+3.39}$ & \cellcolor{vizorange!19}$\mathbf{+6.78}$ & $\mathbf{+6.78}$ \\
Original, random push & \cellcolor{vizaxis!6}$-2.12$ to $-1.69$ & $-2.54$ to $-0.85$ & \cellcolor{vizorange!6}$+0.85$ to $+1.69$ & $-0.64$ to $+0.64$ & \cellcolor{vizaxis!6}$-3.60$ to $+0.85$ & $-2.75$ to $+1.91$ \\
\bottomrule
\end{tabular}
\end{scriptsize}
\end{center}
\end{table*}

\subsection{A Residual-Stream Probe}
\label{app:probe}

\textbf{Setup.} We load each model in Hugging Face Transformers 4.57 with BF16 weights and run one forward pass per request with the chat template and generation prompt, without generating. Truncation sets the per-layer top-$k$ of every MoE block. We record the residual stream after each layer and the next-token distribution at the final prompt token, which produces the first generated token. The refusal direction at layer $l$ is $d_l=\bar h_l^{\mathrm{harm}}-\bar h_l^{\mathrm{safe}}$ over 256 AdvBench behaviors \citep{zou2023universal} and 256 Alpaca instructions without input \citep{taori2023alpaca}, computed with the original model on plain requests; the position of a request is $(h_l\cdot\hat d_l-m_l)/\|d_l\|$, where $m_l$ is the midpoint of the two class means along $\hat d_l$. The refusal-opener probability is the total probability of the first tokens of ``I'm sorry'', ``I am sorry'', ``I cannot'', ``I can't'', ``Sorry'', ``I apologize'', and ``As an AI'', with and without a leading space. Qwen3-30B (FP8), Qwen2-57B (GPTQ), and GPT-OSS-20B are not probed: the first two weight formats do not load in this stack on our GPUs, and GPT-OSS writes a reasoning channel before its answer, so its refusal is not decided at the first token.

\begin{table*}[h]
\caption{Residual-stream probe at the final prompt token (second half of the layers), with the refusal direction estimated from plain AdvBench and Alpaca requests; along the template-specific direction of the causal patching, truncation's shift under refusal suppression is 0.17 and 0.12 gaps on Qwen1.5 and OLMoE. Pos.: original position along the refusal direction relative to the harmful--harmless midpoint; Shift: change under truncation; both in units of the harmful--harmless gap. $\Delta\log P$: change in the log-probability of a refusal opener at the first generated token. $\Delta$ASR: change in pp under Llama Guard 3; shading: its size and sign.}
\label{tab:probe}
\vskip 0.05in
\begin{center}
\begin{scriptsize}
\setlength{\tabcolsep}{3pt}
\begin{tabular}{@{}llrrrrrrrr@{}}
\toprule
 & & \multicolumn{4}{c}{Plain} & \multicolumn{4}{c}{Refusal suppression} \\
\cmidrule(lr){3-6}\cmidrule(l){7-10}
Model & $k\to k'$ & Pos. & Shift & $\Delta\log P$ & $\Delta$ASR & Pos. & Shift & $\Delta\log P$ & $\Delta$ASR \\
\midrule
Qwen1.5-MoE & $4\to 3$ & $+0.29$ & $-0.081$ & $-0.04$ & $+0.00$ & $-0.22$ & $-0.052$ & $-0.93$ & \cellcolor{vizorange!51}$+18.43$ \\
Qwen1.5-MoE & $4\to 2$ & $+0.29$ & $-0.240$ & $-0.29$ & \cellcolor{vizorange!14}$+5.08$ & $-0.22$ & $-0.081$ & $-1.38$ & \cellcolor{vizorange!55}$+22.25$ \\
DeepSeek-V2-Lite & $6\to 4$ & $+0.31$ & $-0.077$ & $-0.22$ & \cellcolor{vizorange!9}$+3.39$ & $-0.25$ & $-0.031$ & $-0.06$ & \cellcolor{vizorange!18}$+6.57$ \\
DeepSeek-V2-Lite & $6\to 3$ & $+0.31$ & $-0.255$ & $-0.65$ & \cellcolor{vizorange!29}$+10.59$ & $-0.25$ & $-0.045$ & $-0.41$ & \cellcolor{vizorange!37}$+13.35$ \\
OLMoE & $8\to 6$ & $+0.29$ & $-0.045$ & $-0.02$ & \cellcolor{vizaxis!6}$-0.64$ & $-0.23$ & $-0.038$ & $-0.58$ & \cellcolor{vizorange!26}$+9.53$ \\
OLMoE & $8\to 4$ & $+0.29$ & $-0.214$ & $-0.20$ & \cellcolor{vizorange!13}$+4.87$ & $-0.23$ & $-0.109$ & $-2.34$ & \cellcolor{vizorange!55}$+36.02$ \\
DeepSeek-MoE-16B & $6\to 4$ & $+0.28$ & $-0.047$ & $-0.09$ & $+0.00$ & $-0.18$ & $+0.003$ & $+0.41$ & \cellcolor{vizaxis!6}$-0.64$ \\
DeepSeek-MoE-16B & $6\to 3$ & $+0.28$ & $-0.118$ & $-0.28$ & \cellcolor{vizorange!14}$+5.08$ & $-0.18$ & $-0.017$ & $+0.25$ & \cellcolor{vizorange!38}$+13.77$ \\
Moonlight-16B & $6\to 4$ & $+0.30$ & $+0.015$ & $+0.15$ & \cellcolor{vizaxis!13}$-4.87$ & $-0.32$ & $-0.024$ & $+0.33$ & \cellcolor{vizaxis!15}$-5.30$ \\
Moonlight-16B & $6\to 3$ & $+0.30$ & $+0.008$ & $+0.57$ & \cellcolor{vizaxis!22}$-8.05$ & $-0.32$ & $-0.031$ & $+0.67$ & \cellcolor{vizaxis!15}$-5.30$ \\
Ling-lite-1.5 & $6\to 4$ & $+0.28$ & $+0.053$ & $-0.02$ & \cellcolor{vizorange!6}$+0.85$ & $-0.12$ & $+0.016$ & $-0.01$ & \cellcolor{vizorange!6}$+0.85$ \\
Ling-lite-1.5 & $6\to 3$ & $+0.28$ & $+0.070$ & $-0.08$ & \cellcolor{vizorange!9}$+3.39$ & $-0.12$ & $+0.028$ & $+0.35$ & \cellcolor{vizorange!6}$+0.21$ \\
\bottomrule
\end{tabular}
\end{scriptsize}
\end{center}
\end{table*}

\textbf{Results.} Table~\ref{tab:probe} gives the measurements at both operating points. Across the 36 combinations of model, operating point, and framing (Table~\ref{tab:probe} lists the 24 without prefix injection), the shift along the refusal direction tracks the change in ASR with Spearman $\rho=-0.48$; after removing each model's mean, $\rho=-0.12$, and over the six model means $\rho=-0.66$, so the association is between models. The change in refusal-opener probability tracks the change in ASR more closely ($\rho=-0.75$), but it is measured at the token whose choice decides most flips, so it restates the first-token result rather than adding a residual-stream one; its separation of forward flips from stable refusals varies widely (AUC 0.25 to 0.95; several combinations have fewer than 20 forward flips). On DeepSeek-MoE-16B at $k'=3$ under refusal suppression, ASR rises by $+13.77$\pp while the refusal-opener probability also rises. On plain requests and under refusal suppression, the ASR change is nearly proportional to the change in the refusal-opener log-probability, with a common slope: a fit through the origin over the 24 combinations (six models, two operating points, two framings) gives $-15.2$\pp per nat ($r=-0.91$), and refitting without each model gives $-14.4$ to $-15.8$ (median absolute error on the left-out model 1.4\pp). The sign agrees in 16 of 17 combinations with $|\Delta\mathrm{ASR}|\ge2$\pp; the exception is DeepSeek-MoE-16B at $k'=3$ under refusal suppression, whose increase does not run through the first token. Without it and DeepSeek-V2-Lite under refusal suppression (where the two engines agree on 54\% of first tokens), $r=-0.98$ and the leave-one-model-out error is at most 5.6\pp. Under prefix injection, whose template dictates the opening, the slope is $-3.1$\pp per nat ($r=-0.58$). One prefill pass of both models thus predicts the sign and approximate size of the change for five of the six probed models.

\textbf{Limits of the probe.} The direction comes from plain requests, while the jailbreak templates append the same text to every request, so positions under a template reflect that text and the midpoint is not a decision boundary under it. The opener set reduces to the first tokens ``I'', ``As'', and ``Sorry'' (and ``I'm'' for Moonlight), with and without a leading space, and the refusal-suppression template forbids apologies and the word ``as''. The probe runs in Transformers while the labels come from vLLM: the greedy first token agrees between the two engines on 91 to 100\% of prompts for five models in every framing, and on DeepSeek-V2-Lite on 91\% of plain, 54\% of refusal-suppression, and 73\% of prefix-injection prompts.

\textbf{A one-token probe.} We also recorded the original and modified models' distributions over the first generated token with one prefill pass per prompt. The share of prompts whose top first token changes detects a perturbation but does not predict the direction of the ASR change: across 36 condition and framing pairs on two models its Spearman correlation with the ASR change is 0.39, and restoring the mass on DeepSeek-V2-Lite at $k'=3$ changes the first token of 56.6\% of refusal-suppression prompts while lowering ASR by 12.92\pp. A first-token check can flag a serving change for a full audit but cannot replace it.

\section{Additional Expert-Reduction Methods}
\label{app:additional}

\subsection{A Published Budget-Allocation Method}
\label{app:lexi}

LExI~\citep{lexi} allocates a fixed routed-expert budget $B$ across layers, choosing per-layer $k_j$ by Gaussian-input sensitivity profiling and an evolutionary search. We use a paper-based reimplementation rather than official code. The paper does not specify the gate convention; we preserve each model's native non-renormalizing routed weights. Shared experts and dense layers remain unchanged and do not count toward $B$. The original routed budgets are 96, 128, and 156 on Qwen1.5-MoE, OLMoE-0125, and DeepSeek-V2-Lite, respectively.

\textbf{Evaluation protocol.} This is a separate evaluation on the 448-behavior split of Study~1 (Appendix~\ref{app:study1}): the 288 AdvBench requests and 160 HarmBench standard behaviors selected with seed 0. We use the StrongREJECT refusal-suppression template, native model chat templates, greedy decoding with at most 256 new tokens, and Llama Guard 3. Each condition uses the same 112 mixed batches of four targets and 12 benign Alpaca fillers; only the 448 targets enter ASR. The model checkpoints are Qwen1.5-MoE-A2.7B-Chat, OLMoE-1B-7B-0125-Instruct, and DeepSeek-V2-Lite-Chat. This differs from Study~2's 472 public requests and 512-token cap, so absolute ASR is not directly comparable to Table~2 of the paper. ``Original'' below means the unmodified model with its native routing and expert budget.

For the protected condition, \method uses original-model routing for the entire prefill and the forward passes producing generated tokens 1--8, then applies the same LExI allocation starting with token 9. The KV cache from the protected passes is retained. Allocation, prompts, batches, and decoding settings are fixed across the paired conditions.

\textbf{ASR and opening-token protection.} The observed original/LExI ASR is 35.7/51.8\% on Qwen1.5, 15.6/46.4\% on OLMoE, and 51.3/67.0\% on DeepSeek-V2-Lite. With \method, ASR is 36.4, 18.5, and 53.3\%, respectively. These point estimates show that protecting the opening removes most of the ASR increase at each tested allocation. On Qwen1.5, the protected arm differs from the original model by 0.7\pp (exact McNemar $p=0.58$). The Qwen1.5 budget sweep below reports paired tests for the unprotected allocations.

At the DeepSeek budget $B=130$, LExI uses $k=6$ on layers 1--20, $k=5$ on layer 21, and $k=1$ on layers 22--26. Shared experts and the dense first layer retain their original computation.

\begin{table}[H]
\centering
\small
\begin{tabular}{@{}lrrr@{}}
\toprule
Setting & Removed & ASR & $\Delta$ (pp) \\
\midrule
Original ($B=96$) & 0\% & 35.7\% & --- \\
LExI $B=48$ & 50\% & 37.9\% & $+2.2$ \\
LExI $B=56$ & 41.7\% & 38.6\% & $+2.9$ \\
LExI $B=72$ & 25\% & 46.2\% & $+10.5^{*}$ \\
LExI $B=80$ & 16.7\% & 51.8\% & $+16.1^{*}$ \\
\midrule
Uniform $k=3$ & 25\% & 60.3\% & $+24.6^{*}$ \\
\bottomrule
\end{tabular}
\caption{Qwen1.5 LExI budget sweep on the same 448 targets. Removed: fraction of the original routed budget removed; $\Delta$: ASR change against the original model. $^{*}$Exact McNemar $p<0.05$. The two non-significant budgets are retained: the largest increase occurs at the least aggressive tested LExI budget.}
\label{tab:lexi_budgets}
\end{table}

\subsection{A Per-Token Expert-Skipping Method}
\label{app:ace}

ACE~\citep{xu2026ace} skips experts per token. We adapt its released HF/Triton implementation to Qwen1.5-MoE-A2.7B-Chat, OLMoE-1B-7B-0125-Instruct, and DeepSeek-V2-Lite-Chat; none of these three checkpoints is officially supported. Thresholds target approximately 25\% and 50\% skipping for Qwen1.5 and OLMoE, and 10\% for the reported DeepSeek setting. Threshold calibration uses WikiText text, without safety-label tuning. After skipping, our ports rescale surviving routed weights to their original total mass, rather than to one. Dense layers and shared experts remain unchanged. The protected arm uses the same threshold, preserves all prefill computation and the passes producing tokens 1--8, and enables ACE starting with token 9, retaining the KV cache.

\textbf{Protocol and implementation controls.} The safety experiment uses the same 448 targets as the LExI comparison, the StrongREJECT refusal-suppression template, native chat templates, Llama Guard 3 on the original request and final answer, greedy decoding, and a 256-token cap. Qwen1.5 retains its completed mixed batches of four targets and 12 Alpaca fillers and repetition penalty 1.05; OLMoE and DeepSeek use 28 target-only batches of 16. Conditions within each model share batches and decoding settings. These runs differ from the main evaluation's 472 public prompts, 512-token cap, repetition penalty 1.0, and vLLM implementation; absolute ASR should not be compared across the two harnesses.

ACE with skipping disabled ($\tau=0$) is an implementation control, not the baseline of the main ACE results. Its ASR is 35.94\%, 15.85\%, and 51.56\% on Qwen1.5, OLMoE, and DeepSeek, against 36.38\%, 14.96\%, and 53.12\% for the original models. The respective paired 95\% Tango intervals for its change are $[-2.51,1.55]$, $[-0.96,2.89]$, and $[-4.51,1.29]$\pp. Their original/control next-token losses are 2.3509/2.3499, 2.4640/2.4644, and 2.3705/2.3710. The DeepSeek adapter preserves native FP32 routing and both shared experts, and uses the checkpoint's tokenizer and required YaRN $m_{\mathrm{scale}}^2$ attention factor. DeepSeek agrees with the independent vLLM original-model safety run within sampling uncertainty (53.12\% versus 51.3\%, exact McNemar $p=0.29$) and matches its first token on 407 of 448 targets.

\begin{table*}[t]
\caption{ACE paired ASR changes (pp), 95\% Tango intervals, and upper 90\% bounds against the original model. $^{*}$Holm-significant exact McNemar test.}
\label{tab:ace_stats}
\centering
\scriptsize
\setlength{\tabcolsep}{4pt}
\begin{tabular}{@{}llrrrr@{}}
\toprule
Model & Skip target & ACE $-$ Original & Protected $-$ Original & Protected $-$ ACE & Upper 90\% \\
\midrule
Qwen1.5 & 25\% & $+9.15\ [5.77,12.83]^{*}$ & $-2.01\ [-4.12,-0.23]$ & $-11.16\ [-14.84,-7.85]^{*}$ & $-0.54$ \\
 & 50\% & $+20.98\ [15.99,26.01]^{*}$ & $-0.45\ [-2.22,1.23]$ & $-21.43\ [-26.34,-16.59]^{*}$ & $0.91$ \\
\midrule
OLMoE-0125 & 25\% & $+2.46\ [0.02,5.10]$ & $+1.12\ [-1.06,3.42]$ & $-1.34\ [-4.25,1.49]$ & $3.01$ \\
 & 50\% & $+12.95\ [9.11,17.04]^{*}$ & $+1.79\ [-0.65,4.38]$ & $-11.16\ [-15.24,-7.31]^{*}$ & $3.92$ \\
\bottomrule
\end{tabular}
\end{table*}

\textbf{Direction and non-inferiority.} On Qwen1.5 and OLMoE, the approximately 25\% condition is the primary ACE experiment; 50\% is an additional stress setting. Table~\ref{tab:ace_stats} reports paired changes against the original model and direct protection effects. Reported significance uses Holm correction over the full three-model test family at each budget and contrast. Excluding DeepSeek's high-skip outputs from safety interpretation does not reduce this correction. Non-inferiority requires the upper bound of the 90\% Tango interval against the original model to be below $+3$\pp. Protection meets this criterion on Qwen1.5 at both reported budgets. OLMoE's bounds are 3.01 and 3.92\pp, so neither establishes non-inferiority, despite point estimates close to the original model. The 95\% intervals and exact tests address different inferential questions; an interval excluding zero need not imply a Holm-significant exact test.

\textbf{DeepSeek operating point.} At approximately 10\% ACE skipping, DeepSeek-V2-Lite has 57.1\% ASR with ACE and 53.1\% with \method. Its 25\% and 50\% settings are excluded from safety interpretation because accuracy and perplexity checks indicate unreliable model output.

\textbf{Realized skipping.} Whole-run skipping for ordinary/protected ACE at the 25\% setting is 23.5/10.1\% on Qwen1.5 and 23.4/8.8\% on OLMoE; at the 50\% setting it is 48.7/20.7\% and 48.9/17.7\%, respectively. After the protected opening, protected ACE skips 25.0\% and 23.7\% at the lighter setting, and 50.1\% and 48.9\% at the heavier setting. Protecting prefill and the opening therefore reduces whole-run expert savings substantially; safety comparisons are at matched thresholds, not matched total expert work.

\clearpage

\end{document}